\documentclass{article}

\input{contents/packages}

\icmltitlerunning{Uncertainty for Active Learning on Graphs}

\begin{document}

\twocolumn[
\icmltitle{Uncertainty for Active Learning on Graphs}

\icmlsetsymbol{equal}{*}

\begin{icmlauthorlist}
\icmlauthor{Dominik Fuchsgruber}{equal,tum,mdsi}
\icmlauthor{Tom Wollschläger}{equal,tum,mdsi}
\icmlauthor{Bertrand Charpentier}{tum}
\icmlauthor{Antonio Oroz}{tum}
\icmlauthor{Stephan Günnemann}{tum,mdsi}
\end{icmlauthorlist}

\icmlaffiliation{tum}{School of Computation, Information and Technology, Technical University of Munich, Germany}
\icmlaffiliation{mdsi}{Munich Data Science Institute, Germany}

\icmlcorrespondingauthor{Dominik Fuchsgruber}{d.fuchsgruber@tum.de}
\icmlcorrespondingauthor{Tom Wollschläger}{tom.wollschlaeger@tum.de}

\icmlkeywords{Machine Learning, ICML, Active Learning, Uncertainty Estimation, Uncertainty, Graphs, Geometric Deep Learning}

\vskip 0.3in
]

\printAffiliationsAndNotice{\icmlEqualContribution} %

\begin{abstract}
Uncertainty Sampling is an Active Learning strategy that aims to improve the data efficiency of machine learning models by iteratively acquiring labels of data points with the highest uncertainty. While it has proven effective for independent data its applicability to graphs remains under-explored. We propose the first extensive study of Uncertainty Sampling for node classification:
\textbf{(1)} 
We benchmark Uncertainty Sampling beyond predictive uncertainty and highlight a significant performance gap to other Active Learning strategies. \textbf{(2)} We develop ground-truth Bayesian uncertainty estimates in terms of the data generating process and prove their effectiveness in guiding Uncertainty Sampling toward optimal queries. We confirm our results on synthetic data and design an approximate approach that consistently outperforms other uncertainty estimators on real datasets.
\textbf{(3)} Based on this analysis, we relate pitfalls in modeling uncertainty to existing methods. Our analysis enables and informs the development of principled uncertainty estimation on graphs.
\end{abstract}

\section{Introduction}

\looseness=-1
Applications in machine learning are often limited by their data efficiency. This encompasses effort spent on experimental design \citep{sverchkov2017review} or the cost of training on large datasets ~\citep{cui2022allie}. To remedy these problems, Active Learning (AL) allows the learner to query an oracle (e.g. users, machines, or experiments) to label specific data points considered \emph{informative}, thus saving labeling labor and training effort that would have been spent on \emph{uninformative} labeled data.

\looseness=-1
Uncertainty Sampling (US) methods \citep{8579074, Joshi2009MulticlassAL} rely on uncertainty estimates to measure the informativeness of labeling each data point. Intuitively, areas where a learner lacks knowledge are assigned high uncertainty. Moreover, methods should distinguish the (irreducible) \textit{aleatoric} uncertainty and the (reducible) \textit{epistemic} uncertainty which the \emph{total} uncertainty about a prediction is composed of \citep{KIUREGHIAN2009105, kendall2017uncertainties}. This disentanglement is particularly important for AL: The learner might not benefit much from labeling instances with high irreducible uncertainty while acquiring knowledge about data points for which uncertainty stems from reducible sources can be highly informative. This suggests epistemic uncertainty as a sensible acquisition function \citep{nguyen2022howto}.

\looseness=-1
For independent and identically distributed (i.i.d.) data, US methods for AL---in particular US methods disentangling aleatoric and epistemic uncertainty---have demonstrated high data efficiency in benchmarks \citep{,  Joshi2009MulticlassAL, DBLP:journals/corr/GalIG17,8579074,kirsch2019batchbald, nguyen2022howto,schmidt2023streambased}. 
Existing efforts in AL for interdependent data like graphs neglect the compositional nature of uncertainty \citep{zhu2003combining,jun2016graph, regol2020active, wu2021active, zhang2022propagation} and limit themselves to a singular measure of total uncertainty.
This leaves it unclear \begin{inparaenum}[(i)]
    \item to which extent uncertainty estimators can effectively inform US for graph data, and 
    \item whether disentangling aleatoric and epistemic uncertainty has similar benefits to AL as in i.i.d. settings.
\end{inparaenum}
The complex nature of graph data makes investigating US methods for AL particularly challenging. Uncertainty estimates should not only capture information about node features independently but also model information about their relationships \citep{stadler2021graph}.
\begin{figure*}
    \centering
    \vspace{-2mm}
    \includegraphics[width=.99\textwidth]{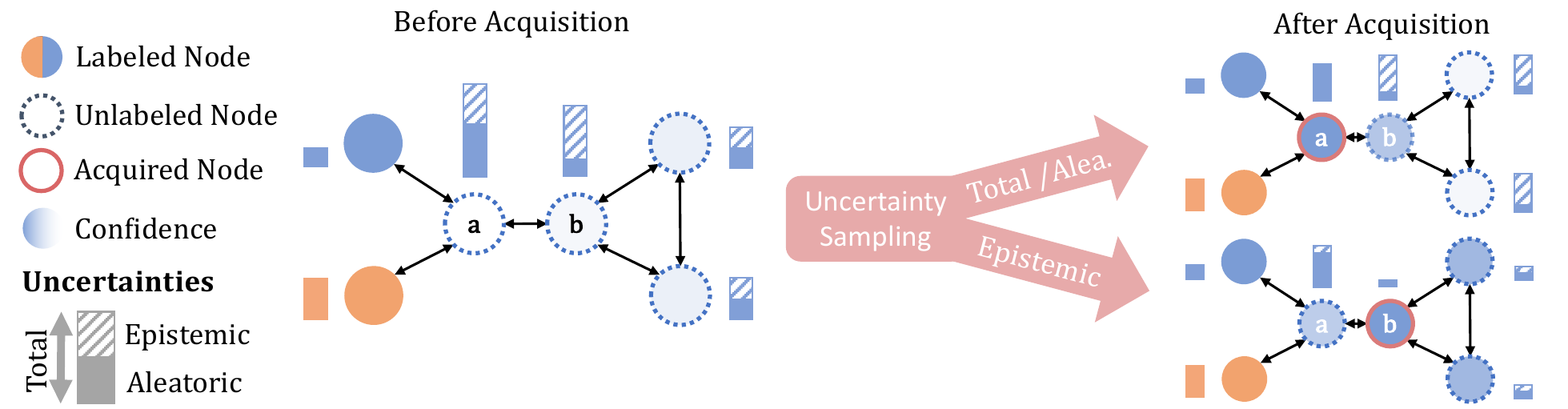}
        \vspace{-1mm}
    \caption{US can be realized by acquiring the label of a node with with maximal total, aleatoric or epistemic uncertainty. The former two include irreducible effects leading to label node $a$ while the latter isolates epistemic factors and queries the label of node $b$, thereby increasing the confidence in correctly predicting the remaining unlabeled nodes the most.}
    \label{fig:uncertainty_picture}
    \vspace{-4mm}
\end{figure*} 

In this work, we examine US for node classification problems. We critically evaluate and benchmark state-of-the-art uncertainty estimators against traditional AL strategies and find that US (as well as many other approaches) fall short of surpassing random sampling. Motivated by the effectiveness of US for i.i.d. data, we formally approach the question of whether US methods for graphs align well with the AL objective at all. We derive novel ground-truth uncertainty measures from the underlying data-generating process and disentangle its aleatoric and epistemic components. This generative perspective enables a principled perspective on handling instance interdependence in uncertainty estimation that we leverage for AL:
We prove that querying epistemically uncertain nodes is equivalent to maximizing the relative gain in confidence a Bayesian classifier puts on correctly predicting all unlabeled nodes.

\Cref{fig:uncertainty_picture} shows that acquiring the most epistemically uncertain label can improve the confidence of the classifier in the correct prediction more than selecting a node associated with high total uncertainty that mainly stems from irreducible factors. While node $a$ is mainly uncertain because of its link to an orange node, $b$'s uncertainty comes from the the lack of labels in its neighbourhood. By employing our proposed estimates on a Contextual Stochastic Block Model (CSBM) \citep{Deshpande2018ContextualSB}, we empirically confirm the validity of our findings on synthetic data. Our analysis reveals that US is an effective strategy only if uncertainty \emph{is disentangled} into aleatoric and epistemic components. 

The main \textbf{contributions} of our work are:
\begin{itemize}[leftmargin=*]
\item We provide the first extensive AL benchmark\footnote{Find our code at \href{https://www.cs.cit.tum.de/daml/graph-active-learning/}{cs.cit.tum.de/daml/graph-active-learning/}} 
for node classification including both a broad range of state-of-the-art uncertainty estimators for US and traditional acquisition strategies. Many traditional methods and uncertainty estimators do not outperform random acquisition.

\item We derive ground-truth aleatoric and epistemic uncertainty for a Bayesian classifier and formally prove the alignment of US with AL.

\item We empirically confirm the efficacy of epistemic US %
both on synthetic and real data by employing an approximation to our proposed ground-truth uncertainty that outperforms SOTA uncertainty estimators off-the-shelf.
This enables future uncertainty estimators to achieve competitive AL performance by building on the principles of our work.

\end{itemize}

\section{Background}\label{sec:background}
\textbf{AL for Semi-Supervised Node classification.} Let $G = (\set V, \set E)$, be a graph with a set of nodes $\set V$ and  edges $\set E \subseteq \set \{\{i, j\} : i, j \in \set V\}$. With $n = \abs{V}$ number of nodes the adjacency matrix is defined as $\smash{\matr{A} \in \{0, 1\}^{n\times n}}$, where $A_{ij} = A_{ji} = 1$ if there exists an edge between node $i$ and node $j$. The features are represented as matrix $\smash{\matr X \in \mathbb{R}^{n \times d}}$. In node classification, every node has a label represented by a vector $\vect y \in [C]^n$. We decompose the set of all nodes into $\set V = \unlabeled \cup \labeled$ where $\labeled$ contains nodes with observed labels $\vect y_\labeled$ and $\unlabeled$ contains remaining unobserved nodes $\vect y_\unlabeled$ that we want to infer, where $\set V = \unlabeled \cup \labeled$. We consider pool-based AL: In each iteration, the learner queries an oracle for a label $\vect y_i$ from the set of unobserved labels $\vect y_\unlabeled$, adds it to the set of observed labels $\vect y_\labeled$ and retrains the model. 

\textbf{Uncertainty in Machine Learning.} Alongside the predicted label $\vect y_i$, it is crucial to consider the associated uncertainty \citep{KIUREGHIAN2009105, kendall2017uncertainties}. Aleatoric uncertainty $\uncertainty^{\alea}$ is the inherent uncertainty that comes from elements like experimental randomness. It can not be reduced by acquiring more data. %
On the other hand, Epistemic uncertainty $\uncertainty^{\epi}$ reflects knowledge gaps, which can be addressed by data acquisition. %
A commonly employed conceptualization \citep{depeweg2018decomposition} defines the total predictive uncertainty $\uncertainty^{\total}$ as encapsulating both reducible and irreducible factors, i.e. $\uncertainty^{\total} = \uncertainty^{\epi} + \uncertainty^{\alea}$.

\textbf{Contextual Stochastic Block Model.} We approach US on graphs sampled from an explicit generative process $p(\matr{A}, \matr{X}, \vect{y})$. To that end, we generate data from a Contextual Stochastic Blockmodel (CSBM) \citep{Deshpande2018ContextualSB} enabling a well-principled study of exact ground-truth uncertainty estimators. We first independently sample the node labels $\vect y$ independently from a prior $p(\vy)$. Node features are generated from class-conditional normal distributions $p(\bm{X}_i \mid \vy_i) \sim \mathcal{N}(\bm{\mu}_{\vy_i}, \sigma_x^2 \matr I)$ and edges are introduced independently according to an affiliation matrix $\matr F \in [0, 1]^{c \times c}$ as $p(\matr A_{i,j} \mid \vy_i, \vy_j) \sim \text{Ber}(\matr F_{\vy_i, \vy_j})$. We defer an in-depth description to \Cref{appendix:datasets}.

\section{Related Work} \label{sec:related_work}

\textbf{Active Learning on Independent and Identically Distributed Data.}
AL has seen substantial exploration in the context of i.i.d. data \citep{ren2021survey}. Approaches can be divided into three main categories: diversity-based, uncertainty-based, or a combination thereof \citep{zhan_comparative_2022}. Diversity or representation-based methods query data samples that best represent the full dataset, i.e. they opt for a diverse set of data points. Approaches like KMeans or Coreset  opt to minimize the difference in model loss between the selected training set and the whole data set \citep{sener2018active}. Other approaches use adversarial techniques to estimate the representativeness and diversity of new samples \citep{sinha_variational_2019, shui_deep_2020}. Uncertainty-based approaches query instances that the classifier is most uncertain about. It is commonly computed from the predictive distribution of a classifier calculating its entropy \citep{shannon_mathematical_1948}, the margin between the two most likely labels, or as the probability of the least confident label \citep{wang_new_2014}. \citet{houlsby_bayesian_2011} introduce Bayesian Active Learning by Disagreement (BALD), which queries points with high mutual information between the model parameters and the class label. Another line of work uses a loss-prediction module as a proxy for uncertainty \citep{Yoo_2019_CVPR,schmidt2023streambased}. \citet{nguyen_epistemic_2019} leverage disentangled uncertainties \citep{kendall2017uncertainties} for AL and propose that epistemic uncertainty is a better proxy for US than aleatoric estimates. Other approaches linearly combine diversity- and uncertainty-based measures \cite{yin_deep_2017} or employ a two-step optimization scheme \cite{ash_deep_2020, zhan_comparative_2022}. {BADGE \citep{ash2019deep} queries diverse and uncertain instances in the gradient space of the model using clustering. Another line of work constructs similarity graphs between instances to enforce diversity among queries \citep{pmlr-v40-Dasarathy15}. GALAXY \citep{pmlr-v162-zhang22k} proposes further refines this approach by mitigating class imbalance. In contrast, our work concerns settings where the adjacency matrix $A$ is explicitly given
.}

\textbf{Active Learning on Interdependent Graph Data.}
While a plethora of studies exist on AL for i.i.d. data, only a limited amount of work addresses interdependent data like graphs. Previous methods approach AL on graphs from different perspectives, including random fields \citep{zhu2003combining, ma2013sigma, ji2012variance, berberidis2018data}, risk minimization \citep{jun2016graph, regol2020active}, adversarial learning \citep{li2020seal}, knowledge transfer between graphs \citep{hu2020graph} or querying cheap soft labels \citep{zhang2022information}. US is typically only considered in terms of the predictive distribution \citep{Madhawa2020ActiveLF} which does not disentangle aleatoric and epistemic components. Also, other uncertainty-reliant approaches do not make that distinction \citep{cai2017active, ijcai2018p296, li2022smartquery} even though the literature on i.i.d. data suggests that US benefits from disentangled uncertainty estimators \citep{nguyen2022howto, Sharma2016EvidencebasedUS}. The exploration of US for AL on graphs beyond total uncertainty remains uncharted territory. Our work targets this gap and showcases the unrealized potential of epistemic uncertainty estimators for AL on graph data.

\textbf{Uncertainty Estimation on Graphs.}
US strategies necessitate accurate uncertainty estimates which can be obtained in different ways.
In classification, the predictive distribution of deterministic classifiers has been used to obtain aleatoric uncertainty \citep{stadler2021graph}. The Graph Convolutional Network (\textbf{GCN}) \citep{kipf2017semisupervised} iteratively updates the representation $H^{(l)}$ of each node using a linear transformation $W^{(l)}$ and subsequent diffusion along the edges and non-linearity $\sigma$. Formally, a GCN layer is expressed as $f(H^{(l)}, A) = \sigma(AH^{(l)}W^{(l)})$. The \textbf{APPNP} model \citep{DBLP:journals/corr/abs-1810-05997} first transforms the nodes features independently and then diffuses predictions according to approximate Personalized Page Rank (PPR) scores.
Bayesian approaches model the posterior distribution over the model parameters. 
\textbf{Ensembles} \citep{lakshminarayanan2017simple} fall into this category. They approximate the posterior over model parameters through a collection of independently trained models. \textbf{Monte Carlo Dropout} (MC-Dropout) \cite{gal2016dropout} instead emulates a distribution over model parameters by applying dropout at inference time. DropEdge \citep{Rong2020DropEdge} proposed to additionally drop edges to reduce over-fitting and over-smoothing. Variational Bayes methods place a prior on the model parameters (\textbf{BGCN}) and sample different parameter sets for each forward pass \citep{blundell2015weight}. They allow access to disentangled uncertainty estimates by approximating a distribution over predictions from multiple forward passes. Commonly employed measures of epistemic uncertainty are the mutual information between the model weights and predicted labels \citep{gawlikowski2023survey} or the variance in confidence about the predicted label \citep{stadler2021graph}. Evidential methods like \textbf{GPN} \citep{stadler2021graph} disentangle epistemic and aleatoric uncertainty by outputting the parameters of a Dirichlet prior to the categorical predictive distribution. This method has shown strong performance in detecting distribution shifts. Finally, Gaussian processes on graphs, while potentially leading to strong uncertainty estimates in specific domains \cite{wollschlaeger2023uncertainty}, they do not disentangle aleatoric and epistemic uncertainty \citep{liu2020uncertainty,borovitskiy2021gp}.

\section{Benchmarking Uncertainty Sampling Approaches for Active Learning on Graphs}\label{sec:al_methods}

Previous studies on non-uncertainty-based AL on graphs find that AL strategies struggle to consistently outperform random sampling \citep{Madhawa2020ActiveLF}. We, therefore, ask the question of whether US shows any merit when using state-of-the-art uncertainty estimation. We are the first to design a comprehensive AL benchmark for node classification that not only considers traditional methods but also includes a variety of uncertainty estimators for US. Our work answers the research question:

\begin{mybox}
    \textit{Does Uncertainty Sampling using state-of-the-art uncertainty estimators work on graph data?}
    
    \textbf{No}, no uncertainty estimator outperforms random sampling. Most non-uncertainty strategies fail to consistently improve over random queries as well.
    \vspace{-.1cm}
\end{mybox}
 
\textbf{Experimental Setup.} We evaluate AL on five common citation benchmark datasets for node classification: \textbf{CoraML} \citep{bandyopadhyay2005link}, \textbf{Citeseer} \citep{sen2008collective, giles1998citeseer}, \textbf{PubMed} \citep{namata2012query} as well as the co-purchase graphs \textbf{Amazon Photos} and \textbf{Amazon Computers} \citep{mcauley2015image}. We evaluate the models of \Cref{sec:related_work}: \textbf{GCN}, \textbf{APPNP}, \textbf{MC-Dropout}, \textbf{BGCN}, \textbf{GPN} and \textbf{Ensembles}
and report average results over multiple AL runs (see \Cref{appendix:experimental_setup}).

When acquiring multiple labels per iteration, greedily selecting the most promising candidates might overestimate the performance improvement \citep{kirsch2019batchbald}. We therefore only acquire a single label in each iteration, enabling us to analyze the performance of different acquisition strategies without having to consider the potential side effects of batched acquisition. We initially label one node per class and fix the acquisition budget to $4C$. %

\begin{figure}[t]
    \centering
\input{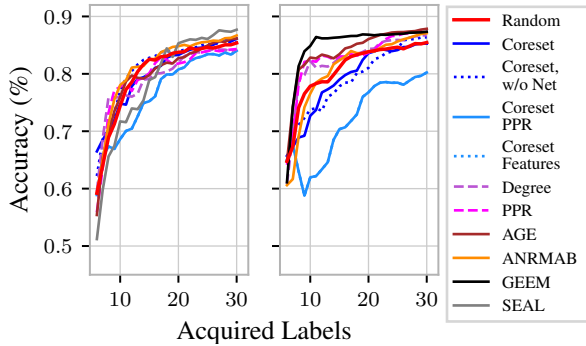}
\vspace{0.2mm}
    \caption{Accuracy of AL strategies on Citeseer using a GCN (left) / SGC (right) classifier. Except for GEEM, which is only tractable for SGCs, traditional AL can not significantly outperform random selection.}
    \label{fig:al_citeseer}
      \vspace{-6mm}
\end{figure}

In addition to a qualitative evaluation of the accuracy curves of different strategies in \Cref{fig:al_citeseer}, we report the accuracy after the labeling budget is exhausted in \Cref{tab:final_acc_with_sd}. Good acquisition functions should achieve higher accuracy at a lower amount of queries, which in turn results in a larger area under (AUC) the visualized curves. After normalization, this metric quantifies the average accuracy which we report as a summary of the AL performance in \Cref{tab:auc_no_sd}. 

We proceed as follows: First, we benchmark traditional AL approaches and find that only one consistently outperforms \textbf{Random} selection that acquires labels with uniform probability. Then, we show that US fails to even match the performance of random queries in many instances, contrasting its successful application to i.i.d. data. We supply the corresponding AUC and final accuracy scores as well as visualizations on all datasets in \Cref{appendix:additional_metrics_and_plots}.

\textbf{Non-Uncertainty-based Strategies.} We first examine strategies that do not exclusively rely on uncertainty.
\begin{inparaenum}[(i)]
    \item \textbf{Coreset} \citep{sener2018active} opts to find a core-set cover of the training pool by selecting nodes that maximize the minimal distance in the latent space of a classifier to already acquired instances.
    \item \textbf{Coreset-PPR} is similar to Coreset, but we use inverse Personalized Page Rank (PPR) scores as a distance measure to select structurally dissimilar nodes.
    \item \textbf{Coreset Features.} Distances between nodes are computed only in terms of input features.
    \item \textbf{Degree} and \textbf{PPR.} We acquire nodes with the highest corresponding centrality measure.
    \item \textbf{AGE} \citep{cai2017active} and  \textbf{ANRMAB} \citep{ijcai2018p296} combine total predictive uncertainty, informativeness, and representativeness metrics.
    \item \textbf{GEEM} \citep{regol2020active} uses risk minimization to select the next query. Because of its high computational cost, we follow its authors and employ an SGC \citep{wu2019simplifying} backbone.
    \item \textbf{SEAL} \citep{li2020seal} uses adversarial learning to identify nodes dissimilar from the labeled set.
\end{inparaenum}
If applicable, we also consider setting $\matr A = \matr I$ to exclude structural information similar to \cite{stadler2021graph}.

\underline{{Observations:}} \Cref{fig:al_citeseer} shows that only GEEM identifies a training set that is significantly more informative than random sampling. The performance of GEEM, however, comes at a high computational cost, as it requires training $\mathcal{O}(nC)$ models in each acquisition which makes it intractable for larger datasets and models beyond SGC. Structural Coreset strategies work well on both co-purchase networks but do not show strong results on other graphs. This highlights a notable gap between many commonly used acquisition functions and a potential optimal strategy.

\begin{figure}[t]
    \centering
\input{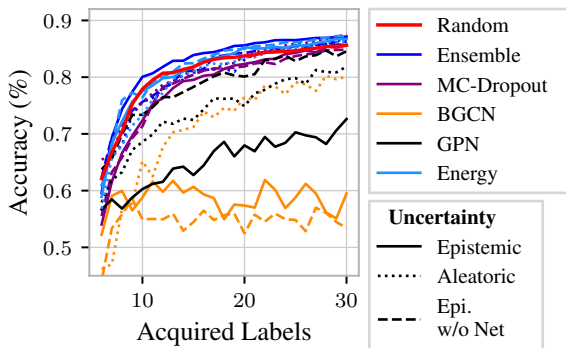}
      \vspace{1mm}
      \caption{US on Citeseer. No method significantly outperforms random selection.}
      \label{fig:uncertainty_citeseer}
      \vspace{-6.5mm}
\end{figure}

\textbf{Uncertainty Sampling.} We evaluate US using different uncertainty estimators: 
\begin{inparaenum}[(i)]
    \item \textbf{Aleatoric.} Like \cite{stadler2021graph}, we compute $\uncertainty^{\alea} = - \max_c \vect p_c$.
    \item \textbf{Epistemic.} For models perform multiple predictions (MC-Dropout, Ensembles, BGCNs), we compute epistemic uncertainty as the variance of the confidence $\smash{\uncertainty^\epi = \Var\left[ \vect p_{\hat{c}}\right]}$ in the predicted class $\smash{\hat{c}}$. For GPN, we follow the authors and use evidence as a measure of epistemic confidence.
    \item \textbf{Energy.} Energy-based models (EBMs) \citep{liu2021energybased, wu2023energybased} relate uncertainty to the energy $\smash{\uncertainty = -\tau \log \sum_c \exp\left( \vect l_i \right)}$ of the predicted logits $\vect l$. We apply this estimator to the deterministic GCN, APPNP, and SGC models as a surrogate for epistemic uncertainty.
\end{inparaenum}
 Again, we ablate each strategy by withholding structural information ($\matr A = \matr I$).

\begin{table*}[ht!]
\centering
\caption{Average AUC ($\uparrow$) for different acquisition strategies on different models and datasets. We mark the best strategy per model in bold and underline the runner-up. For each dataset, we highlight the overall best strategy with a $\dag$ symbol.}
\label{tab:auc_no_sd}
\resizebox{2\columnwidth}{!}{
\begin{tabular}{ll|c|ccccc|c|c|cc|cc}\Xhline{3\arrayrulewidth}
\multirow{3}{*}{\rotatebox[origin=c]{90}{}} &  & \multicolumn{1}{c|}{\makecell{\textbf{Baselines}}} & \multicolumn{7}{c|}{\makecell{\textbf{Non-Uncertainty}}} & \multicolumn{4}{c}{\makecell{\textbf{Uncertainty}}}\\
\hline & \makecell[r]{\textbf{Inputs}} & \multicolumn{1}{c|}{\makecell{}} & \multicolumn{5}{c|}{\makecell{$ A$ \& $ X$}} & \multicolumn{1}{c|}{\makecell{$ A$}} & \multicolumn{1}{c|}{\makecell{$ X$}} & \multicolumn{2}{c|}{\makecell{$ A$ \& $ X$}} & \multicolumn{2}{c}{\makecell{$ X$}}\\
 & \textbf{Model} & \makecell{Random} & \makecell{Coreset} & \makecell{AGE} & \makecell{ANRMAB} & \makecell{GEEM} & \makecell{SEAL} & \makecell{Coreset\\PPR} & \makecell{Coreset\\Inputs} & \makecell{Epi./\\(Energy)} & \makecell{Alea.} & \makecell{Epi./\\(Energy)} & \makecell{Alea.}\\
\hline \multirow{7}{*}{\rotatebox[origin=c]{90}{\makecell{CoraML}}} & GCN & $62.51$ & $64.35$ & $64.12$ & $64.24$ & \small{n/a} & \npboldmath{$66.07$} & $59.53$ & $61.26$ & $63.97$ & $61.30$ & \underline{$65.65$} & $64.33$\\
 & APPNP & $67.72$ & $67.32$ & $66.12$ & $69.49$ & \small{n/a} & \small{n/a} & \npboldmath{$71.04$} & $64.49$ & $64.92$ & $67.68$ & \underline{$69.59$} & $66.69$\\
 & Ensemble & $63.89$ & $60.55$ & $64.80$ & \underline{$65.10$} & \small{n/a} & \small{n/a} & $62.65$ & $65.07$ & $63.47$ & $64.03$ & $64.80$ & \npboldmath{$65.82$}\\
 & MC-Dropout & \npboldmath{$64.94$} & $64.37$ & \underline{$64.44$} & $64.06$ & \small{n/a} & \small{n/a} & $62.92$ & $64.35$ & $59.17$ & $63.69$ & $61.82$ & $63.87$\\
 & BGCN & $45.76$ & \underline{$49.37$} & \npboldmath{$51.25$} & $47.23$ & \small{n/a} & \small{n/a} & $39.43$ & $44.85$ & $44.45$ & $46.61$ & $42.74$ & $48.11$\\
 & GPN & $56.50$ & \small{n/a} & \small{n/a} & \small{n/a} & \small{n/a} & \small{n/a} & \npboldmath{$58.04$} & $54.02$ & $54.75$ & $57.16$ & $55.89$ & \underline{$57.21$}\\
 & SGC & $63.85$ & $65.23$ & \underline{$67.56$} & $61.14$ & \npboldmath{$71.39^\dagger$} & \small{n/a} & $60.24$ & $59.18$ & $67.51$ & $65.66$ & $65.05$ & $67.13$\\
\hline \multirow{7}{*}{\rotatebox[origin=c]{90}{\makecell{Pubmed}}} & GCN & $61.56$ & \underline{$62.61$} & \npboldmath{$69.48$} & $60.31$ & \small{n/a} & $58.62$ & $61.71$ & $56.71$ & $59.64$ & $61.85$ & $59.66$ & $60.34$\\
 & APPNP & \underline{$64.61$} & $63.88$ & \npboldmath{$70.18^\dagger$} & $63.83$ & \small{n/a} & \small{n/a} & $64.21$ & $56.87$ & $63.09$ & $62.37$ & $62.23$ & $63.95$\\
 & Ensemble & $59.26$ & \underline{$64.25$} & \npboldmath{$68.26$} & $60.40$ & \small{n/a} & \small{n/a} & $61.89$ & $56.36$ & $63.70$ & $61.37$ & $59.71$ & $61.15$\\
 & MC-Dropout & $58.30$ & \underline{$62.97$} & \npboldmath{$65.24$} & $60.50$ & \small{n/a} & \small{n/a} & $61.43$ & $56.01$ & $58.67$ & $59.07$ & $59.23$ & $62.22$\\
 & BGCN & $53.59$ & \npboldmath{$59.29$} & $56.93$ & $52.68$ & \small{n/a} & \small{n/a} & $53.40$ & $51.40$ & $55.19$ & $52.81$ & \underline{$57.09$} & $54.62$\\
 & GPN & \underline{$59.76$} & \small{n/a} & \small{n/a} & \small{n/a} & \small{n/a} & \small{n/a} & \npboldmath{$62.08$} & $54.34$ & $58.82$ & $57.24$ & $56.25$ & $59.37$\\
 & SGC & $56.79$ & $64.48$ & \npboldmath{$69.20$} & $60.49$ & \underline{$64.82$} & \small{n/a} & $62.15$ & $52.25$ & $62.04$ & $61.55$ & $61.04$ & $60.74$\\
\hline \multirow{7}{*}{\rotatebox[origin=c]{90}{\makecell{AmazonPhotos}}} & GCN & $79.06$ & $78.58$ & $75.17$ & \underline{$79.97$} & \small{n/a} & $71.16$ & $70.20$ & \npboldmath{$82.71$} & $74.66$ & $74.61$ & $79.96$ & $79.63$\\
 & APPNP & $79.29$ & \underline{$81.04$} & $79.02$ & $80.35$ & \small{n/a} & \small{n/a} & $76.37$ & \npboldmath{$84.24$} & $79.72$ & $77.45$ & $80.48$ & $77.69$\\
 & Ensemble & $82.23$ & $80.44$ & $77.45$ & $82.77$ & \small{n/a} & \small{n/a} & $74.93$ & \underline{$84.04$} & \npboldmath{$84.46$} & $77.85$ & $80.50$ & $81.25$\\
 & MC-Dropout & \underline{$80.32$} & $76.63$ & $74.75$ & $80.21$ & \small{n/a} & \small{n/a} & $75.32$ & \npboldmath{$82.45$} & $72.42$ & $73.16$ & $69.68$ & $78.80$\\
 & BGCN & $71.22$ & $67.15$ & $65.69$ & $70.69$ & \small{n/a} & \small{n/a} & $59.34$ & \npboldmath{$73.39$} & $70.83$ & $67.83$ & \underline{$72.21$} & $69.19$\\
 & GPN & $62.80$ & \small{n/a} & \small{n/a} & \small{n/a} & \small{n/a} & \small{n/a} & $55.59$ & \npboldmath{$65.07$} & $54.78$ & $60.53$ & \underline{$62.90$} & $62.41$\\
 & SGC & $80.52$ & $82.32$ & $74.01$ & $80.92$ & \npboldmath{$86.43^\dagger$} & \small{n/a} & $66.94$ & \underline{$84.24$} & $84.01$ & $71.43$ & $80.75$ & $76.38$\\
\Xhline{3\arrayrulewidth}
\end{tabular}

}
\vspace{-4mm}
\end{table*}

 \underline{{Observations:}} %
 In contrast to literature on i.i.d. data
 \citep{8579074, DBLP:journals/corr/GalIG17,nguyen2022howto}, we observe none of the US approaches to be effective in \Cref{fig:uncertainty_citeseer}. While sampling instances with high aleatoric uncertainty matches the performance of random queries, we find epistemic uncertainty to even underperform in many instances. This is surprising, as the efficacy of epistemic US has been demonstrated for i.i.d. data \citep{nguyen2022howto, kirsch2019batchbald}. Only ensemble models match the performance of random sampling and slightly outperform on some datasets. GPN and energy-based approaches can not guide US toward effective queries. This is an intriguing result as both uncertainty estimators have shown to be highly effective for of out-of-distribution detection \citep{stadler2021graph,wu2023energybased}. %

\section{Ground-Truth Uncertainty from the Data Generating Process}\label{sec:gnd_csbm}
As no existing US method yields satisfactory results, we formally answer the following research question:

\begin{mybox}
    \textit{Does US on graphs align with the AL objective?}
    \textbf{Yes}, we formally show that acquiring the node with maximal epistemic uncertainty optimizes the gain in the posterior probability of the ground-truth labels $\vect y_{\mathcal{U}}$ of unobserved nodes. 
\end{mybox}

\looseness=-1
Evaluating the quality of uncertainty estimates is inherently difficult as generally, ground-truth values are unavailable for both the overall predictive uncertainty $u^\text{total}$ and its constituents $u^\text{alea}, u^\text{epi}$. Additionally, since epistemic uncertainty pertains to the knowledge of the classifier, it cannot be defined in a model-agnostic manner. Therefore, we analyze uncertainty from the perspective of the underlying (potentially unknown) data-generating process $p(\matr X, \matr A, \vect y)$ with respect to a Bayesian classifier. This lends itself to a definition of ground-truth uncertainty. In the following, we propose confidence measures and relate them to uncertainty as their inverse $\confidence := \uncertainty^{-1}$. This allows us to state the main theoretical result of our work: The optimality of US using epistemic uncertainty.

\begin{definition}\label{def:alea_classifier}
    We define the parametrized Bayesian classifier $f_{\theta}^*(\matr A, \matr X, \vect y_\labeled^\gt)$ in terms of the data generating process $p(\matr A, \matr X, \vect y)$ as the prediction $\smash{\vect c \in [C]^{\lvert \unlabeled \rvert}}$ that maximizes:
    \begin{equation}
        \mathbb{E}_{p(\theta \mid \matr A, \matr X, \vect y_\labeled^\gt)}\left[\prob{\vect y_\unlabeled = \vect c \mid \matr A, \matr X, \vect y_\labeled = \vect y_\labeled^\gt, \theta}\right]
    \end{equation}
    
\end{definition}
\vspace{-5pt}
Here, we denote with $\smash[]{\vect y_\labeled^\gt}$ the labels of already observed instances. The predictive distribution $p\smash{(\vect y_\unlabeled \mid \matr A, \matr X, \vect y_\labeled^\gt)}$ encapsulates the total confidence of $\smash{f^*_\theta}$. The classifier averages its prediction according to a learnable posterior distribution $p(\theta \mid \matr A, \matr X, \vect y_\labeled)$ over its parameters, e.g. weights of a GNN. Marginalization yields the total confidence $\smash{\confidence^\total}$. 

\begin{definition}\label{def:conf_total}
    The total confidence $\confidence^\total(i, c)$ of $f^*_{\theta}$ in predicting label $c$ for node $i$ is defined as:
    \begin{multline}
         \mathbb{E}_{p(\theta \mid \matr A, \matr X, \vect y_\labeled^\gt)} \left[ \prob{\vect y_i = c \mid \matr A, \matr X, \vect y_\labeled=\vect y_\labeled^\gt, \theta} \right]
    \end{multline}
\end{definition}
\vspace{-5pt}
Intuitively, the total confidence captures aleatoric factors through the inherent randomness in the data generating process $\smash{p(\vect y, \matr A, \matr X)}$. Epistemic uncertainty is incorporated by conditioning on a limited set of observed labels $\smash{\vect y_\labeled^\gt}$. 
With a growing labeled set irreducible errors will increasingly dominate total predictive uncertainty.
In the extreme case where all labels but one have been observed, i.e. $\smash{\labeled = \mathcal{V} \setminus \{v_i\}}$, remaining uncertainty only stems from aleatoric factors.
\begin{definition}\label{def:conf_alea}
    The aleatoric confidence $\confidence^\alea(i, c)$ of $f^*_{\theta}$ in predicting label $c$ for node $i$ is defined as:
    \begin{equation}
         \mathbb{E}_{p({{\theta}} \mid \matr A, \matr X, \vect y_{{-i}}^\gt)} \left[ \prob{\vect y_i = c \mid \matr A, \matr X, \vect y_{-i}=\vect y_{-i}^\gt, {{\theta}}} \right]
    \end{equation}
\end{definition}
\vspace{-5pt}
Here, we denote with $\smash{\vect y_{-i} = \vect y_{-i}^{\gt}}$ that all nodes excluding the predicted node $i$ are observed as their true values. All remaining lack of confidence is deemed irreducible. Lastly, we define epistemic confidence by comparing aleatoric factors to the overall confidence. As both are defined probabilistically, we consider their ratio:
\begin{definition}\label{def:conf_epi}
    The epistemic confidence of $f^*_{\theta}$ in predicting label $c$ for node $i$ is defined as:
    \begin{equation}
        \confidence^\epi(i, c) := \confidence^\total(i, c) / \confidence^\alea(i, c)
        \label{eq:epi_as_ratio}
    \end{equation}
\end{definition}

These definitions directly imply a notion of uncertainty: $\smash{\uncertainty^\epi(i,c) = \uncertainty^\total(i,c) / \uncertainty^\alea(i,c)}$. Epistemic US thus labels node $i$ when the associated total uncertainty $\smash{\uncertainty^\total(i, \vect y_i^\gt)}$ is large compared to its aleatoric uncertainty $\smash{\uncertainty^\alea(i, \vect y_i^\gt)}$. It favors uncertain nodes \textit{when the uncertainty stems from non-aleatoric sources}. 

\begin{remark}
{In most applications, including US, monotonoic transformations of an uncertainty estimator do not affect its behaviour. Therefore, uncertainty can equivalently be defined as a difference between log-likelihoods instead of likelihood ratios. This definition recovers the well-established additive nature of uncertainty:
$
    \log \uncertainty^\total(i, \vect y_i^\gt) - \log \uncertainty^\alea(i, \vect y_i^\gt) = \log \uncertainty^\epi(i, \vect y_i^\gt)
$. The same holds for logarithmic confidence definitions. Notably, the core result of our work, which we state next, also holds when defining uncertainty in terms of log-likelihoods.}
\end{remark}

\begin{theorem} \label{lem:acquisition}
    Epistemic uncertainty $\smash{\uncertainty^\epi(i, \vect y_i^\gt)}$ of a node $i$ is equivalent to the relative gain its acquisition provides to the posterior over the remaining true labels:
\begin{equation*}
    \uncertainty^\epi(i, \vect y_i^\gt)=\frac{\prob{
        \vect y_{\unlabeled - i} = \vect y_{\unlabeled - i}^\gt \mid \matr A, \matr X, \vect y_\labeled, \vect y_i= \vect y_i^\gt
    }}{\prob{
        \vect y_{\unlabeled - i} = \vect y_{\unlabeled - i}^\gt \mid \matr A, \matr X, \vect y_\labeled
    }}
    \label{eq:acquisition}
\end{equation*}
Hence, acquiring the most epistemically uncertain node is an optimal AL strategy for $f^*_{\theta}$.
\end{theorem}

We provide a proof of \Cref{lem:acquisition} in \Cref{appendix:proofs}. Here, we refer to all unobserved labels excluding $\smash{\vect y_i}$ as $\smash{\vect y_{\unlabeled - i}}$. The ratio that is optimized by epistemic US corresponds to the relative increase in the posterior of the true unobserved labels. That is, it compares the probability the classifier $f^*_{\theta}$ assigns to the remaining unobserved true labels after acquiring the ground-truth label 
$\smash{\vect y_i^\gt}$ of node $i$ as opposed to not acquiring its label. High values indicate that the underlying classifier will be significantly more likely to predict the true labels of the remaining nodes after the corresponding query. 
Thus, a query that maximizes epistemic uncertainty will push the classifier toward predicting the true labels for all remaining unlabeled nodes. This holds for any Bayesian classifier that specifies a posterior $p(\theta \mid \matr A, \matr X, \vect y_\labeled)$ over the parameters of the generative process. For example, fitting the parameters of a GNN is an instance of this framework, where all probability mass is put on one estimate of $\theta$. Approaches like Bayesian GNNs explicitly specify this posterior distribution. However, computing exact disentangled uncertainty requires access to unavailable labels $\vect y_\unlabeled$ and is therefore impractical. Our analysis motivates the development of tractable approximations to these quantities. Novel US approaches can directly benefit from the theoretical optimality guarantees that this work provides. %

\section{Uncertainty Sampling with Ground-Truth Uncertainty}

To support our theoretical claims, we employ US using the proposed ground-truth epistemic uncertainty as an acquisition function. Since for real-world datasets, the data generating process is not known, we first focus our analysis on CSBMs defined in \Cref{sec:background} and discuss a practical approximation later in \Cref{sec:gnd_real}. This allows us to compute the uncertainty estimates of \Cref{def:conf_alea,def:conf_epi,def:conf_total} directly by evaluating the explicit joint likelihood of the generative process $\smash{p(\matr A, \matr X, \vect y)}$ (see \Cref{appendix:computing_alea_and_total_on_csbms}). While the optimality of epistemic US holds for any data-generating process, we focus on CSBMs as they have been extensively studied as proxies for real data in node classification \citep{Palowitch_2022}. To isolate the effect of correctly disentangling uncertainty, we also assume the parameters of the underlying CSBM to be known to the Bayesian classifier $f^*_\theta$. Therefore, any discrepancies in US performance are purely linked to the disentanglement into aleatoric and epistemic factors.

\begin{mybox}
    \vspace{-.1cm}
    \textit{Is US effective in practice?}
    
    \textbf{Yes}, we observe a significant improvement over random acquisition using the proposed ground-truth uncertainty. It is crucial to \textbf{disentangle} uncertainty into aleatoric and epistemic factors.
    \vspace{-.15cm}
\end{mybox}

 We compare the performance of US using the proposed uncertainty measures to contemporary uncertainty estimators over 5 graphs with $100$ nodes and $7$ classes sampled from a CSBM distribution $p(\matr A, \matr X, \vect y)$ in \Cref{fig:al_csbm}. We report similar findings for larger graphs in \Cref{appendix:us_with_gnd}.
 
 \underline{Observations}: 
In agreement with \Cref{lem:acquisition}, epistemic uncertainty significantly outperforms random queries as well as aleatoric and total uncertainty which we explain formally in the following \Cref{prop:aleatoric_uncertainty,prop:total_uncertainty}. We continue to analyze which aspects of uncertainty modelling are crucial for its successful in AL.

\textbf{Disentangling Uncertainty}. 
We first discuss why acquiring nodes with high total uncertainty $\smash{\uncertainty^\total}$ performs worse than isolating epistemic factors: Total uncertainty favors not only informative queries but also tends to acquire labels of nodes that are associated with a high aleatoric uncertainty $\smash{\uncertainty^\alea}$.

\begin{proposition}
    \label{prop:total_uncertainty}
    Total uncertainty $\smash{\uncertainty^\total(i, \vect y_i^\gt)}$ of a node $i$ is proportional to the posterior over the unobserved true labels $\smash{\vect y_{\unlabeled - i}^\gt}$ after acquiring its label $\smash{\vect y_i^\gt}$: 
\begin{equation*}
      \uncertainty^\total(i, \vect y_i^\gt) \propto \prob{\vect y_{\unlabeled -i} =  \vect y_{\unlabeled-i}^\gt \mid \matr A, \matr X, \vect y_\labeled, \vect y_i= \vect y_i^\gt}
\end{equation*}
\end{proposition}

We provide a proof of \Cref{prop:total_uncertainty} in \Cref{appendix:proofs}. Acquiring nodes with maximal total uncertainty maximizes the posterior of the remaining unlabeled set $\smash{\vect y_{\unlabeled 
- i}}$. This is problematic as one way to increase this posterior probability is to remove an aleatorically uncertain node $i$ from the unlabeled set. Such a query will not push the posterior of the remaining nodes in $\smash{\vect y_\unlabeled}$ towards their true labels and instead improve the posterior by excluding nodes that are inherently difficult to predict. In contrast, the epistemic acquisition evaluates the joint posterior in relation to the effect of removing node $i$ from the unlabeled set (see \Cref{lem:acquisition}). In fact, acquiring aleatorically uncertain nodes directly removes inherently ambiguous nodes from the unlabeled set.
\begin{proposition}
    \label{prop:aleatoric_uncertainty}
    Aleatoric uncertainty $\smash{\uncertainty^\alea(i, \vect y_i^\gt)}$ of a node $i$ is proportional to the posterior over the unobserved true labels $\vect y_{\unlabeled - i}^\gt$ \emph{without} acquiring its label $\vect y_i^\gt$: 
\begin{equation*}
   \uncertainty^\alea(i, \vect y_i^\gt) \propto \prob{\vect y_{\unlabeled -i} =  \vect y_{\unlabeled-i}^\gt \mid \matr A, \matr X, \vect y_\labeled}
\end{equation*}
\end{proposition}
We provide a proof for \Cref{prop:aleatoric_uncertainty} in \Cref{appendix:proofs}. \Cref{prop:aleatoric_uncertainty} explains why we observe aleatoric US to be ineffective in \Cref{fig:al_csbm}. It optimizes the posterior of the remaining labels $\smash{\vect y_{\unlabeled - i}^\gt}$ without considering the acquisition of $\smash{\vect y_i^\gt}$. Such queries do not align with AL as they neglect the additional information obtained in each iteration. To optimize predictions on all remaining nodes it is crucial to properly disentangle uncertainty into aleatoric and epistemic components and acquire epistemically uncertain labels.

\looseness=-1
\textbf{Modelling the Data Generating Process}. We also investigate the importance of the uncertainty estimator to faithfully model the true data-generating process $p(\matr A, \matr X, \vect y)$. To that end, we ablate our proposed uncertainty measures but only consider a Bayesian classifier that erroneously exclusively models present edges $(i, j) \in \mathcal{E}$ while neglecting $(i, j) \notin \mathcal{E}$: $\hat{p}(\matr A, \matr X, \vect y) := \allowbreak \prod_{i < j, (i, j) \in {\mathcal{E}}} p(\matr A_{i, j} \mid \vect y_i, \vect y_j) \prod_i p(\matr X_i) \prod p(\vect y_i)$.

\looseness=-1
We specifically pick this inaccurate model to ignore non-existing edges because of its strong resemblance to contemporary GNN architectures used at the backbone of uncertainty estimators discussed in \Cref{sec:al_methods}. They rely on variations of the message-passing framework which propagates information exclusively along existing edges $\mathcal{E}$. In \Cref{fig:al_csbm}, we observe that employing disentangled ground-truth uncertainty based on an inaccurate generative process neglecting non-existing edges harms US even with proper uncertainty disentanglement. Thus, our analysis reveals another potential shortcoming of contemporary uncertainty estimators for graphs: They may fail to accurately learn the underlying data-generating process and thus be incapable of assessing uncertainty faithfully.

\begin{figure}[t]
    \centering
\input{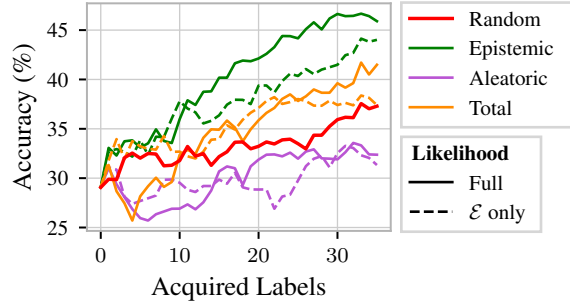}
    \caption{US on a CSBM with 100 nodes and 7 classes using $f^*_\theta$. Ground-truth epistemic uncertainty significantly outperforms other estimators and random queries.}
    \label{fig:al_csbm}
    \vspace{-5mm}
\end{figure}

\textbf{Pitfalls in US for Graphs.} Overall, we identify two aspects required for effective US on graphs that may explain the lackluster performance of existing estimators: Most importantly, uncertainty needs to be \emph{disentangled into aleatoric and epistemic components}. Only epistemic US theoretically and empirically aligns well with AL. Furthermore, we observe an additional performance gain when the Bayesian classifier faithfully \emph{models the data generating process}. Hence, we propose that uncertainty estimators should be designed with special care for isolating epistemic uncertainty while at the same time being expressive enough to capture the unknown underlying data-generating process.

\section{Disentangling Uncertainty on Real Data}\label{sec:gnd_real}

Our theoretical analysis discusses the alignment of AL with epistemic US, employing ground-truth uncertainties. In practice, these are not available, as they require full knowledge of the generative process and access to unavailable labels. We address this gap by proposing a simple approximate disentangled uncertainty estimator that closely follows our analysis. Our results underline the applicability of our theoretical findings to real-world problems.

 In the following, we outline the concepts behind two different paradigms to approximate ground-truth uncertainty. We provide an extensive formal description of both algorithms in \Cref{appendix:gt_approximation}.

\textbf{Multiple Pseudo-Labels (MP).} According to \Cref{eq:epi_as_ratio}, one way to obtain epistemic confidence is to approximate total and aleatoric confidence separately and take their ratio. Total confidence is defined in terms of the unknown underlying data generating process $p(\matr A, \matr X, \vect y)$ (\Cref{def:conf_total}). We approximate it directly from the predictive distribution of the classifier $f_\theta$, thereby assuming $f_\theta$ to implicitly model the data distribution. 

\begin{equation}
    \textcolor{black}{\confidence^\total(i, c) \approx f_\theta(\matr A, \matr X)_{i, c}}
\end{equation}

Aleatoric confidence (\Cref{def:conf_alea}) also conditions the marginal distribution over $\vect y_i$ on unavailable labels $\vect y_{\unlabeled - i}$. Therefore, we use the predictions of $f_\theta$ as pseudo-labels $\hat{\vect y}_i = \argmax_c f_\theta(\matr A, \matr X)_{i, c}$ and temporarily augment the dataset with \emph{multiple pseudo-labels} by setting $\vect y_{\unlabeled - i} = \smash{\hat{\vect y}_{\unlabeled - i}}$. The predictive distribution of an auxiliary classifier $f_{\hat{\theta}_i}$ trained on this dataset approximates aleatoric confidence. 

\begin{equation}
    \textcolor{black}{\confidence^\alea(i, c) \approx f_{\hat{\theta}_i}(\matr A, \matr X)_{i, c}}
\end{equation}

\textcolor{black}{This implies that for each query $\mathcal{O}(n)$ auxiliary classifiers need to be trained on $n-1$ nodes each.} Lastly, \Cref{lem:acquisition} requires us to evaluate epistemic confidence $\smash{\confidence^\epi(i, \vect y_i^\gt)}$ on the unavailable true label $\vect y_i^\gt$ that we substitute with $\hat{\vect y}_i$ as well. \textcolor{black}{We compute: $\smash{\confidence^\epi(i, \vect y_i^\gt) = \confidence^\alea(i, \vect y_i^\gt) / \confidence^\total(i, \vect y_i^\gt)}$.}

\textbf{Expected Single Pseudo-Label (ESP).} The right-hand side of \Cref{lem:acquisition} is equivalent to epistemic uncertainty and can also be approximated as a proxy for acquisition. 
For both the numerator and denominator, we estimate the joint probabilities over $\vect y_{\unlabeled - i}$ as a product of marginal probabilities given by a classifier. The probability of the denominator, 
can be calculated directly from the predictive distribution of $f_\theta$ \textcolor{black}{(see \Cref{appendix:gt_approximation})}. 

\begin{equation}
    \textcolor{black}{\prob{
        \vect y_{\unlabeled - i} = \hat{\vect y}_{\unlabeled - i} \mid \matr A, \matr X, \vect y_\labeled
    } \propto f_\theta(\matr A, \matr X)_{i, \hat{\vect y}_i}^{-1}}
\end{equation}

Similarly to MP, we compute the probability in the numerator, $\smash{\prob{
        \vect y_{\unlabeled - i} = \vect y_{\unlabeled - i}^\gt \mid \matr A, \matr X, \vect y_\labeled, \vect y_i= \vect y_i^\gt
    }}$, by training a separate classifier after augmenting the training data with a label for $\vect y_i$. Since for each node $i$, we condition only on one additional label $\vect y_i$ at a time, we do not use the pseudo-label $\hat{\vect y}_i$ but instead take an \emph{expectation} over all realizations $c \in [C]$ with respect to the predictive distribution of $f_\theta$.

\begin{equation}
\begin{split}
    \prob{\vect y_{\unlabeled - i} = \vect y_{\unlabeled - i}^\gt \mid \matr A, \matr X, \vect y_\labeled, \vect y_i = \vect y_i^\gt}
    \approx \\ \mathbb{E}_{c \sim f_\theta(\matr A, \matr X)_{i, :}} \left[\prob{\vect y_{\unlabeled - i} = \vect y_{\unlabeled - i}^\gt \mid \matr A, \matr X, \vect y_\labeled, \vect y_i = c}\right]
    \end{split}
\end{equation}

\textcolor{black}{In contrast to the MP algorithm, ESP trains $\mathcal{O}(n * c)$ auxiliary classifiers in each iteration. However, the auxiliary models used in the MP approximation are trained on $n-1$ pseudo-labels whereas the training sets used in ESP contain $\lvert O\rvert + 1$ nodes. This results in the ESP algorithm being faster in practice than its MP counterpart despite its worse runtime complexity.}

\begin{figure}
    \centering
\input{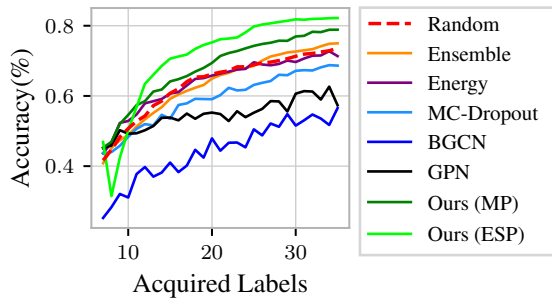}
    \vspace{-1mm}
      \caption{Approximation of disentangled uncertainty against existing epistemic US methods on CoraML. Our framework works well both using multiple pseudo-labels (MP) or taking an expectation over each of them (ESP).}
      \label{fig:ours_cora_ml}
      \vspace{-6mm}
\end{figure}

\begin{table}[h!]
\caption{Average AUC ($\uparrow$) of our proposed approximate ground-truth uncertainty versus existing epistemic US. The best estimator is boldfaced and the runner-up is underlined. Our approximate uncertainty consistently outperforms existing epistemic estimators.}
\label{tab:auc_ours}
\centering
\resizebox{1\columnwidth}{!}{
\begin{tabular}{l|ccccccc}\Xhline{3\arrayrulewidth}
 & Ensemble & \makecell{MC-\\Dropout} & Energy & GPN & BGCN & \makecell{Ours\\(MP)} & \makecell{Ours\\(ESP)}\\\Xhline{3\arrayrulewidth}
CoraML & $63.47$ & $59.17$ & $63.97$ & $54.75$ & $44.45$ & \underline{$68.16$} & \npboldmath{$71.45$}\\
Citeseer & $82.94$ & $78.86$ & $81.59$ & $65.31$ & $58.68$ & \npboldmath{$83.96$} & \underline{$83.43$}\\
Pubmed & \underline{$63.70$} & $58.67$ & $59.64$ & $58.82$ & $55.19$ & \small{n/f} & \npboldmath{$64.36$}\\
Photos & \underline{$84.46$} & $72.42$ & $74.66$ & $54.78$ & $70.83$ & $75.37$ & \npboldmath{$85.52$}\\
Computers & \underline{$68.38$} & $51.02$ & $59.62$ & $39.21$ & $58.64$ & \small{n/f} & \npboldmath{$72.54$}\\
\Xhline{3\arrayrulewidth}
\end{tabular}

}
\vspace{-2
mm}
\end{table}

We visualize the AL curves of epistemic US using these approaches in \Cref{fig:ours_cora_ml} and \Cref{appendix:gt_approximation}. On all datasets, our proposed approximate epistemic uncertainty outperforms other US techniques, often matching the best-performing non-US AL strategy. We observe that ESP approximations perform better in most cases. \textcolor{black}{While both algorithms estimate the same quantity they rely on different approximations. Therefore, it is expected that they behave differently in practice.} While MP substitutes the labels of all unobserved nodes $\unlabeled$ in the dataset with pseudo-labels simultaneously, the ESP method only considers one pseudo-label at a time and factors in the belief of $f_\theta$ by taking an expectation over possible labels. This makes it less prone to erroneous pseudo-labels which, especially for small training sets, are to be expected. 
Both MP and ESP approximations are also exposed to miscalibration of $f_\theta$ and $f_{\hat{\theta}}$, a well-known issue in GNN training \cite{hsu2022makes}. As we discussed in \Cref{sec:gnd_csbm}, message passing GNNs may also not be suited to model the underlying data distribution well. Despite these limitations, approximating and disentangling uncertainty by following our theoretical analysis elevates US from often sub-random to competitive performance levels.

While the performance of our approximate approach to ground-truth uncertainty is surprisingly good and consistently outperforms all existing US methods, our main point is not to propose this as a novel AL strategy.
Instead, these results underline the high applicability of our theoretical findings to the development of uncertainty estimators on graphs. We show that epistemic uncertainty is a provably well-suited acquisition proxy for AL on graphs. It identifies highly informative queries in practice when taking into account the theoretical insights of our work.

\section{Applicability to Indepenent Data}\label{sec:iid_data_discussion}

\textcolor{black}{We define uncertainty in terms of an (unknown) data generating process. Therefore, both the definitions and the theoretical alignment of US with AL could also be applied to i.i.d. data. Our framework is particularly useful for AL on graphs for two reasons: First, a generative perspective enables a theoretically well-motivated and sound approach to instance interdependence. Second, the effectiveness of our proposed approximate estimator in \Cref{sec:gnd_real} hinges on faithfully approximating the generative process. In contrast to other domains, node classification problems can be approached with simple feature transformations when accounting for network effects \citep{wu2019simplifying}. Therefore, the inductive biases of GNNs can ease the approximation of graph generative processes. We empirically verify the need for considering interdependence in \Cref{appendix:gt_features_only}.}

\section{Limitations}
Our work highlights the potential of US for AL on graphs and the role of epistemic uncertainty as an optimal guide.
While we propose an effective off-the-shelf approximation of our theoretical arguments on real data, our goal is not primarily to provide an efficient AL strategy to be deployed in-practice: \textcolor{black}{As our method requires training auxiliary classifiers for each query it shares its runtime complexity of $\mathcal{O}(n * c)$ with the best-performing non-US method, GEEM, and is therefore limited to lightweight models and small datasets.}
Our study serves as a principled exploration into the landscape of US on graphs, aiming to inspire and inform future research in developing uncertainty estimators with consideration for the insights our paper provides. We leave the development of uncertainty estimation strategies that align with our theoretical work to future research.

\section{Conclusion}
Our study sheds light on the potential and challenges of AL on graphs. An extensive benchmark reveals that most strategies only marginally outperform random queries at best, and existing uncertainty estimators inadequately guide US. We introduce ground-truth uncertainty estimates for node classification and prove the alignment of US with AL. An empirical analysis on synthetic and real data shows that epistemic US is a highly effective AL strategy when uncertainty is properly disentangled. We believe this to be a highly relevant result for uncertainty estimation on graphs which, so far, neglected AL as an evaluation criterion. Our work, thus, lays the necessary theoretical groundwork for developing principled uncertainty estimators on graphs. \textcolor{black}{While our analysis can also be transferred to i.i.d. settings, interdependent problems like link prediction or classification may directly benefit from the generative perspective to uncertainty estimation this work introduces.}

\section*{Acknowledgements}

We want to give special thank to Jan Schuchardt for giving helpful suggestions on an early draft of the manuscript. The research presented has been performed in the frame of the RADLEN project funded by TUM Georg Nemetschek Institute Artificial Intelligence for the Built World (GNI). It is further supported by the Bavarian Ministry of Economic Affairs, Regional Development and Energy with funds from the Hightech Agenda Bayern.

\section*{Impact Statement}

Our paper aims at advancing the field of uncertainty estimation on interdependent data which contributes to making AI systems more reliable. By formally establishing a connection to Active Learning, we enable novel acquisition strategies in the domain of graphs to be more data efficient which in turn can help in reducing their negative environmental impact. Nonetheless, advances in the field of Machine Learning are always exposed to non-benign usage. While we acknowledge the various well-established societal and ethical implications of such research we believe the potential positive impacts of our paper outweigh these risks. We encourage both researchers and practitioners building on our analysis to take special care not only for potential misuse but also to thoroughly validate uncertainty estimators that our work proposes, especially in safety-critical domains.

\bibliography{icml2024}

\begin{thebibliography}{70}
\providecommand{\natexlab}[1]{#1}
\providecommand{\url}[1]{\texttt{#1}}
\expandafter\ifx\csname urlstyle\endcsname\relax
  \providecommand{\doi}[1]{doi: #1}\else
  \providecommand{\doi}{doi: \begingroup \urlstyle{rm}\Url}\fi

\bibitem[Ash et~al.(2019)Ash, Zhang, Krishnamurthy, Langford, and Agarwal]{ash2019deep}
Ash, J.~T., Zhang, C., Krishnamurthy, A., Langford, J., and Agarwal, A.
\newblock Deep batch active learning by diverse, uncertain gradient lower bounds.
\newblock \emph{arXiv preprint arXiv:1906.03671}, 2019.

\bibitem[Ash et~al.(2020)Ash, Zhang, Krishnamurthy, Langford, and Agarwal]{ash_deep_2020}
Ash, J.~T., Zhang, C., Krishnamurthy, A., Langford, J., and Agarwal, A.
\newblock Deep {Batch} {Active} {Learning} by {Diverse}, {Uncertain} {Gradient} {Lower} {Bounds}, 2020.

\bibitem[Bandyopadhyay et~al.(2005)Bandyopadhyay, Maulik, Holder, Cook, and Getoor]{bandyopadhyay2005link}
Bandyopadhyay, S., Maulik, U., Holder, L.~B., Cook, D.~J., and Getoor, L.
\newblock Link-based classification.
\newblock \emph{Advanced methods for knowledge discovery from complex data}, pp.\  189--207, 2005.

\bibitem[Beluch et~al.(2018)Beluch, Genewein, Nurnberger, and Kohler]{8579074}
Beluch, W.~H., Genewein, T., Nurnberger, A., and Kohler, J.~M.
\newblock The power of ensembles for active learning in image classification.
\newblock In \emph{2018 IEEE/CVF Conference on Computer Vision and Pattern Recognition}, pp.\  9368--9377, 2018.
\newblock \doi{10.1109/CVPR.2018.00976}.

\bibitem[Berberidis \& Giannakis(2018)Berberidis and Giannakis]{berberidis2018data}
Berberidis, D. and Giannakis, G.~B.
\newblock Data-adaptive active sampling for efficient graph-cognizant classification.
\newblock \emph{IEEE Transactions on Signal Processing}, 66:\penalty0 5167--5179, 2018.

\bibitem[Blundell et~al.(2015)Blundell, Cornebise, Kavukcuoglu, and Wierstra]{blundell2015weight}
Blundell, C., Cornebise, J., Kavukcuoglu, K., and Wierstra, D.
\newblock Weight uncertainty in neural networks, 2015.

\bibitem[Borovitskiy et~al.(2021)Borovitskiy, Azangulov, Terenin, Mostowsky, Deisenroth, and Durrande]{borovitskiy2021gp}
Borovitskiy, V., Azangulov, I., Terenin, A., Mostowsky, P., Deisenroth, M., and Durrande, N.
\newblock Mat{é}rn gaussian processes on graphs.
\newblock In \emph{Proceedings of The 24th International Conference on Artificial Intelligence and Statistics}, volume 130, pp.\  2593--2601. PMLR, 2021.

\bibitem[Cai et~al.(2017)Cai, Zheng, and Chang]{cai2017active}
Cai, H., Zheng, V.~W., and Chang, K. C.-C.
\newblock Active learning for graph embedding, 2017.

\bibitem[Cui et~al.(2022)Cui, Tang, Katariya, Rao, Agrawal, Subbian, and Lee]{cui2022allie}
Cui, L., Tang, X., Katariya, S., Rao, N., Agrawal, P., Subbian, K., and Lee, D.
\newblock Allie: Active learning on large-scale imbalanced graphs.
\newblock In \emph{Proceedings of the ACM Web Conference 2022}. Association for Computing Machinery, 2022.

\bibitem[Dasarathy et~al.(2015)Dasarathy, Nowak, and Zhu]{pmlr-v40-Dasarathy15}
Dasarathy, G., Nowak, R., and Zhu, X.
\newblock S2: An efficient graph based active learning algorithm with application to nonparametric classification.
\newblock In Grünwald, P., Hazan, E., and Kale, S. (eds.), \emph{Proceedings of The 28th Conference on Learning Theory}, volume~40 of \emph{Proceedings of Machine Learning Research}, pp.\  503--522, Paris, France, 03--06 Jul 2015. PMLR.
\newblock URL \url{https://proceedings.mlr.press/v40/Dasarathy15.html}.

\bibitem[Depeweg et~al.(2018)Depeweg, Hernández-Lobato, Doshi-Velez, and Udluft]{depeweg2018decomposition}
Depeweg, S., Hernández-Lobato, J.~M., Doshi-Velez, F., and Udluft, S.
\newblock Decomposition of uncertainty in bayesian deep learning for efficient and risk-sensitive learning, 2018.

\bibitem[Deshpande et~al.(2018)Deshpande, Montanari, Mossel, and Sen]{Deshpande2018ContextualSB}
Deshpande, Y., Montanari, A., Mossel, E., and Sen, S.
\newblock Contextual stochastic block models.
\newblock In \emph{Neural Information Processing Systems}, 2018.

\bibitem[Gal \& Ghahramani(2016)Gal and Ghahramani]{gal2016dropout}
Gal, Y. and Ghahramani, Z.
\newblock Dropout as a bayesian approximation: Representing model uncertainty in deep learning, 2016.

\bibitem[Gal et~al.(2017)Gal, Islam, and Ghahramani]{DBLP:journals/corr/GalIG17}
Gal, Y., Islam, R., and Ghahramani, Z.
\newblock Deep bayesian active learning with image data.
\newblock \emph{CoRR}, abs/1703.02910, 2017.

\bibitem[Gao et~al.(2018)Gao, Yang, Zhou, Wu, Pan, and Hu]{ijcai2018p296}
Gao, L., Yang, H., Zhou, C., Wu, J., Pan, S., and Hu, Y.
\newblock Active discriminative network representation learning.
\newblock In \emph{Proceedings of the Twenty-Seventh International Joint Conference on Artificial Intelligence, {IJCAI-18}}, pp.\  2142--2148. International Joint Conferences on Artificial Intelligence Organization, 2018.
\newblock \doi{10.24963/ijcai.2018/296}.

\bibitem[Gawlikowski et~al.(2023)Gawlikowski, Tassi, Ali, Lee, Humt, Feng, Kruspe, Triebel, Jung, Roscher, et~al.]{gawlikowski2023survey}
Gawlikowski, J., Tassi, C. R.~N., Ali, M., Lee, J., Humt, M., Feng, J., Kruspe, A., Triebel, R., Jung, P., Roscher, R., et~al.
\newblock A survey of uncertainty in deep neural networks.
\newblock \emph{Artificial Intelligence Review}, pp.\  1--77, 2023.

\bibitem[Giles et~al.(1998)Giles, Bollacker, and Lawrence]{giles1998citeseer}
Giles, C.~L., Bollacker, K.~D., and Lawrence, S.
\newblock Citeseer: An automatic citation indexing system.
\newblock In \emph{Proceedings of the third ACM conference on Digital libraries}, pp.\  89--98, 1998.

\bibitem[Gosch et~al.(2023)Gosch, Sturm, Geisler, and G\"unnemann]{gosch2023revisiting}
Gosch, L., Sturm, D., Geisler, S., and G\"unnemann, S.
\newblock Revisiting robustness in graph machine learning, 2023.

\bibitem[Grathwohl et~al.(2019)Grathwohl, Wang, Jacobsen, Duvenaud, Norouzi, and Swersky]{grathwohl2019your}
Grathwohl, W., Wang, K.-C., Jacobsen, J.-H., Duvenaud, D., Norouzi, M., and Swersky, K.
\newblock Your classifier is secretly an energy based model and you should treat it like one.
\newblock \emph{arXiv preprint arXiv:1912.03263}, 2019.

\bibitem[Houlsby et~al.(2011)Houlsby, HuszÃ¡r, Ghahramani, and Lengyel]{houlsby_bayesian_2011}
Houlsby, N., HuszÃ¡r, F., Ghahramani, Z., and Lengyel, M.
\newblock Bayesian {Active} {Learning} for {Classification} and {Preference} {Learning}, 2011.

\bibitem[Hsu et~al.(2022)Hsu, Shen, Tomani, and Cremers]{hsu2022makes}
Hsu, H. H.-H., Shen, Y., Tomani, C., and Cremers, D.
\newblock What makes graph neural networks miscalibrated?
\newblock \emph{Advances in Neural Information Processing Systems}, 35:\penalty0 13775--13786, 2022.

\bibitem[Hu et~al.(2020)Hu, Xiong, Qu, Yuan, Côté, Liu, and Tang]{hu2020graph}
Hu, S., Xiong, Z., Qu, M., Yuan, X., Côté, M.-A., Liu, Z., and Tang, J.
\newblock Graph policy network for transferable active learning on graphs, 2020.

\bibitem[Jaakkola(2000)]{jaakkola2000tutorial}
Jaakkola, T.~S.
\newblock Tutorial on variational approximation methods.
\newblock \emph{Advanced mean field methods: theory and practice}, pp.\  129--159, 2000.

\bibitem[Ji \& Han(2012)Ji and Han]{ji2012variance}
Ji, M. and Han, J.
\newblock A variance minimization criterion to active learning on graphs.
\newblock In \emph{Artificial Intelligence and Statistics}, pp.\  556--564. PMLR, 2012.

\bibitem[Joshi et~al.(2009)Joshi, Porikli, and Papanikolopoulos]{Joshi2009MulticlassAL}
Joshi, A.~J., Porikli, F.~M., and Papanikolopoulos, N.
\newblock Multi-class active learning for image classification.
\newblock \emph{2009 IEEE Conference on Computer Vision and Pattern Recognition}, pp.\  2372--2379, 2009.

\bibitem[Jun \& Nowak(2016)Jun and Nowak]{jun2016graph}
Jun, K.-S. and Nowak, R.
\newblock Graph-based active learning: A new look at expected error minimization.
\newblock In \emph{2016 IEEE Global Conference on Signal and Information Processing (GlobalSIP)}, pp.\  1325--1329. IEEE, 2016.

\bibitem[Kendall \& Gal(2017)Kendall and Gal]{kendall2017uncertainties}
Kendall, A. and Gal, Y.
\newblock What uncertainties do we need in bayesian deep learning for computer vision?, 2017.

\bibitem[Kingma \& Ba(2014)Kingma and Ba]{kingma2014adam}
Kingma, D.~P. and Ba, J.
\newblock Adam: A method for stochastic optimization.
\newblock \emph{arXiv preprint arXiv:1412.6980}, 2014.

\bibitem[Kipf \& Welling(2017)Kipf and Welling]{kipf2017semisupervised}
Kipf, T.~N. and Welling, M.
\newblock Semi-supervised classification with graph convolutional networks, 2017.

\bibitem[Kirsch et~al.(2019)Kirsch, van Amersfoort, and Gal]{kirsch2019batchbald}
Kirsch, A., van Amersfoort, J., and Gal, Y.
\newblock Batchbald: Efficient and diverse batch acquisition for deep bayesian active learning, 2019.

\bibitem[Kiureghian \& Ditlevsen(2009)Kiureghian and Ditlevsen]{KIUREGHIAN2009105}
Kiureghian, A.~D. and Ditlevsen, O.
\newblock Aleatory or epistemic? does it matter?
\newblock \emph{Structural Safety}, 31:\penalty0 105--112, 2009.
\newblock ISSN 0167-4730.
\newblock \doi{https://doi.org/10.1016/j.strusafe.2008.06.020}.

\bibitem[Klicpera et~al.(2018)Klicpera, Bojchevski, and G{\"{u}}nnemann]{DBLP:journals/corr/abs-1810-05997}
Klicpera, J., Bojchevski, A., and G{\"{u}}nnemann, S.
\newblock Personalized embedding propagation: Combining neural networks on graphs with personalized pagerank.
\newblock \emph{CoRR}, abs/1810.05997, 2018.

\bibitem[Lakshminarayanan et~al.(2017)Lakshminarayanan, Pritzel, and Blundell]{lakshminarayanan2017simple}
Lakshminarayanan, B., Pritzel, A., and Blundell, C.
\newblock Simple and scalable predictive uncertainty estimation using deep ensembles, 2017.

\bibitem[Li et~al.(2022)Li, Wu, Rakesh, Lin, Yang, and Wang]{li2022smartquery}
Li, X., Wu, Y., Rakesh, V., Lin, Y., Yang, H., and Wang, F.
\newblock Smartquery: An active learning framework for graph neural networks through hybrid uncertainty reduction.
\newblock In \emph{Proceedings of the 31st ACM International Conference on Information \& Knowledge Management}, pp.\  4199--4203, 2022.

\bibitem[Li et~al.(2020)Li, Yin, and Chen]{li2020seal}
Li, Y., Yin, J., and Chen, L.
\newblock Seal: Semisupervised adversarial active learning on attributed graphs.
\newblock \emph{IEEE Transactions on Neural Networks and Learning Systems}, 32:\penalty0 3136--3147, 2020.

\bibitem[Liu et~al.(2021)Liu, Wang, Owens, and Li]{liu2021energybased}
Liu, W., Wang, X., Owens, J.~D., and Li, Y.
\newblock Energy-based out-of-distribution detection, 2021.

\bibitem[Liu et~al.(2020)Liu, Li, Chen, Hu, and Huang]{liu2020uncertainty}
Liu, Z.-Y., Li, S.-Y., Chen, S., Hu, Y., and Huang, S.-J.
\newblock Uncertainty aware graph gaussian process for semi-supervised learning.
\newblock 34, 2020.

\bibitem[Ma et~al.(2013)Ma, Garnett, and Schneider]{ma2013sigma}
Ma, Y., Garnett, R., and Schneider, J.
\newblock $\sigma$-optimality for active learning on gaussian random fields.
\newblock \emph{Advances in Neural Information Processing Systems}, 26, 2013.

\bibitem[Madhawa \& Murata(2020)Madhawa and Murata]{Madhawa2020ActiveLF}
Madhawa, K. and Murata, T.
\newblock Active learning for node classification: An evaluation.
\newblock \emph{Entropy}, 22, 2020.

\bibitem[Mariadassou et~al.(2010)Mariadassou, Robin, and Vacher]{mariadassou2010uncovering}
Mariadassou, M., Robin, S., and Vacher, C.
\newblock Uncovering latent structure in valued graphs: a variational approach.
\newblock 2010.

\bibitem[McAuley et~al.(2015)McAuley, Targett, Shi, and Van Den~Hengel]{mcauley2015image}
McAuley, J., Targett, C., Shi, Q., and Van Den~Hengel, A.
\newblock Image-based recommendations on styles and substitutes.
\newblock In \emph{Proceedings of the 38th international ACM SIGIR conference on research and development in information retrieval}, pp.\  43--52, 2015.

\bibitem[McCallum et~al.(2000)McCallum, Nigam, Rennie, and Seymore]{mccallum2000automating}
McCallum, A.~K., Nigam, K., Rennie, J., and Seymore, K.
\newblock Automating the construction of internet portals with machine learning.
\newblock \emph{Information Retrieval}, 3:\penalty0 127--163, 2000.

\bibitem[Namata et~al.(2012)Namata, London, Getoor, Huang, and Edu]{namata2012query}
Namata, G., London, B., Getoor, L., Huang, B., and Edu, U.
\newblock Query-driven active surveying for collective classification.
\newblock In \emph{10th international workshop on mining and learning with graphs}, volume~8, pp.\ ~1, 2012.

\bibitem[Nguyen et~al.(2019)Nguyen, Destercke, and HÃ¼llermeier]{nguyen_epistemic_2019}
Nguyen, V.-L., Destercke, S., and HÃ¼llermeier, E.
\newblock Epistemic {Uncertainty} {Sampling}, 2019.

\bibitem[Nguyen et~al.(2022)Nguyen, Shaker, and Hüllermeier]{nguyen2022howto}
Nguyen, V.-L., Shaker, M., and Hüllermeier, E.
\newblock How to measure uncertainty in uncertainty sampling for active learning.
\newblock \emph{Machine Learning}, 111, 2022.
\newblock \doi{10.1007/s10994-021-06003-9}.

\bibitem[Palowitch et~al.(2022)Palowitch, Tsitsulin, Mayer, and Perozzi]{Palowitch_2022}
Palowitch, J., Tsitsulin, A., Mayer, B., and Perozzi, B.
\newblock {GraphWorld}.
\newblock In \emph{Proceedings of the 28th {ACM} {SIGKDD} Conference on Knowledge Discovery and Data Mining}. {ACM}, 2022.
\newblock \doi{10.1145/3534678.3539203}.

\bibitem[Regol et~al.(2020)Regol, Pal, Zhang, and Coates]{regol2020active}
Regol, F., Pal, S., Zhang, Y., and Coates, M.
\newblock Active learning on attributed graphs via graph cognizant logistic regression and preemptive query generation.
\newblock In \emph{International Conference on Machine Learning}, pp.\  8041--8050. PMLR, 2020.

\bibitem[Ren et~al.(2021)Ren, Xiao, Chang, Huang, Li, Gupta, Chen, and Wang]{ren2021survey}
Ren, P., Xiao, Y., Chang, X., Huang, P.-Y., Li, Z., Gupta, B.~B., Chen, X., and Wang, X.
\newblock A survey of deep active learning, 2021.

\bibitem[Rong et~al.(2020)Rong, Huang, Xu, and Huang]{Rong2020DropEdge}
Rong, Y., Huang, W., Xu, T., and Huang, J.
\newblock Dropedge: Towards deep graph convolutional networks on node classification.
\newblock In \emph{International Conference on Learning Representations}, 2020.

\bibitem[Schmidt \& Günnemann(2023)Schmidt and Günnemann]{schmidt2023streambased}
Schmidt, S. and Günnemann, S.
\newblock Stream-based active learning by exploiting temporal properties in perception with temporal predicted loss, 2023.

\bibitem[Sen et~al.(2008)Sen, Namata, Bilgic, Getoor, Galligher, and Eliassi-Rad]{sen2008collective}
Sen, P., Namata, G., Bilgic, M., Getoor, L., Galligher, B., and Eliassi-Rad, T.
\newblock Collective classification in network data.
\newblock \emph{AI magazine}, 29:\penalty0 93--93, 2008.

\bibitem[Sener \& Savarese(2018)Sener and Savarese]{sener2018active}
Sener, O. and Savarese, S.
\newblock Active learning for convolutional neural networks: A core-set approach, 2018.

\bibitem[Shannon(1948)]{shannon_mathematical_1948}
Shannon, C.~E.
\newblock A mathematical theory of communication, 1948.

\bibitem[Sharma \& Bilgic(2016)Sharma and Bilgic]{Sharma2016EvidencebasedUS}
Sharma, M. and Bilgic, M.
\newblock Evidence-based uncertainty sampling for active learning.
\newblock \emph{Data Mining and Knowledge Discovery}, 31:\penalty0 164--202, 2016.

\bibitem[Shui et~al.(2020)Shui, Zhou, C~Gagné, and Wang]{shui_deep_2020}
Shui, C., Zhou, F., C~Gagné, C., and Wang, B.
\newblock Deep {Active} {Learning}: {Unified} and {Principled} {Method} for {Query} and {Training}, 2020.

\bibitem[Sinha et~al.(2019)Sinha, Ebrahimi, and Darrell]{sinha_variational_2019}
Sinha, S., Ebrahimi, S., and Darrell, T.
\newblock Variational {Adversarial} {Active} {Learning}, 2019.

\bibitem[Stadler et~al.(2021)Stadler, Charpentier, Geisler, Z{\"u}gner, and G{\"u}nnemann]{stadler2021graph}
Stadler, M., Charpentier, B., Geisler, S., Z{\"u}gner, D., and G{\"u}nnemann, S.
\newblock Graph posterior network: Bayesian predictive uncertainty for node classification.
\newblock \emph{Advances in Neural Information Processing Systems}, 34:\penalty0 18033--18048, 2021.

\bibitem[Sverchkov \& Craven(2017)Sverchkov and Craven]{sverchkov2017review}
Sverchkov, Y. and Craven, M.
\newblock A review of active learning approaches to experimental design for uncovering biological networks.
\newblock \emph{PLOS Computational Biology}, 13, 2017.
\newblock \doi{10.1371/journal.pcbi.1005466}.

\bibitem[Wang \& Shang(2014)Wang and Shang]{wang_new_2014}
Wang, D. and Shang, Y.
\newblock A new active labeling method for deep learning.
\newblock In \emph{2014 {International} {Joint} {Conference} on {Neural} {Networks} ({IJCNN})}. IEEE, 2014.
\newblock \doi{10.1109/IJCNN.2014.6889457}.

\bibitem[Wollschläger et~al.(2023)Wollschläger, Gao, Charpentier, Ketata, and Günnemann]{wollschlaeger2023uncertainty}
Wollschläger, T., Gao, N., Charpentier, B., Ketata, M.~A., and Günnemann, S.
\newblock Uncertainty estimation for molecules: Desiderata and methods, 2023.

\bibitem[Wu et~al.(2019)Wu, Souza, Zhang, Fifty, Yu, and Weinberger]{wu2019simplifying}
Wu, F., Souza, A., Zhang, T., Fifty, C., Yu, T., and Weinberger, K.
\newblock Simplifying graph convolutional networks.
\newblock In \emph{International conference on machine learning}, pp.\  6861--6871. PMLR, 2019.

\bibitem[Wu et~al.(2023)Wu, Chen, Yang, and Yan]{wu2023energybased}
Wu, Q., Chen, Y., Yang, C., and Yan, J.
\newblock Energy-based out-of-distribution detection for graph neural networks, 2023.

\bibitem[Wu et~al.(2021)Wu, Xu, Singh, Yang, and Dubrawski]{wu2021active}
Wu, Y., Xu, Y., Singh, A., Yang, Y., and Dubrawski, A.
\newblock Active learning for graph neural networks via node feature propagation, 2021.

\bibitem[Yin et~al.(2017)Yin, Qian, Cao, Li, Wei, Zheng, and Davidson]{yin_deep_2017}
Yin, C., Qian, B., Cao, S., Li, X., Wei, J., Zheng, Q., and Davidson, I.
\newblock Deep {Similarity}-{Based} {Batch} {Mode} {Active} {Learning} with {Exploration}-{Exploitation}.
\newblock In \emph{2017 {IEEE} {International} {Conference} on {Data} {Mining} ({ICDM})}, 2017.
\newblock \doi{10.1109/ICDM.2017.67}.

\bibitem[Yoo \& Kweon(2019)Yoo and Kweon]{Yoo_2019_CVPR}
Yoo, D. and Kweon, I.~S.
\newblock Learning loss for active learning.
\newblock In \emph{Proceedings of the IEEE/CVF Conference on Computer Vision and Pattern Recognition (CVPR)}, June 2019.

\bibitem[Zhan et~al.(2022)Zhan, Wang, Huang, Xiong, Dou, and Chan]{zhan_comparative_2022}
Zhan, X., Wang, Q., Huang, K.-h., Xiong, H., Dou, D., and Chan, A.~B.
\newblock A {Comparative} {Survey} of {Deep} {Active} {Learning}, 2022.

\bibitem[Zhang et~al.(2022{\natexlab{a}})Zhang, Katz-Samuels, and Nowak]{pmlr-v162-zhang22k}
Zhang, J., Katz-Samuels, J., and Nowak, R.
\newblock {GALAXY}: Graph-based active learning at the extreme.
\newblock In Chaudhuri, K., Jegelka, S., Song, L., Szepesvari, C., Niu, G., and Sabato, S. (eds.), \emph{Proceedings of the 39th International Conference on Machine Learning}, volume 162 of \emph{Proceedings of Machine Learning Research}, pp.\  26223--26238. PMLR, 17--23 Jul 2022{\natexlab{a}}.
\newblock URL \url{https://proceedings.mlr.press/v162/zhang22k.html}.

\bibitem[Zhang et~al.(2022{\natexlab{b}})Zhang, Wang, You, Cao, Huang, Shan, Yang, and Cui]{zhang2022information}
Zhang, W., Wang, Y., You, Z., Cao, M., Huang, P., Shan, J., Yang, Z., and Cui, B.
\newblock Information gain propagation: a new way to graph active learning with soft labels, 2022{\natexlab{b}}.

\bibitem[Zhang et~al.(2022{\natexlab{c}})Zhang, Wang, You, Cao, Huang, Shan, Yang, and Cui]{zhang2022propagation}
Zhang, W., Wang, Y., You, Z., Cao, M., Huang, P., Shan, J., Yang, Z., and Cui, B.
\newblock Information gain propagation: A new way to graph active learning with soft labels.
\newblock In \emph{ICLR}, 2022{\natexlab{c}}.

\bibitem[Zhu et~al.(2003)Zhu, Lafferty, and Ghahramani]{zhu2003combining}
Zhu, X., Lafferty, J., and Ghahramani, Z.
\newblock Combining active learning and semi-supervised learning using gaussian fields and harmonic functions.
\newblock In \emph{ICML 2003 workshop on the continuum from labeled to unlabeled data in machine learning and data mining}, volume~3, 2003.

\end{thebibliography}
\bibliographystyle{icml2024}

\newpage
\appendix
\onecolumn

\clearpage

\section{Proofs}\label{appendix:proofs}

\begin{customthm}{1}
Epistemic uncertainty $\smash{\uncertainty^\epi(i, \vect y_i^\gt)}$ of a node $i$ is equivalent to the relative gain its acquisition provides to the posterior over the remaining true labels:
\begin{equation*}
    \uncertainty^\epi(i, \vect y_i^\gt)=\frac{\prob{
        \vect y_{\unlabeled - i} = \vect y_{\unlabeled - i}^\gt \mid \matr A, \matr X, \vect y_\labeled, \vect y_i= \vect y_i^\gt
    }}{\prob{
        \vect y_{\unlabeled - i} = \vect y_{\unlabeled - i}^\gt \mid \matr A, \matr X, \vect y_\labeled
    }}
\end{equation*}
Hence, acquiring the most epistemically uncertain node is an optimal AL strategy for $f^*_{\theta}$.
\end{customthm}
\begin{proof}
\begin{equation*}\begin{aligned}
    \confidence^\alea(i, \vect y_i^\gt) &={\int \prob{\vect y_i = \vect y_i^\gt \mid \matr A, \matr X, \vect y_{-i} = \vect y_{-i}^{\gt}, \theta} p(\theta \mid \matr A, \matr X, \vect y_{-i} = \vect y_{-i}^{\gt})d\theta} \\
    &= \prob{\vect y_i = \vect y_i^\gt \mid \matr A, \matr X, \vect y_{-i} = \vect y_{-i}^{\gt}}\\
    &= \prob{\vect y_i = \vect y_i^\gt | \matr A, \matr X, \vect y_{\labeled} , \vect y_{\unlabeled - i} = \vect y_{\unlabeled - i}^{\gt}} \\
    &= \frac{\prob{\vect y_{\unlabeled - i} = \vect y_{\unlabeled - i}^{\gt} \mid \matr A, \matr X, \vect y_{\labeled} , \vect y_i = \vect y_i^\gt}}{\prob{\vect y_{\unlabeled -i} = \vect y_{\unlabeled - i}^{\gt} \mid \matr A, \matr X, \vect y_{\labeled} }} \prob{\vect y_i = \vect y_i^\gt \mid \matr X, \matr A, \vect y_{\labeled}} \\
    &= \frac{\prob{\vect y_{\unlabeled - i} = \vect y_{\unlabeled - i}^{\gt} \mid \matr A, \matr X, \vect y_{\labeled} , \vect y_i = \vect y_i^\gt}}{\prob{\vect y_{\unlabeled - i} = \vect y_{\unlabeled -i}^{\gt} | \matr A, \matr X, \vect y_{\labeled} }} \confidence^\total(i, \vect y_i^\gt) \\
    \uncertainty^\epi(i, \vect y_i^\gt) &= \frac{\confidence^\alea(i, \vect y_i^\gt)}{\confidence^\total(i, \vect y_i^\gt)} = \frac{\prob{\vect y_{\unlabeled - i} = \vect y_{\unlabeled - i}^{\gt} \mid \matr A, \matr X, \vect y_{\labeled} , \vect y_i = \vect y_i^\gt}}{\prob{\vect y_{\unlabeled - i} = \vect y_{\unlabeled -i}^{\gt} | \matr A, \matr X, \vect y_{\labeled} }} 
\end{aligned}\end{equation*}

First, we insert our definition of aleatoric confidence and marginalize $\theta$. Then, we split $\vect y_{-i}$ into two parts: observed $\vect y_{\labeled}$ and unobserved $\vect y_{\unlabeled - i}$. As we assign the ground-truth labels to both, the exact partition is not relevant. Next, we use Bayes law to get a distribution over the unobserved node labels $\vect y_{\unlabeled}$. {Similarly to the first step, we marginalize $\theta$ to obtain $\smash{\confidence^\total(i, \vect y_i^\gt)}$.} In the last step, we see that the right term matches our definition of total confidence $\smash{\confidence^{\total}}$.
\end{proof}

\begin{customproposition}{1}
    Total uncertainty $\smash{\uncertainty^\total(i, \vect y_i^\gt)}$ of a node $i$ is proportional to the posterior over the unobserved true labels $\smash{\vect y_{\unlabeled - i}^\gt}$ after acquiring its label $\smash{\vect y_i^\gt}$: 
\begin{equation*}
      \uncertainty^\total(i, \vect y_i^\gt) \propto \prob{\vect y_{\unlabeled -i} =  \vect y_{\unlabeled-i}^\gt \mid \matr A, \matr X, \vect y_\labeled, \vect y_i= \vect y_i^\gt}
\end{equation*}
\end{customproposition}
\begin{proof}
\begin{equation*}\begin{aligned}
    \uncertainty^\total(i, \vect y_i^\gt) &= 
    \frac{1}{\confidence^\total(i, \vect y_i^\gt)} \\ &=\frac{\confidence^\alea(i, \vect y_i^\gt)}{\confidence^\total(i, \vect y_i^\gt)} \frac{1}{\confidence^\alea(i, \vect y_i^\gt)} \\ 
    &= \frac{\prob{\vect y_{\unlabeled} = \vect y_{\unlabeled - i}^{\gt} \mid \dots, \vect y_i = \vect y_i^\gt}}{\prob{\vect y_{\unlabeled - i} = \vect y_{\unlabeled -i}^{\gt} | \dots}}\frac{1}{\confidence^\alea(i, \vect y_i^\gt)} \\ 
     &=\frac{\prob{\vect y_{\unlabeled - i} = \vect y_{\unlabeled - i}^{\gt} \mid \dots, \vect y_i = \vect y_i^\gt}}{\prob{\vect y_{\unlabeled - i} = \vect y_{\unlabeled -i}^{\gt} | \dots}} \frac{1}{\prob{\vect y_i = \vect y_i^\gt \mid \vect y_{\unlabeled_i} = \vect y_{\unlabeled - i}^\gt, \dots}} \\
    &= \frac{\prob{\vect y_{\unlabeled - i} = \vect y_{\unlabeled - i}^{\gt} \mid \dots, \vect y_i = \vect y_i^\gt}}{\prob{\vect y_{\unlabeled - i} = \vect y_{\unlabeled -i}^{\gt} | \dots}} \frac{\prob{\vect y_{\unlabeled - i} = \vect y_{\unlabeled -i}^{\gt} | \dots}}{\prob{\vect y_i = \vect y_i^\gt, \vect y_{\unlabeled - i} = \vect y_{\unlabeled - i}^\gt \mid \vect \dots}} \\
    &= \frac{\prob{\vect y_{\unlabeled - i} = \vect y_{\unlabeled - i}^{\gt} \mid \dots, \vect y_i = \vect y_i^\gt}}{\prob{\vect y_i = \vect y_i^\gt \vect y_{\unlabeled - i} = \vect y_{\unlabeled - i}^\gt \mid \vect \dots}} \\
    &= \frac{\prob{\vect y_{\unlabeled - i} = \vect y_{\unlabeled - i}^{\gt} \mid \dots, \vect y_i = \vect y_i^\gt}}{\prob{ \vect y_{\unlabeled} = \vect y_{\unlabeled}^\gt \mid \vect \dots}} \\
    &\propto {\prob{\vect y_{\unlabeled - i} = \vect y_{\unlabeled - i}^{\gt} \mid \dots, \vect y_i = \vect y_i^\gt}}
\end{aligned}\end{equation*}

Here, we abbreviated $\matr A, \matr X, \vect y_\labeled $ with $\dots$ for clarity. First, we insert the result from \Cref{lem:acquisition}. Then, we use the definition of conditional probability to replace the inverse of the aleatoric uncertainty. After canceling terms, we observe that $\prob{ \vect y_{\unlabeled} = \vect y_{\unlabeled}^\gt \mid \vect \dots}$ is a constant with respect to $i$ and arrive at the desired result that the total uncertainty is proportional to the posterior over the true labels $\prob{\vect y_{\unlabeled - i} = \vect y_{\unlabeled -i}^{\gt} | \dots, \vect y_i = \vect y_i^\gt}$.{As in \Cref{prop:total_uncertainty}, we implicitly marginalized the learnable parameters $\theta$ of the Bayesian classifier.}

\end{proof}

\begin{customproposition}{2}
    Aleatoric uncertainty $\smash{\uncertainty^\alea(i, \vect y_i^\gt)}$ of a node $i$ is proportional to the posterior over the unobserved true labels $\vect y_{\unlabeled - i}^\gt$ \emph{without} acquiring its label $\vect y_i^\gt$: 
\begin{equation*}
   \uncertainty^\alea(i, \vect y_i^\gt) \propto \prob{\vect y_{\unlabeled -i} =  \vect y_{\unlabeled-i}^\gt \mid \matr A, \matr X, \vect y_\labeled}
\end{equation*}
\end{customproposition}
\begin{proof} 
 \begin{equation*}\begin{aligned}
    \uncertainty^\alea(i, \vect y_i^\gt) &= \frac{1}{\confidence^\alea(i, \vect y_i^\gt)} \\ &= \frac{\confidence^\total(i, \vect y_i^\gt)}{\confidence^\alea(i, \vect y_i^\gt)}\frac{1}{\confidence^\total(i, \vect y_i^\gt)} \\
    &= \frac{\prob{\vect y_{\unlabeled - i} = \vect y_{\unlabeled -i}^{\gt} | \dots}}{\prob{\vect y_{\unlabeled} = \vect y_{\unlabeled - i}^{\gt} \mid \dots, \vect y_i = \vect y_i^\gt}}\frac{1}{\confidence^\total(i, \vect y_i^\gt)} \\
    &= \frac{\prob{\vect y_{\unlabeled - i} = \vect y_{\unlabeled -i}^{\gt} | \dots}}{\prob{\vect y_{\unlabeled} = \vect y_{\unlabeled - i}^{\gt} \mid \dots, \vect y_i = \vect y_i^\gt}}\frac{1}{\prob{\vect y_i = \vect y_i^{\gt} \mid \dots}} \\
    &= \frac{\prob{\vect y_{\unlabeled - i} = \vect y_{\unlabeled -i}^{\gt} | \dots}}{\prob{\vect y_{\unlabeled -i } = \vect y_{\unlabeled - i}^{\gt}, \vect y_i = \vect y_i^\gt \mid \dots}} \\
    &= \frac{\prob{\vect y_{\unlabeled - i} = \vect y_{\unlabeled -i}^{\gt} | \dots}}{\prob{\vect y_{\unlabeled} = \vect y_{\unlabeled}^{\gt} \mid \dots}} \\
    & \propto \prob{\vect y_{\unlabeled - i} = \vect y_{\unlabeled -i}^{\gt} | \dots}
\end{aligned}\end{equation*}

Again, we abbreviated $\matr A, \matr X, \vect y_\labeled $ with $\dots$ for clarity. First, we insert the result from \Cref{lem:acquisition} into $\smash{{\confidence^\total(i, \vect y_i^\gt)} / {\confidence^\alea(i, \vect y_i^\gt)}}$. Then, insert the definition of total confidence. We apply the law of conditional probability, we see that $\prob{ \vect y_{\unlabeled} = \vect y_{\unlabeled}^\gt \mid \vect \dots}$ is a constant with respect to $i$ and arrive at the desired result that the aleatoric uncertainty is proportional to the posterior over the true labels $\prob{\vect y_{\unlabeled - i} = \vect y_{\unlabeled -i}^{\gt} | \dots}$. 

\end{proof}

\section{Experimental Setup}\label{appendix:experimental_setup}
\subsection{Active Learning} We discuss AL on graphs, where given an initial set of labeled instances $\set O \subset \set V$ we aim to acquire a set of the unlabeled nodes $\set U \subset \set V$ that is optimal in terms of improving the performance of the classifier on the entire graph. That is, we \begin{inparaenum}[(1)]
     \item initially label one node randomly drawn from each class.
     \item We then re-initialize the model weights and train the classifier until convergence.
     \item We employ an acquisition strategy to select one or more unlabeled node(s).
     \item We add the acquired label(s) to the training set and repeat the procedure at step (2) until some acquisition budget is exhausted.
\end{inparaenum}
In contrast to domains where model training is expensive \citep{8579074, DBLP:journals/corr/GalIG17, kirsch2019batchbald}, we re-train the classifier after each acquisition iteration. If not stated otherwise, we only acquire one node label in each iteration and fix the acquisition budget to $5C$. The resulting final training pools, therefore, contain fewer instances compared to dataset splits commonly used in other work \citep{kipf2017semisupervised,stadler2021graph}.

\subsection{Hyperparameters}\label{appendix:hyperparameters}
As hyperparameter tuning may be unrealistic in AL \citep{regol2020active}, we do not finetune them on a validation set. One potential strategy to realize hyperparameters is to randomly sample them from the search space over which hyperparameter optimization would be performed. This, however, is expected to lead to notably worse performance for most GNN architectures \citep{regol2020active}. Since the goal of our work is to showcase that even under optimal circumstances, contemporary uncertainty estimators can not enable US to be effective, we select one set of hyperparameters reported to be effective by the literature and employ it on all datasets. Specifically, we chose the following values:

\begin{table}[H]
\centering
\resizebox{\textwidth}{!}{
\begin{tabular}{ l | c|c|c|c|c|c|c|c|c }
\textbf{Model} & \makecell{Hidden \\ Dimensions} & \makecell{Learning \\ Rate} & \makecell{Max \\ Epochs} & \makecell{Weight \\ Decay} & \makecell{Teleport \\ Probability} & \makecell{Power \\ Iteration \\ Steps} & Dropout & \makecell{Flow \\ Dimension} & \makecell{Number \\ Radial Flow \\ Layers}\\
\hline
 GCN & [64] & 0.001 & 10,000 & 0.001 & n/a & n/a. & 0.0 & n/a & n/a.\\
 APPNP & [64] & 0.001 & 10,000 & 0.001 & 0.2 & 10 & 0.0 & n/a & n/a\\
 GPN & [64] & 0.001 & 10,000 & 0.001 & 0.2 & 10 & 0.5 & 16 & 10\\    
\end{tabular}

}
\caption{Hyperparameters of different GNN backbones}
\end{table}

\subsection{Datasets}\label{appendix:datasets}

\textbf{Real-World Datasets.} We evaluate AL approaches on three common node classification benchmark citation datasets: CoraML \citep{mccallum2000automating, sen2008collective, bandyopadhyay2005link, giles1998citeseer}, Citeseer \citep{sen2008collective, bandyopadhyay2005link, giles1998citeseer}, PubMed \citep{namata2012query}. 
In all datasets, nodes are papers and edges model citations. We consider undirected edges only and select the largest connected component in the dataset. All datasets use bag-of-words representations and we normalize the input features $\vect x_i$ node-wise to have a $L_2$-norm of $1$. We report the statistics of each dataset in \Cref{tab:data}.

\begin{table}[h!]
\centering
\resizebox{\textwidth}{!}{
\begin{tabular}{l|c|c|c|c|c|c|c|c|c}
    Dataset & 
    \makecell{\textbf{\#Nodes $n$}} & \textbf{\#Edges $m$} & \makecell{\textbf{\#Input} \\ \textbf{Features $d$}} & \textbf{\#Classes $c$} & \makecell{\textbf{Edge} \\ \textbf{ Density} $\frac{m}{n^2}$} & \textbf{Homophily $p$} & \makecell{\textbf{Intra-Class} \\ \textbf{Edge Density $p$}}& \makecell{\textbf{Inter-Class} \\ \textbf{Edge Density $q$}} & \textbf{SNR $\frac{p}{q}$}\\
    \hline
    CoraML & $2,810$ & $15,962$ & $2,879$ & $7$ & $0.20$\% & $78.44$\% & $1.69$\% & $0.06\%$ & $28.51$ \\
    Citeseer & $1,681$ & $5,804$ & $602$ & $6$ & $0.20$\% & $92.76$\% & $4.89$\% & $0.03\%$ & $141.36$ \\
    PubMed & $19,717$ & $88,648$ & $500$ & $3$ & $0.02$\% & $80.24$\% & $0.07$\% & $0.01\%$ & $9.48$ \\
    AmazonComputers & $13,381$ & $491,556$ & $767$ & $10$ & $0.27\%$ & $77.72\%$ & $3.60\%$ & $0.07\%$ & $54.74\%$ \\
    AmazonPhotos & $7,484$ & $238,086$ & $745$ & $8$ &  $42.47\%$ & $82.72\%$ & $3.77\%$ & $0.10\%$ & $36.85\%$
\end{tabular}

}
\caption{Dataset statistics.}
\label{tab:data}
\end{table}

We additionally compute the edge density $\frac{m}{n^2}$ as well as the homophily which is the fraction of edges that link between nodes of the same class. For comparability with CSBMs, we also report the average empirical inter-class edge probabilities $p$ and intra-class edge probabilities $q$ as well as their ratio, the structural signal-to-noise rate (SNR).

\textbf{CSBMs.}

{
CSBMs define the following generative process: First, node labels are sampled independently from a prior $p(\vect y)$. For each node, features are then drawn independently from a class-conditional normal distribution $p(\matr X_i \mid \vect y_i) \sim \mathcal{N}(\vect{\mu}_{\vect y_i}, \sigma_x^2 \matr I)$. Each edge in the graph is generated independently according to an affiliation matrix $\matr F \in \mathbb{R}^{c \times c}$, i.e. $p(\matr A_{i,j} \mid \vect y_i, \vect y_j) \sim \text{Ber}(\matr{F}_{\vect y_i, \vect y_j})$. This gives rise to an explicit joint distribution over the graph that factorizes as $\smash{p(\matr{A}, \matr{X}, \vect y) = \prod_{i < j}p(\matr A_{i,j} \mid \vect y_i, \vect y_j) \prod_{i}p(\matr X_i \mid \vect y_i) \prod_i  p(\vect y_i)}$. 
}

We generate CSBM graphs with a fixed number of nodes, classes, and input features according to \Cref{sec:background}. That is, we first sample class labels $\vect y_i$ for each node independently from a uniform prior $\mathbb{P}\bigl[\vect y_i\bigr] = 1 / C$. We then create links according to the affiliation matrix $\matr F$, where $\mathbb{P}\bigl[\matr A_{i,j} = 1 \vert \vect y_i, \vect y_j \bigr] = \matr F_{\vect y_i, \vect y_j}$. If not stated otherwise, we use a symmetric and homogeneous affiliation matrix $\matr F$, where we set all diagonal elements to a given intra-class edge probability $p$ and all off-diagonal elements to a given inter-class edge probability $q$. We enforce a structural signal-to-noise ratio (SNR) $\sigma_A = p / q$ by specifying an expected node degree $\E\left[\text{deg}(v) \right]$ and then solving for:

\begin{equation}
    q = \frac{\E\left[\text{deg}(v)\right] c}{n - 1} \frac{1}{\sigma_A + c - 1}
\end{equation}

For each class, we first deterministically draw $C$ vectors of dimension $d$ that all have a pairwise distance of $\delta_X$. We then use a random rotation to obtain class centers $\vect \mu_c$. For each node, we then independently sample its features $\vect x_i$ from a normal distribution $\mathcal{N}(\matr X_i \vert \vect\mu_{\vect y_i}, \sigma_X \matr I)$. We refer to the quotient $\frac{\delta_X}{\sigma_X}$ as the feature signal-to-noise ratio. The dimension of the feature space is given by $d = \max(c, \lceil n / \log^2n \rceil)$, following \cite{gosch2023revisiting}.

\subsection{Model Details}

If applicable, we use the same GCN \citep{kipf2017semisupervised} backbone for all models. That is, we employ one hidden layer of dimension of $64$. The APPNP \citep{DBLP:journals/corr/abs-1810-05997} model uses an MLP with one hidden layer of the same dimension. We diffuse the predictions using $10$ steps of power iteration and a teleport probability of $0.2$. 

If not stated otherwise, ensembles are composed of 10 architecturally identical GCNs. In the case of MCD, we apply dropout with probability $0.1$ to each neuron of the GCN backbone independently. The BGCN mimics the GCN but models weights and biases as normal distributions. We follow \citep{blundell2015weight} and regularize these towards a standard normal distribution. Consequentially, we apply a weight of $\lambda = 0.1$ to the regularization loss. 

We follow \cite{stadler2021graph} to configure the GPN model: We use 10 radial flow layers to implement the class-conditional density model with a dimension of 16. For diffusion, we implement the same configuration as for APPNP. Other hyperparameters are set as in \citep{stadler2021graph}. 

{For SEAL, we follow the authors and do not re-train the model after each acquisition \citep{li2020seal}. Furthermore, we pick the number of training iterations for the discriminator to be $n_d = 5$ and the number of iterations for the generator $n_g = 10$, a combination that we observed to be successful on one dataset and kept fixed for all others as the authors do not provide them directly and hyperparameter optimization is unrealistic in AL \citep{regol2020active}.}

\subsection{Training Details}

Apart from the GPN, all of the aforementioned models are trained towards the binary-cross-entropy objective using the ADAM \citep{kingma2014adam} optimizer with a learning rate of $10^{-3}$ and weight decay of $10^{-3}$. We also perform early stopping on the validation loss with a patience of $100$ iterations. We implement our models in PyTorch and PyTorch Geometric and train on two types of machines: 
\begin{inparaenum}[(i)]
    \item Xeon E5-2630 v4 CPU @ 2.20GHz with a NVIDA GTX 1080TI GPU and 128 GB of RAM.
    \item AMD EPYC 7543 CPU @ 2.80GHz with a NVIDA A100 GPU and 128 GB of RAM
\end{inparaenum}. 

For each dataset, classifier, and acquisition function, we report results averaged over five different dataset splits and five distinct model initializations. In each dataset split, we a priori fix 20\% of all nodes as a test set that is reused in every subsequent dataset split and can never be acquired by any strategy.

\section{Additional Metrics and Plots}\label{appendix:additional_metrics_and_plots}

We supply an evaluation of contemporary non-uncertainty-based acquisition strategies on different models as well as various uncertainty estimators for US over all datasets listed in \Cref{appendix:datasets}. We briefly summarize the key insights:

\begin{figure}[ht]
    \centering
    \begin{subfigure}{.49\textwidth}
        \centering
\input{files/pgfs/non_uncertainty/cora_ml_gcn.pgf}
      \caption{AL strategies on CoraML using a GCN.}
      \label{fig:non_uncertainty_cora_ml_gcn}
    \end{subfigure}
    \hfill
    \begin{subfigure}{.49\textwidth}
      \centering
\input{files/pgfs/uncertainty/cora_ml.pgf}
      \caption{US on CoraML.}
      \label{fig:uncertainty_cora_ml}
    \end{subfigure}
    \begin{subfigure}{.49\textwidth}
        \centering
\input{files/pgfs/non_uncertainty/cora_ml_appnp.pgf}
      \caption{AL strategies on CoraML using APPNP.}
      \label{fig:non_uncertainty_cora_ml_appnp}
    \end{subfigure}
    \hfill
    \begin{subfigure}{.49\textwidth}
      \centering
\input{files/pgfs/non_uncertainty/cora_ml_sgc.pgf}
      \caption{AL strategies on CoraML using a SGC.}
      \label{fig:non_uncertainty_cora_ml_sgc}
    \end{subfigure}
    \begin{subfigure}{.49\textwidth}
        \centering
\input{files/pgfs/non_uncertainty/citeseer_gcn.pgf}
      \caption{AL strategies on Citeseer using a GCN.}
      \label{fig:non_uncertainty_citeseer_gcn}
    \end{subfigure}
    \hfill
    \begin{subfigure}{.49\textwidth}
      \centering
\input{files/pgfs/uncertainty/citeseer.pgf}
      \caption{US on Citeseer.}
      \label{fig:uncertainty_citeseer2}
    \end{subfigure}
    \begin{subfigure}{.49\textwidth}
        \centering
\input{files/pgfs/non_uncertainty/citeseer_appnp.pgf}
      \caption{AL strategies on Citeseer using APPNP.}
      \label{fig:non_uncertainty_citeseer_appnp}
    \end{subfigure}
    \hfill
    \begin{subfigure}{.49\textwidth}
      \centering
\input{files/pgfs/non_uncertainty/citeseer_sgc.pgf}
      \caption{AL strategies on Citeseer using a SGC.}
      \label{fig:non_uncertainty_citeseer_sgc}
    \end{subfigure}
    \caption{Accuracy curves of AL strategies, both non-uncertainty-based as well as US on different datasets for different models.}
\end{figure}
\begin{figure}[ht]\ContinuedFloat
    \centering
    \begin{subfigure}{.49\textwidth}
        \centering
\input{files/pgfs/non_uncertainty/pubmed_gcn.pgf}
      \caption{AL strategies on PubMed using a GCN.}
      \label{fig:non_uncertainty_pubmed_gcn}
    \end{subfigure}
    \hfill
    \begin{subfigure}{.49\textwidth}
      \centering
\input{files/pgfs/uncertainty/pubmed.pgf}
      \caption{US on PubMed.}
      \label{fig:uncertainty_pubmed}
    \end{subfigure}
    
    \begin{subfigure}{.49\textwidth}
        \centering
\input{files/pgfs/non_uncertainty/pubmed_appnp.pgf}
      \caption{AL strategies on PubMed using APPNP.}
      \label{fig:non_uncertainty_pubmed_appnp}
    \end{subfigure}
    \hfill
    \begin{subfigure}{.49\textwidth}
      \centering
\input{files/pgfs/non_uncertainty/pubmed_sgc.pgf}
      \caption{AL strategies on PubMed using a SGC.}
      \label{fig:non_uncertainty_pubmed_sgc}
    \end{subfigure}
    \begin{subfigure}{.49\textwidth}
        \centering
\input{files/pgfs/non_uncertainty/amazon_photos_gcn.pgf}
      \caption{AL strategies on AmazonPhotos using a GCN.}
      \label{fig:non_uncertainty_amazon_photos_gcn}
    \end{subfigure}
    \hfill
    \begin{subfigure}{.49\textwidth}
      \centering
\input{files/pgfs/uncertainty/amazon_photos.pgf}
      \caption{US on AmazonPhotos.}
      \label{fig:uncertainty_amazon_photos}
    \end{subfigure}
    \begin{subfigure}{.49\textwidth}
        \centering
\input{files/pgfs/non_uncertainty/amazon_photos_appnp.pgf}
      \caption{AL strategies on AmazonPhotos using APPNP.}
      \label{fig:non_uncertainty_amazon_photos_appnp}
    \end{subfigure}
    \hfill
    \begin{subfigure}{.49\textwidth}
      \centering
\input{files/pgfs/non_uncertainty/amazon_photos_sgc.pgf}
      \caption{AL strategies on AmazonPhotos using a SGC.}
      \label{fig:non_uncertainty_amazon_photos_sgc}
    \end{subfigure}
    \caption{Accuracy curves of AL strategies, both non-uncertainty-based as well as US on different datasets for different models (cont.).}
\end{figure}
\begin{figure}[ht]\ContinuedFloat
    \centering
    \begin{subfigure}{.49\textwidth}
        \centering
\input{files/pgfs/non_uncertainty/amazon_computers_gcn.pgf}
      \caption{AL strategies on AmazonComputers using a GCN.}
      \label{fig:non_uncertainty_amazon_computers_gcn}
    \end{subfigure}
    \hfill
    \begin{subfigure}{.49\textwidth}
      \centering
\input{files/pgfs/uncertainty/amazon_computers.pgf}
      \caption{US on AmazonComputers.}
      \label{fig:uncertainty_amazon_computers}
    \end{subfigure}
    \begin{subfigure}{.49\textwidth}
        \centering
\input{files/pgfs/non_uncertainty/amazon_computers_appnp.pgf}
      \caption{AL strategies on AmazonComputers using APPNP.}
      \label{fig:non_uncertainty_amazon_computers_appnp}
    \end{subfigure}
    \hfill
    \begin{subfigure}{.49\textwidth}
      \centering
\input{files/pgfs/non_uncertainty/amazon_computers_sgc.pgf}
      \caption{AL strategies on AmazonComputers using a SGC.}
      \label{fig:non_uncertainty_amazon_computers_sgc}
    \end{subfigure}
    \label{fig:appendix_all}
    \caption{Accuracy curves of AL strategies, both non-uncertainty-based as well as US on different datasets for different models (cont.).}
\end{figure}

\begin{inparaenum}[(i)]
    \item \textbf{CoraML.} \Cref{fig:non_uncertainty_cora_ml_gcn,fig:uncertainty_cora_ml,fig:non_uncertainty_cora_ml_sgc,fig:non_uncertainty_cora_ml_appnp} show that apart from GEEM no AL strategy is significantly more effective than random sampling. For an SGC classifier, GEEM can identify high-quality training sets. In terms of US approaches, only ensemble methods perform somewhat better than random acquisition.
    \item \textbf{Citeseer.} \Cref{fig:non_uncertainty_citeseer_gcn,fig:uncertainty_citeseer,fig:non_uncertainty_citeseer_sgc,fig:non_uncertainty_citeseer_appnp} show that no AL strategy can match the performance of the GEEM. Again, risk minimization is the strongest performing approach. Only ensembles and energy-based models can compete with random acquisition when concerning epistemic uncertainty estimates. We verify the intriguing trend observed in \Cref{sec:al_methods} that epistemic uncertainty proxies seem to significantly underperform random queries when being employed for US. This supports our conjecture that contemporary estimators may not properly disentangle or only partially model uncertainty.
    \item \textbf{PubMed.} In \Cref{fig:non_uncertainty_pubmed_gcn,fig:uncertainty_pubmed,fig:non_uncertainty_pubmed_sgc,fig:non_uncertainty_pubmed_appnp}, GEEM is outperformed by the AGE baseline. For this dataset, acquiring central nodes (PPR, AGE) appears to be a successful strategy. Interestingly, ANRMAB fails to exploit centrality despite it, in theory, being capable of doing so. US appears to be effective only when employing ensemble methods.
    \item \textbf{AmazonPhotos.} On this dataset, GEEM outperforms a centrality-based (PPR) Coreset approach (see \Cref{fig:non_uncertainty_amazon_photos_gcn,fig:uncertainty_amazon_photos,fig:non_uncertainty_amazon_photos_sgc,fig:non_uncertainty_amazon_photos_appnp}. Concerning US, we again find that only ensemble methods outperform random sampling while other estimators lead to significantly worse performance.
    \item \textbf{Amazon Computers.} Similar to the other co-purchase network, we observe Coreset PPR to be a strong proxy for AL on the AmazonComputers dataset (\Cref{fig:non_uncertainty_amazon_computers_gcn,fig:uncertainty_amazon_computers,fig:non_uncertainty_amazon_computers_sgc,fig:non_uncertainty_amazon_computers_appnp}). For this dataset, all US approaches--including ensembles-- perform significantly worse than a random strategy.
\end{inparaenum}

This affirms the statements of \Cref{sec:al_methods}: Among non-uncertainty strategies, only GEEM can consistently outperform random queries. In turn, most US approaches consistently underperform random queries, a trend that may be indicative of improperly disentangled uncertainty. Only ensembles show small merit in most cases and, at least, do not fall short of random queries. 

We supplement our findings by reporting the accuracy of each classifier after the acquisition budget is exhausted in \Cref{tab:final_acc_with_sd}. The corresponding rankings align well with the respective AUC scores \Cref{tab:auc_with_sd}. Both metrics are averaged over all 5 dataset splits and model initializations and we also report the standard deviations.

\newpage

\begin{sidewaystable}[h!]
\centering
\caption{Average AUC ($\uparrow$) for different acquisition strategies on different models and datasets and the corresponding standard deviation over 5 different dataset splits and 5 model initializations each. We mark the best strategy per model in bold and underline the second best. For each dataset, we additionally mark the overall best model and strategy with the $\dag$ symbol.}

\label{tab:auc_with_sd}
\resizebox{\columnwidth}{!}{
\begin{tabular}{ll|c|ccccccc|ccc|cc|cc|cc}\Xhline{3\arrayrulewidth}
\multirow{3}{*}{\rotatebox[origin=c]{90}{}} &  & \multicolumn{1}{c|}{\makecell{\textbf{Baselines}}} & \multicolumn{12}{c|}{\makecell{\textbf{Non-Uncertainty}}} & \multicolumn{4}{c}{\makecell{\textbf{Uncertainty}}}\\
\hline & \makecell[r]{\textbf{Inputs}} & \multicolumn{1}{c|}{\makecell{}} & \multicolumn{7}{c|}{\makecell{$ A$ \& $ X$}} & \multicolumn{3}{c|}{\makecell{$ A$}} & \multicolumn{2}{c|}{\makecell{$ X$}} & \multicolumn{2}{c|}{\makecell{$ A$ \& $ X$}} & \multicolumn{2}{c}{\makecell{$ X$}}\\
 & \textbf{Model} & \makecell{Random} & \makecell{Coreset} & \makecell{AGE} & \makecell{ANRMAB} & \makecell{GEEM} & \makecell{SEAL} & \makecell{GALAXY} & \makecell{Badge} & \makecell{Coreset\\PPR} & \makecell{PPR} & \makecell{Degree} & \makecell{Coreset\\w/o Net} & \makecell{Coreset\\Inputs} & \makecell{Epi./\\(Energy)} & \makecell{Alea.} & \makecell{Epi./\\(Energy)} & \makecell{Alea.}\\
\hline \multirow{7}{*}{\rotatebox[origin=c]{90}{\makecell{CoraML}}} & GCN & $62.51$\tiny{$\pm 4.31$} & $64.35$\tiny{$\pm 4.27$} & $64.12$\tiny{$\pm 3.67$} & $64.24$\tiny{$\pm 4.27$} & \small{n/a} & \npboldmath{$66.07$}\tiny{$\pm 3.94$} & $63.24$\tiny{$\pm 5.49$} & $64.51$\tiny{$\pm 4.14$} & $59.53$\tiny{$\pm 7.57$} & $62.16$\tiny{$\pm 4.53$} & $58.39$\tiny{$\pm 5.01$} & $60.73$\tiny{$\pm 4.33$} & $61.26$\tiny{$\pm 4.35$} & $63.97$\tiny{$\pm 6.29$} & $61.30$\tiny{$\pm 5.63$} & \underline{$65.65$}\tiny{$\pm 4.28$} & $64.33$\tiny{$\pm 4.17$}\\
 & APPNP & $67.72$\tiny{$\pm 5.70$} & $67.32$\tiny{$\pm 4.20$} & $66.12$\tiny{$\pm 3.36$} & $69.49$\tiny{$\pm 5.52$} & \small{n/a} & \small{n/a} & \small{n/a} & \small{n/a} & \npboldmath{$71.04$}\tiny{$\pm 6.28$} & $65.52$\tiny{$\pm 3.48$} & $64.57$\tiny{$\pm 4.79$} & $62.01$\tiny{$\pm 5.26$} & $64.49$\tiny{$\pm 4.38$} & $64.92$\tiny{$\pm 7.02$} & $67.68$\tiny{$\pm 5.09$} & \underline{$69.59$}\tiny{$\pm 4.28$} & $66.69$\tiny{$\pm 5.03$}\\
 & Ensemble & $63.89$\tiny{$\pm 6.15$} & $60.55$\tiny{$\pm 5.52$} & $64.80$\tiny{$\pm 4.33$} & $65.10$\tiny{$\pm 7.29$} & \small{n/a} & \small{n/a} & \small{n/a} & \small{n/a} & $62.65$\tiny{$\pm 7.08$} & \underline{$65.47$}\tiny{$\pm 4.26$} & $62.41$\tiny{$\pm 3.04$} & $60.52$\tiny{$\pm 7.37$} & $65.07$\tiny{$\pm 5.11$} & $63.47$\tiny{$\pm 6.20$} & $64.03$\tiny{$\pm 6.68$} & $64.80$\tiny{$\pm 5.14$} & \npboldmath{$65.82$}\tiny{$\pm 3.01$}\\
 & MC-Dropout & \npboldmath{$64.94$}\tiny{$\pm 4.93$} & $64.37$\tiny{$\pm 4.58$} & \underline{$64.44$}\tiny{$\pm 4.37$} & $64.06$\tiny{$\pm 4.94$} & \small{n/a} & \small{n/a} & \small{n/a} & \small{n/a} & $62.92$\tiny{$\pm 6.38$} & $63.91$\tiny{$\pm 5.68$} & $61.65$\tiny{$\pm 3.90$} & $59.58$\tiny{$\pm 4.58$} & $64.35$\tiny{$\pm 4.29$} & $59.17$\tiny{$\pm 5.57$} & $63.69$\tiny{$\pm 4.39$} & $61.82$\tiny{$\pm 5.19$} & $63.87$\tiny{$\pm 4.87$}\\
 & BGCN & $45.76$\tiny{$\pm 3.26$} & $49.37$\tiny{$\pm 3.47$} & \underline{$51.25$}\tiny{$\pm 4.48$} & $47.23$\tiny{$\pm 3.04$} & \small{n/a} & \small{n/a} & \small{n/a} & \small{n/a} & $39.43$\tiny{$\pm 3.87$} & \npboldmath{$54.64$}\tiny{$\pm 2.54$} & $48.68$\tiny{$\pm 3.68$} & $43.65$\tiny{$\pm 3.73$} & $44.85$\tiny{$\pm 3.15$} & $44.45$\tiny{$\pm 2.28$} & $46.61$\tiny{$\pm 3.07$} & $42.74$\tiny{$\pm 3.57$} & $48.11$\tiny{$\pm 3.16$}\\
 & GPN & $56.50$\tiny{$\pm 4.73$} & \small{n/a} & \small{n/a} & \small{n/a} & \small{n/a} & \small{n/a} & \small{n/a} & \small{n/a} & $58.04$\tiny{$\pm 4.56$} & \npboldmath{$59.88$}\tiny{$\pm 3.05$} & \underline{$58.74$}\tiny{$\pm 3.53$} & \small{n/a} & $54.02$\tiny{$\pm 3.83$} & $54.75$\tiny{$\pm 3.21$} & $57.16$\tiny{$\pm 5.25$} & $55.89$\tiny{$\pm 4.29$} & $57.21$\tiny{$\pm 5.12$}\\
 & SGC & $63.85$\tiny{$\pm 6.05$} & $65.23$\tiny{$\pm 5.14$} & \underline{$67.56$}\tiny{$\pm 3.20$} & $61.14$\tiny{$\pm 5.50$} & \npboldmath{$71.39^\dagger$}\tiny{$\pm 3.37$} & \small{n/a} & \small{n/a} & \small{n/a} & $60.24$\tiny{$\pm 7.98$} & $65.05$\tiny{$\pm 2.85$} & $61.60$\tiny{$\pm 4.34$} & $62.94$\tiny{$\pm 5.41$} & $59.18$\tiny{$\pm 5.42$} & $67.51$\tiny{$\pm 3.64$} & $65.66$\tiny{$\pm 5.70$} & $65.05$\tiny{$\pm 4.71$} & $67.13$\tiny{$\pm 4.11$}\\
\hline \multirow{7}{*}{\rotatebox[origin=c]{90}{\makecell{Citeseer}}} & GCN & $80.34$\tiny{$\pm 3.20$} & $80.36$\tiny{$\pm 2.81$} & $80.14$\tiny{$\pm 3.00$} & \underline{$81.59$}\tiny{$\pm 3.11$} & \small{n/a} & $79.19$\tiny{$\pm 4.86$} & $81.06$\tiny{$\pm 3.59$} & $78.41$\tiny{$\pm 4.78$} & $81.51$\tiny{$\pm 2.97$} & $79.93$\tiny{$\pm 2.16$} & $79.72$\tiny{$\pm 2.47$} & $81.41$\tiny{$\pm 2.28$} & $76.87$\tiny{$\pm 5.65$} & \underline{$81.59$}\tiny{$\pm 3.36$} & $80.15$\tiny{$\pm 5.10$} & \npboldmath{$82.26$}\tiny{$\pm 2.82$} & $76.79$\tiny{$\pm 4.07$}\\
 & APPNP & $80.77$\tiny{$\pm 5.34$} & \underline{$82.62$}\tiny{$\pm 2.53$} & \npboldmath{$83.00$}\tiny{$\pm 2.59$} & $82.31$\tiny{$\pm 3.10$} & \small{n/a} & \small{n/a} & \small{n/a} & \small{n/a} & $80.70$\tiny{$\pm 4.49$} & $81.37$\tiny{$\pm 4.85$} & $81.36$\tiny{$\pm 3.32$} & $80.89$\tiny{$\pm 5.26$} & $79.20$\tiny{$\pm 7.16$} & $80.37$\tiny{$\pm 6.14$} & $77.77$\tiny{$\pm 8.04$} & $81.94$\tiny{$\pm 4.95$} & $79.15$\tiny{$\pm 4.96$}\\
 & Ensemble & $81.14$\tiny{$\pm 3.99$} & $77.61$\tiny{$\pm 4.28$} & \underline{$82.37$}\tiny{$\pm 2.44$} & $81.43$\tiny{$\pm 3.62$} & \small{n/a} & \small{n/a} & \small{n/a} & \small{n/a} & $81.59$\tiny{$\pm 3.71$} & $81.42$\tiny{$\pm 2.61$} & $80.31$\tiny{$\pm 2.45$} & $80.18$\tiny{$\pm 4.60$} & $79.13$\tiny{$\pm 4.71$} & \npboldmath{$82.94$}\tiny{$\pm 2.72$} & $80.26$\tiny{$\pm 6.82$} & $81.06$\tiny{$\pm 4.56$} & $77.76$\tiny{$\pm 5.03$}\\
 & MC-Dropout & \npboldmath{$81.90$}\tiny{$\pm 3.19$} & $78.97$\tiny{$\pm 3.47$} & $79.86$\tiny{$\pm 3.31$} & $80.46$\tiny{$\pm 2.99$} & \small{n/a} & \small{n/a} & \small{n/a} & \small{n/a} & \underline{$81.09$}\tiny{$\pm 3.38$} & $79.86$\tiny{$\pm 3.58$} & $80.82$\tiny{$\pm 2.85$} & $79.40$\tiny{$\pm 4.15$} & $78.62$\tiny{$\pm 4.98$} & $78.86$\tiny{$\pm 5.47$} & $80.97$\tiny{$\pm 5.28$} & $78.85$\tiny{$\pm 4.73$} & $76.02$\tiny{$\pm 4.46$}\\
 & BGCN & $70.17$\tiny{$\pm 3.64$} & $73.86$\tiny{$\pm 2.14$} & \npboldmath{$76.30$}\tiny{$\pm 1.42$} & $70.70$\tiny{$\pm 3.28$} & \small{n/a} & \small{n/a} & \small{n/a} & \small{n/a} & $69.32$\tiny{$\pm 3.41$} & $73.77$\tiny{$\pm 2.73$} & \underline{$75.55$}\tiny{$\pm 1.47$} & $68.59$\tiny{$\pm 3.91$} & $67.51$\tiny{$\pm 3.64$} & $58.68$\tiny{$\pm 7.96$} & $71.23$\tiny{$\pm 3.80$} & $54.85$\tiny{$\pm 6.96$} & $70.33$\tiny{$\pm 4.14$}\\
 & GPN & $77.07$\tiny{$\pm 5.06$} & \small{n/a} & \small{n/a} & \small{n/a} & \small{n/a} & \small{n/a} & \small{n/a} & \small{n/a} & $72.83$\tiny{$\pm 6.88$} & \underline{$79.52$}\tiny{$\pm 2.53$} & \npboldmath{$80.65$}\tiny{$\pm 1.69$} & \small{n/a} & $71.33$\tiny{$\pm 7.55$} & $65.31$\tiny{$\pm 10.21$} & $74.28$\tiny{$\pm 7.63$} & $78.73$\tiny{$\pm 5.24$} & $77.28$\tiny{$\pm 4.85$}\\
 & SGC & $81.04$\tiny{$\pm 5.06$} & $79.38$\tiny{$\pm 4.44$} & \underline{$84.21$}\tiny{$\pm 1.95$} & $81.03$\tiny{$\pm 3.28$} & \npboldmath{$85.25^\dagger$}\tiny{$\pm 1.80$} & \small{n/a} & \small{n/a} & \small{n/a} & $81.59$\tiny{$\pm 4.15$} & $83.25$\tiny{$\pm 2.74$} & $82.60$\tiny{$\pm 1.49$} & $79.21$\tiny{$\pm 7.73$} & $72.04$\tiny{$\pm 5.66$} & $75.13$\tiny{$\pm 9.12$} & $78.85$\tiny{$\pm 6.36$} & $72.94$\tiny{$\pm 8.35$} & $82.90$\tiny{$\pm 1.77$}\\
\hline \multirow{7}{*}{\rotatebox[origin=c]{90}{\makecell{Pubmed}}} & GCN & $61.56$\tiny{$\pm 4.52$} & $62.61$\tiny{$\pm 5.20$} & \npboldmath{$69.48$}\tiny{$\pm 4.12$} & $60.31$\tiny{$\pm 6.68$} & \small{n/a} & $58.62$\tiny{$\pm 5.27$} & $62.05$\tiny{$\pm 5.73$} & $63.34$\tiny{$\pm 6.01$} & $61.71$\tiny{$\pm 7.24$} & \underline{$66.07$}\tiny{$\pm 4.01$} & $63.84$\tiny{$\pm 5.40$} & $60.33$\tiny{$\pm 6.04$} & $56.71$\tiny{$\pm 6.44$} & $59.64$\tiny{$\pm 6.02$} & $61.85$\tiny{$\pm 5.43$} & $59.66$\tiny{$\pm 5.23$} & $60.34$\tiny{$\pm 5.98$}\\
 & APPNP & $64.61$\tiny{$\pm 6.05$} & $63.88$\tiny{$\pm 6.47$} & \npboldmath{$70.18^\dagger$}\tiny{$\pm 4.50$} & $63.83$\tiny{$\pm 5.85$} & \small{n/a} & \small{n/a} & \small{n/a} & \small{n/a} & $64.21$\tiny{$\pm 6.62$} & \underline{$68.24$}\tiny{$\pm 4.44$} & $66.09$\tiny{$\pm 4.25$} & $65.51$\tiny{$\pm 4.18$} & $56.87$\tiny{$\pm 8.10$} & $63.09$\tiny{$\pm 7.08$} & $62.37$\tiny{$\pm 5.82$} & $62.23$\tiny{$\pm 7.74$} & $63.95$\tiny{$\pm 6.34$}\\
 & Ensemble & $59.26$\tiny{$\pm 6.36$} & \underline{$64.25$}\tiny{$\pm 6.60$} & \npboldmath{$68.26$}\tiny{$\pm 4.28$} & $60.40$\tiny{$\pm 6.98$} & \small{n/a} & \small{n/a} & \small{n/a} & \small{n/a} & $61.89$\tiny{$\pm 5.92$} & $63.48$\tiny{$\pm 7.18$} & $62.78$\tiny{$\pm 7.21$} & $59.19$\tiny{$\pm 6.73$} & $56.36$\tiny{$\pm 5.41$} & $63.70$\tiny{$\pm 6.26$} & $61.37$\tiny{$\pm 4.97$} & $59.71$\tiny{$\pm 8.14$} & $61.15$\tiny{$\pm 4.90$}\\
 & MC-Dropout & $58.30$\tiny{$\pm 6.21$} & $62.97$\tiny{$\pm 6.29$} & $65.24$\tiny{$\pm 6.30$} & $60.50$\tiny{$\pm 4.68$} & \small{n/a} & \small{n/a} & \small{n/a} & \small{n/a} & $61.43$\tiny{$\pm 5.41$} & \npboldmath{$65.72$}\tiny{$\pm 3.92$} & \underline{$65.29$}\tiny{$\pm 5.23$} & $57.71$\tiny{$\pm 5.70$} & $56.01$\tiny{$\pm 6.09$} & $58.67$\tiny{$\pm 7.52$} & $59.07$\tiny{$\pm 7.03$} & $59.23$\tiny{$\pm 6.99$} & $62.22$\tiny{$\pm 4.68$}\\
 & BGCN & $53.59$\tiny{$\pm 5.46$} & \npboldmath{$59.29$}\tiny{$\pm 4.21$} & $56.93$\tiny{$\pm 3.14$} & $52.68$\tiny{$\pm 2.92$} & \small{n/a} & \small{n/a} & \small{n/a} & \small{n/a} & $53.40$\tiny{$\pm 4.46$} & \underline{$59.12$}\tiny{$\pm 2.49$} & $55.93$\tiny{$\pm 5.47$} & $53.63$\tiny{$\pm 4.50$} & $51.40$\tiny{$\pm 4.15$} & $55.19$\tiny{$\pm 3.29$} & $52.81$\tiny{$\pm 4.77$} & $57.09$\tiny{$\pm 2.93$} & $54.62$\tiny{$\pm 5.29$}\\
 & GPN & $59.76$\tiny{$\pm 6.46$} & \small{n/a} & \small{n/a} & \small{n/a} & \small{n/a} & \small{n/a} & \small{n/a} & \small{n/a} & \underline{$62.08$}\tiny{$\pm 6.79$} & \npboldmath{$64.64$}\tiny{$\pm 3.83$} & $58.13$\tiny{$\pm 4.94$} & \small{n/a} & $54.34$\tiny{$\pm 7.91$} & $58.82$\tiny{$\pm 8.90$} & $57.24$\tiny{$\pm 8.31$} & $56.25$\tiny{$\pm 6.23$} & $59.37$\tiny{$\pm 6.96$}\\
 & SGC & $56.79$\tiny{$\pm 6.28$} & $64.48$\tiny{$\pm 5.35$} & \npboldmath{$69.20$}\tiny{$\pm 4.83$} & $60.49$\tiny{$\pm 6.30$} & $64.82$\tiny{$\pm 3.68$} & \small{n/a} & \small{n/a} & \small{n/a} & $62.15$\tiny{$\pm 5.47$} & $63.88$\tiny{$\pm 4.65$} & \underline{$65.14$}\tiny{$\pm 4.69$} & $59.57$\tiny{$\pm 4.45$} & $52.25$\tiny{$\pm 6.89$} & $62.04$\tiny{$\pm 5.24$} & $61.55$\tiny{$\pm 6.65$} & $61.04$\tiny{$\pm 5.84$} & $60.74$\tiny{$\pm 6.28$}\\
\hline \multirow{7}{*}{\rotatebox[origin=c]{90}{\makecell{AmazonPhotos}}} & GCN & $79.06$\tiny{$\pm 3.86$} & $78.58$\tiny{$\pm 2.44$} & $75.17$\tiny{$\pm 5.50$} & \underline{$79.97$}\tiny{$\pm 4.20$} & \small{n/a} & $71.16$\tiny{$\pm 3.02$} & $78.87$\tiny{$\pm 4.84$} & $76.97$\tiny{$\pm 5.12$} & $70.20$\tiny{$\pm 8.55$} & $76.30$\tiny{$\pm 3.13$} & $71.00$\tiny{$\pm 5.92$} & $75.40$\tiny{$\pm 4.41$} & \npboldmath{$82.71$}\tiny{$\pm 3.03$} & $74.66$\tiny{$\pm 7.12$} & $74.61$\tiny{$\pm 5.99$} & $79.96$\tiny{$\pm 4.84$} & $79.63$\tiny{$\pm 4.59$}\\
 & APPNP & $79.29$\tiny{$\pm 6.17$} & $81.04$\tiny{$\pm 2.86$} & $79.02$\tiny{$\pm 5.26$} & $80.35$\tiny{$\pm 4.96$} & \small{n/a} & \small{n/a} & \small{n/a} & \small{n/a} & $76.37$\tiny{$\pm 7.76$} & $79.03$\tiny{$\pm 3.83$} & $73.25$\tiny{$\pm 8.33$} & \underline{$82.31$}\tiny{$\pm 3.29$} & \npboldmath{$84.24$}\tiny{$\pm 4.37$} & $79.72$\tiny{$\pm 6.19$} & $77.45$\tiny{$\pm 6.13$} & $80.48$\tiny{$\pm 4.97$} & $77.69$\tiny{$\pm 5.83$}\\
 & Ensemble & $82.23$\tiny{$\pm 2.91$} & $80.44$\tiny{$\pm 4.48$} & $77.45$\tiny{$\pm 3.41$} & $82.77$\tiny{$\pm 4.62$} & \small{n/a} & \small{n/a} & \small{n/a} & \small{n/a} & $74.93$\tiny{$\pm 8.07$} & $77.32$\tiny{$\pm 4.53$} & $75.17$\tiny{$\pm 6.07$} & $76.95$\tiny{$\pm 5.38$} & \underline{$84.04$}\tiny{$\pm 3.15$} & \npboldmath{$84.46$}\tiny{$\pm 3.69$} & $77.85$\tiny{$\pm 6.69$} & $80.50$\tiny{$\pm 4.83$} & $81.25$\tiny{$\pm 4.27$}\\
 & MC-Dropout & \underline{$80.32$}\tiny{$\pm 3.75$} & $76.63$\tiny{$\pm 4.47$} & $74.75$\tiny{$\pm 3.45$} & $80.21$\tiny{$\pm 3.25$} & \small{n/a} & \small{n/a} & \small{n/a} & \small{n/a} & $75.32$\tiny{$\pm 6.70$} & $74.33$\tiny{$\pm 3.46$} & $69.03$\tiny{$\pm 6.60$} & $75.71$\tiny{$\pm 4.62$} & \npboldmath{$82.45$}\tiny{$\pm 2.84$} & $72.42$\tiny{$\pm 6.79$} & $73.16$\tiny{$\pm 6.19$} & $69.68$\tiny{$\pm 8.19$} & $78.80$\tiny{$\pm 4.17$}\\
 & BGCN & $71.22$\tiny{$\pm 3.86$} & $67.15$\tiny{$\pm 4.70$} & $65.69$\tiny{$\pm 5.06$} & $70.69$\tiny{$\pm 3.38$} & \small{n/a} & \small{n/a} & \small{n/a} & \small{n/a} & $59.34$\tiny{$\pm 7.49$} & $64.51$\tiny{$\pm 4.28$} & $61.23$\tiny{$\pm 5.84$} & $69.78$\tiny{$\pm 4.24$} & \npboldmath{$73.39$}\tiny{$\pm 4.16$} & $70.83$\tiny{$\pm 3.36$} & $67.83$\tiny{$\pm 5.83$} & \underline{$72.21$}\tiny{$\pm 2.84$} & $69.19$\tiny{$\pm 5.80$}\\
 & GPN & $62.80$\tiny{$\pm 4.22$} & \small{n/a} & \small{n/a} & \small{n/a} & \small{n/a} & \small{n/a} & \small{n/a} & \small{n/a} & $55.59$\tiny{$\pm 4.09$} & $62.17$\tiny{$\pm 4.08$} & $56.77$\tiny{$\pm 4.94$} & \small{n/a} & \npboldmath{$65.07$}\tiny{$\pm 2.74$} & $54.78$\tiny{$\pm 3.57$} & $60.53$\tiny{$\pm 4.17$} & \underline{$62.90$}\tiny{$\pm 2.61$} & $62.41$\tiny{$\pm 2.75$}\\
 & SGC & $80.52$\tiny{$\pm 4.89$} & $82.32$\tiny{$\pm 2.58$} & $74.01$\tiny{$\pm 5.96$} & $80.92$\tiny{$\pm 4.04$} & \npboldmath{$86.43^\dagger$}\tiny{$\pm 4.13$} & \small{n/a} & \small{n/a} & \small{n/a} & $66.94$\tiny{$\pm 7.86$} & $76.16$\tiny{$\pm 4.06$} & $66.63$\tiny{$\pm 8.12$} & $81.47$\tiny{$\pm 5.08$} & \underline{$84.24$}\tiny{$\pm 3.59$} & $84.01$\tiny{$\pm 4.72$} & $71.43$\tiny{$\pm 8.37$} & $80.75$\tiny{$\pm 4.23$} & $76.38$\tiny{$\pm 7.40$}\\
\hline \multirow{7}{*}{\rotatebox[origin=c]{90}{\makecell{AmazonComputers}}} & GCN & $69.80$\tiny{$\pm 3.43$} & $59.36$\tiny{$\pm 4.83$} & $60.07$\tiny{$\pm 6.85$} & $70.70$\tiny{$\pm 2.66$} & \small{n/a} & $61.51$\tiny{$\pm 4.11$} & \underline{$72.32$}\tiny{$\pm 2.47$} & $58.48$\tiny{$\pm 4.40$} & $61.22$\tiny{$\pm 5.54$} & $57.45$\tiny{$\pm 4.48$} & $58.14$\tiny{$\pm 5.29$} & $65.13$\tiny{$\pm 4.31$} & \npboldmath{$72.34$}\tiny{$\pm 2.75$} & $59.62$\tiny{$\pm 7.26$} & $60.17$\tiny{$\pm 6.98$} & $69.34$\tiny{$\pm 4.75$} & $69.87$\tiny{$\pm 3.77$}\\
 & APPNP & $71.69$\tiny{$\pm 3.34$} & $66.62$\tiny{$\pm 3.55$} & $68.89$\tiny{$\pm 3.32$} & \underline{$72.69$}\tiny{$\pm 3.27$} & \small{n/a} & \small{n/a} & \small{n/a} & \small{n/a} & $65.75$\tiny{$\pm 6.76$} & $70.91$\tiny{$\pm 3.52$} & $66.42$\tiny{$\pm 4.73$} & $70.22$\tiny{$\pm 4.44$} & \npboldmath{$73.83$}\tiny{$\pm 3.56$} & $62.03$\tiny{$\pm 8.44$} & $62.26$\tiny{$\pm 10.26$} & $68.79$\tiny{$\pm 5.46$} & $71.72$\tiny{$\pm 3.79$}\\
 & Ensemble & $72.56$\tiny{$\pm 3.06$} & $64.19$\tiny{$\pm 4.49$} & $60.69$\tiny{$\pm 5.43$} & $72.73$\tiny{$\pm 4.16$} & \small{n/a} & \small{n/a} & \small{n/a} & \small{n/a} & $64.13$\tiny{$\pm 6.76$} & $60.59$\tiny{$\pm 7.02$} & $63.16$\tiny{$\pm 5.33$} & $67.20$\tiny{$\pm 6.40$} & \npboldmath{$75.39^\dagger$}\tiny{$\pm 2.68$} & $68.38$\tiny{$\pm 5.04$} & $68.47$\tiny{$\pm 7.74$} & $69.49$\tiny{$\pm 8.35$} & \underline{$73.67$}\tiny{$\pm 3.30$}\\
 & MC-Dropout & $68.06$\tiny{$\pm 4.65$} & $56.58$\tiny{$\pm 4.30$} & $57.04$\tiny{$\pm 4.61$} & $69.01$\tiny{$\pm 2.92$} & \small{n/a} & \small{n/a} & \small{n/a} & \small{n/a} & $64.43$\tiny{$\pm 6.67$} & $55.88$\tiny{$\pm 4.81$} & $56.72$\tiny{$\pm 6.12$} & $61.86$\tiny{$\pm 3.16$} & \npboldmath{$71.05$}\tiny{$\pm 2.88$} & $51.02$\tiny{$\pm 9.05$} & $59.31$\tiny{$\pm 7.96$} & $60.71$\tiny{$\pm 7.40$} & \underline{$70.66$}\tiny{$\pm 3.97$}\\
 & BGCN & \underline{$60.52$}\tiny{$\pm 3.72$} & $45.65$\tiny{$\pm 3.02$} & $46.72$\tiny{$\pm 3.74$} & $60.32$\tiny{$\pm 4.31$} & \small{n/a} & \small{n/a} & \small{n/a} & \small{n/a} & $39.86$\tiny{$\pm 8.46$} & $43.61$\tiny{$\pm 4.17$} & $45.52$\tiny{$\pm 2.93$} & $58.85$\tiny{$\pm 3.41$} & \npboldmath{$60.79$}\tiny{$\pm 3.89$} & $58.64$\tiny{$\pm 5.96$} & $51.11$\tiny{$\pm 6.23$} & $60.19$\tiny{$\pm 3.82$} & $59.20$\tiny{$\pm 3.65$}\\
 & GPN & $52.26$\tiny{$\pm 5.90$} & \small{n/a} & \small{n/a} & \small{n/a} & \small{n/a} & \small{n/a} & \small{n/a} & \small{n/a} & $34.71$\tiny{$\pm 4.90$} & $49.30$\tiny{$\pm 5.00$} & $45.21$\tiny{$\pm 5.32$} & \small{n/a} & \npboldmath{$56.32$}\tiny{$\pm 3.79$} & $39.21$\tiny{$\pm 3.92$} & $42.95$\tiny{$\pm 4.53$} & \underline{$54.83$}\tiny{$\pm 4.84$} & $53.71$\tiny{$\pm 3.91$}\\
 & SGC & $72.39$\tiny{$\pm 2.74$} & $71.53$\tiny{$\pm 3.86$} & $69.31$\tiny{$\pm 4.37$} & $71.62$\tiny{$\pm 3.79$} & \npboldmath{$74.49$}\tiny{$\pm 3.43$} & \small{n/a} & \small{n/a} & \small{n/a} & $58.52$\tiny{$\pm 0.44$} & $70.35$\tiny{$\pm 3.84$} & $68.70$\tiny{$\pm 3.24$} & $68.14$\tiny{$\pm 5.46$} & \underline{$73.91$}\tiny{$\pm 3.30$} & $61.02$\tiny{$\pm 8.74$} & $59.34$\tiny{$\pm 10.50$} & $66.53$\tiny{$\pm 5.88$} & $69.24$\tiny{$\pm 5.52$}\\
\Xhline{3\arrayrulewidth}
\end{tabular}

}

\end{sidewaystable}

\begin{sidewaystable}[h!]
\centering
\caption{Average final classification accuracy ($\uparrow$) for different acquisition strategies on different models and datasets and the corresponding standard deviation over 5 different dataset splits and 5 model initializations each. We mark the best strategy per model in bold and underline the second best. For each dataset, we additionally mark the overall best model and strategy with the $\dag$ symbol.}
\label{tab:final_acc_with_sd}

\resizebox{\columnwidth}{!}{
\begin{tabular}{ll|c|ccccccc|ccc|cc|cc|cc}\Xhline{3\arrayrulewidth}
\multirow{3}{*}{\rotatebox[origin=c]{90}{}} &  & \multicolumn{1}{c|}{\makecell{\textbf{Baselines}}} & \multicolumn{12}{c|}{\makecell{\textbf{Non-Uncertainty}}} & \multicolumn{4}{c}{\makecell{\textbf{Uncertainty}}}\\
\hline & \makecell[r]{\textbf{Inputs}} & \multicolumn{1}{c|}{\makecell{}} & \multicolumn{7}{c|}{\makecell{$ A$ \& $ X$}} & \multicolumn{3}{c|}{\makecell{$ A$}} & \multicolumn{2}{c|}{\makecell{$ X$}} & \multicolumn{2}{c|}{\makecell{$ A$ \& $ X$}} & \multicolumn{2}{c}{\makecell{$ X$}}\\
 & \textbf{Model} & \makecell{Random} & \makecell{Coreset} & \makecell{AGE} & \makecell{ANRMAB} & \makecell{GEEM} & \makecell{SEAL} & \makecell{GALAXY} & \makecell{Badge} & \makecell{Coreset\\PPR} & \makecell{PPR} & \makecell{Degree} & \makecell{Coreset\\w/o Net} & \makecell{Coreset\\Inputs} & \makecell{Epi./\\(Energy)} & \makecell{Alea.} & \makecell{Epi./\\(Energy)} & \makecell{Alea.}\\
\hline \multirow{7}{*}{\rotatebox[origin=c]{90}{\makecell{CoraML}}} & GCN & $72.80$\tiny{$\pm 3.74$} & $72.95$\tiny{$\pm 4.17$} & $72.54$\tiny{$\pm 3.61$} & $74.64$\tiny{$\pm 3.46$} & \small{n/a} & $73.08$\tiny{$\pm 6.13$} & \underline{$74.67$}\tiny{$\pm 4.82$} & $72.51$\tiny{$\pm 4.05$} & $69.33$\tiny{$\pm 6.04$} & $71.10$\tiny{$\pm 4.26$} & $67.31$\tiny{$\pm 5.93$} & $72.36$\tiny{$\pm 4.27$} & $71.21$\tiny{$\pm 4.18$} & $71.34$\tiny{$\pm 5.39$} & $70.69$\tiny{$\pm 7.56$} & \npboldmath{$75.59$}\tiny{$\pm 2.82$} & $74.08$\tiny{$\pm 3.79$}\\
 & APPNP & $76.74$\tiny{$\pm 4.45$} & $74.33$\tiny{$\pm 4.02$} & $74.86$\tiny{$\pm 2.76$} & \underline{$77.71$}\tiny{$\pm 4.59$} & \small{n/a} & \small{n/a} & \small{n/a} & \small{n/a} & $76.11$\tiny{$\pm 4.08$} & $74.99$\tiny{$\pm 2.24$} & $71.81$\tiny{$\pm 3.78$} & $72.78$\tiny{$\pm 3.84$} & $74.69$\tiny{$\pm 3.38$} & $72.29$\tiny{$\pm 5.83$} & $75.61$\tiny{$\pm 3.88$} & \npboldmath{$77.91$}\tiny{$\pm 3.03$} & $75.29$\tiny{$\pm 3.71$}\\
 & Ensemble & $74.25$\tiny{$\pm 4.81$} & $68.32$\tiny{$\pm 3.58$} & $73.82$\tiny{$\pm 2.05$} & $75.40$\tiny{$\pm 4.56$} & \small{n/a} & \small{n/a} & \small{n/a} & \small{n/a} & $72.76$\tiny{$\pm 4.57$} & $72.96$\tiny{$\pm 2.64$} & $69.36$\tiny{$\pm 3.30$} & $69.77$\tiny{$\pm 5.69$} & $74.12$\tiny{$\pm 3.77$} & $74.98$\tiny{$\pm 3.58$} & \underline{$75.51$}\tiny{$\pm 3.48$} & $74.63$\tiny{$\pm 3.64$} & \npboldmath{$75.62$}\tiny{$\pm 2.29$}\\
 & MC-Dropout & $73.86$\tiny{$\pm 5.46$} & $72.44$\tiny{$\pm 5.59$} & $72.40$\tiny{$\pm 4.09$} & $73.44$\tiny{$\pm 5.36$} & \small{n/a} & \small{n/a} & \small{n/a} & \small{n/a} & $70.91$\tiny{$\pm 5.15$} & $72.06$\tiny{$\pm 5.10$} & $69.58$\tiny{$\pm 3.78$} & $71.42$\tiny{$\pm 5.19$} & $74.06$\tiny{$\pm 1.94$} & $68.64$\tiny{$\pm 4.57$} & \underline{$74.42$}\tiny{$\pm 4.15$} & $71.67$\tiny{$\pm 4.63$} & \npboldmath{$75.86$}\tiny{$\pm 4.55$}\\
 & BGCN & $59.09$\tiny{$\pm 5.86$} & $61.08$\tiny{$\pm 6.05$} & \underline{$61.20$}\tiny{$\pm 8.35$} & $57.80$\tiny{$\pm 6.47$} & \small{n/a} & \small{n/a} & \small{n/a} & \small{n/a} & $46.46$\tiny{$\pm 7.81$} & \npboldmath{$64.79$}\tiny{$\pm 4.96$} & $59.40$\tiny{$\pm 5.58$} & $56.22$\tiny{$\pm 4.89$} & $55.73$\tiny{$\pm 7.12$} & $56.53$\tiny{$\pm 7.54$} & $59.48$\tiny{$\pm 7.56$} & $47.72$\tiny{$\pm 7.13$} & $57.87$\tiny{$\pm 7.44$}\\
 & GPN & $64.24$\tiny{$\pm 5.57$} & \small{n/a} & \small{n/a} & \small{n/a} & \small{n/a} & \small{n/a} & \small{n/a} & \small{n/a} & $59.66$\tiny{$\pm 7.23$} & \npboldmath{$66.85$}\tiny{$\pm 3.48$} & $65.10$\tiny{$\pm 5.94$} & \small{n/a} & $60.52$\tiny{$\pm 6.49$} & $57.34$\tiny{$\pm 8.78$} & $61.08$\tiny{$\pm 8.07$} & $63.55$\tiny{$\pm 7.16$} & \underline{$65.97$}\tiny{$\pm 7.69$}\\
 & SGC & $73.59$\tiny{$\pm 4.99$} & $75.50$\tiny{$\pm 3.70$} & $77.49$\tiny{$\pm 1.44$} & $72.99$\tiny{$\pm 4.52$} & \npboldmath{$81.89^\dagger$}\tiny{$\pm 1.58$} & \small{n/a} & \small{n/a} & \small{n/a} & $66.40$\tiny{$\pm 6.16$} & \underline{$78.02$}\tiny{$\pm 1.26$} & $73.14$\tiny{$\pm 3.15$} & $74.35$\tiny{$\pm 4.61$} & $72.39$\tiny{$\pm 4.66$} & $77.90$\tiny{$\pm 2.49$} & $75.23$\tiny{$\pm 3.96$} & $76.59$\tiny{$\pm 2.79$} & $77.03$\tiny{$\pm 3.59$}\\
\hline \multirow{7}{*}{\rotatebox[origin=c]{90}{\makecell{Citeseer}}} & GCN & $85.38$\tiny{$\pm 2.24$} & $86.13$\tiny{$\pm 1.80$} & $86.15$\tiny{$\pm 1.99$} & $86.68$\tiny{$\pm 1.89$} & \small{n/a} & \npboldmath{$87.69$}\tiny{$\pm 2.00$} & $87.35$\tiny{$\pm 1.64$} & $86.45$\tiny{$\pm 1.39$} & $86.03$\tiny{$\pm 1.72$} & $84.31$\tiny{$\pm 1.58$} & $84.31$\tiny{$\pm 1.59$} & $85.51$\tiny{$\pm 1.90$} & $84.01$\tiny{$\pm 4.56$} & $86.74$\tiny{$\pm 2.45$} & $86.39$\tiny{$\pm 2.89$} & $87.21$\tiny{$\pm 1.98$} & \underline{$87.39$}\tiny{$\pm 1.95$}\\
 & APPNP & $84.82$\tiny{$\pm 3.78$} & $85.96$\tiny{$\pm 2.64$} & \underline{$87.09$}\tiny{$\pm 1.55$} & $86.09$\tiny{$\pm 2.08$} & \small{n/a} & \small{n/a} & \small{n/a} & \small{n/a} & $85.31$\tiny{$\pm 3.32$} & $86.99$\tiny{$\pm 1.48$} & $85.33$\tiny{$\pm 1.68$} & $85.78$\tiny{$\pm 2.98$} & $82.97$\tiny{$\pm 5.03$} & $86.55$\tiny{$\pm 3.81$} & $84.49$\tiny{$\pm 5.34$} & $85.83$\tiny{$\pm 3.38$} & \npboldmath{$87.27$}\tiny{$\pm 1.90$}\\
 & Ensemble & $85.63$\tiny{$\pm 3.26$} & $85.00$\tiny{$\pm 2.65$} & $86.07$\tiny{$\pm 2.35$} & \underline{$86.77$}\tiny{$\pm 2.16$} & \small{n/a} & \small{n/a} & \small{n/a} & \small{n/a} & $86.19$\tiny{$\pm 3.25$} & $85.07$\tiny{$\pm 1.50$} & $84.90$\tiny{$\pm 1.59$} & $86.05$\tiny{$\pm 1.40$} & $83.98$\tiny{$\pm 3.30$} & \npboldmath{$87.15$}\tiny{$\pm 1.54$} & $86.41$\tiny{$\pm 4.64$} & $86.27$\tiny{$\pm 2.37$} & \npboldmath{$87.15$}\tiny{$\pm 2.28$}\\
 & MC-Dropout & $86.64$\tiny{$\pm 1.90$} & $86.52$\tiny{$\pm 1.88$} & $85.50$\tiny{$\pm 1.83$} & $86.07$\tiny{$\pm 2.77$} & \small{n/a} & \small{n/a} & \small{n/a} & \small{n/a} & $86.01$\tiny{$\pm 2.24$} & $84.56$\tiny{$\pm 1.95$} & $85.23$\tiny{$\pm 1.13$} & $85.12$\tiny{$\pm 2.47$} & $84.74$\tiny{$\pm 3.07$} & $85.63$\tiny{$\pm 2.50$} & \underline{$87.59$}\tiny{$\pm 2.09$} & $84.84$\tiny{$\pm 3.57$} & \npboldmath{$88.06^\dagger$}\tiny{$\pm 2.00$}\\
 & BGCN & $78.89$\tiny{$\pm 7.20$} & \underline{$80.45$}\tiny{$\pm 3.60$} & $79.59$\tiny{$\pm 4.58$} & $79.37$\tiny{$\pm 4.66$} & \small{n/a} & \small{n/a} & \small{n/a} & \small{n/a} & $75.14$\tiny{$\pm 6.50$} & $80.44$\tiny{$\pm 4.52$} & \npboldmath{$81.05$}\tiny{$\pm 3.44$} & $76.46$\tiny{$\pm 5.00$} & $76.41$\tiny{$\pm 3.51$} & $59.50$\tiny{$\pm 13.96$} & $80.08$\tiny{$\pm 5.25$} & $53.42$\tiny{$\pm 12.61$} & $78.44$\tiny{$\pm 5.55$}\\
 & GPN & $80.96$\tiny{$\pm 4.76$} & \small{n/a} & \small{n/a} & \small{n/a} & \small{n/a} & \small{n/a} & \small{n/a} & \small{n/a} & $77.06$\tiny{$\pm 7.62$} & \underline{$83.35$}\tiny{$\pm 2.24$} & $82.46$\tiny{$\pm 2.34$} & \small{n/a} & $79.14$\tiny{$\pm 4.86$} & $72.61$\tiny{$\pm 7.77$} & $81.87$\tiny{$\pm 5.27$} & \npboldmath{$84.58$}\tiny{$\pm 3.86$} & $81.60$\tiny{$\pm 3.65$}\\
 & SGC & $85.45$\tiny{$\pm 3.72$} & $85.26$\tiny{$\pm 2.90$} & \npboldmath{$87.84$}\tiny{$\pm 1.10$} & $87.09$\tiny{$\pm 1.86$} & $87.33$\tiny{$\pm 1.95$} & \small{n/a} & \small{n/a} & \small{n/a} & $87.17$\tiny{$\pm 2.14$} & $87.49$\tiny{$\pm 1.05$} & $86.73$\tiny{$\pm 1.73$} & $86.37$\tiny{$\pm 2.99$} & $80.23$\tiny{$\pm 4.86$} & $75.63$\tiny{$\pm 10.15$} & $86.19$\tiny{$\pm 3.54$} & $80.16$\tiny{$\pm 10.35$} & \underline{$87.78$}\tiny{$\pm 1.98$}\\
\hline \multirow{7}{*}{\rotatebox[origin=c]{90}{\makecell{Pubmed}}} & GCN & $67.60$\tiny{$\pm 5.10$} & $70.87$\tiny{$\pm 5.94$} & \npboldmath{$73.79$}\tiny{$\pm 3.92$} & $67.44$\tiny{$\pm 8.55$} & \small{n/a} & $65.01$\tiny{$\pm 7.30$} & $69.23$\tiny{$\pm 5.44$} & $72.14$\tiny{$\pm 5.16$} & $66.73$\tiny{$\pm 5.80$} & \underline{$73.72$}\tiny{$\pm 3.72$} & $67.57$\tiny{$\pm 6.48$} & $67.66$\tiny{$\pm 5.04$} & $59.87$\tiny{$\pm 7.29$} & $66.75$\tiny{$\pm 8.22$} & $67.84$\tiny{$\pm 6.27$} & $68.46$\tiny{$\pm 7.13$} & $65.62$\tiny{$\pm 7.53$}\\
 & APPNP & $69.18$\tiny{$\pm 5.57$} & $68.54$\tiny{$\pm 7.37$} & \npboldmath{$74.98^\dagger$}\tiny{$\pm 4.48$} & $69.86$\tiny{$\pm 5.37$} & \small{n/a} & \small{n/a} & \small{n/a} & \small{n/a} & $69.19$\tiny{$\pm 5.81$} & \underline{$74.33$}\tiny{$\pm 3.52$} & $68.33$\tiny{$\pm 4.29$} & $69.24$\tiny{$\pm 5.44$} & $61.23$\tiny{$\pm 9.30$} & $70.86$\tiny{$\pm 5.61$} & $68.16$\tiny{$\pm 6.96$} & $68.09$\tiny{$\pm 8.57$} & $68.93$\tiny{$\pm 6.57$}\\
 & Ensemble & $66.10$\tiny{$\pm 6.96$} & $70.89$\tiny{$\pm 6.16$} & \npboldmath{$72.65$}\tiny{$\pm 3.98$} & $68.74$\tiny{$\pm 6.24$} & \small{n/a} & \small{n/a} & \small{n/a} & \small{n/a} & $68.24$\tiny{$\pm 5.75$} & $69.44$\tiny{$\pm 6.61$} & $66.11$\tiny{$\pm 6.92$} & $65.03$\tiny{$\pm 5.24$} & $60.25$\tiny{$\pm 6.79$} & \underline{$71.57$}\tiny{$\pm 6.75$} & $67.35$\tiny{$\pm 5.07$} & $66.10$\tiny{$\pm 6.61$} & $68.07$\tiny{$\pm 5.96$}\\
 & MC-Dropout & $66.45$\tiny{$\pm 4.99$} & \underline{$71.19$}\tiny{$\pm 7.79$} & $71.15$\tiny{$\pm 6.03$} & $65.50$\tiny{$\pm 4.56$} & \small{n/a} & \small{n/a} & \small{n/a} & \small{n/a} & $67.88$\tiny{$\pm 6.79$} & \npboldmath{$72.20$}\tiny{$\pm 3.46$} & $70.11$\tiny{$\pm 7.06$} & $62.32$\tiny{$\pm 7.93$} & $60.82$\tiny{$\pm 6.69$} & $61.55$\tiny{$\pm 8.42$} & $67.70$\tiny{$\pm 7.60$} & $63.82$\tiny{$\pm 7.51$} & $69.75$\tiny{$\pm 4.93$}\\
 & BGCN & $62.57$\tiny{$\pm 8.73$} & \underline{$66.70$}\tiny{$\pm 4.98$} & $63.77$\tiny{$\pm 5.06$} & $60.72$\tiny{$\pm 6.22$} & \small{n/a} & \small{n/a} & \small{n/a} & \small{n/a} & $59.02$\tiny{$\pm 9.08$} & \npboldmath{$68.97$}\tiny{$\pm 3.86$} & $58.06$\tiny{$\pm 5.41$} & $60.36$\tiny{$\pm 6.70$} & $54.49$\tiny{$\pm 6.33$} & $62.67$\tiny{$\pm 5.46$} & $58.51$\tiny{$\pm 7.89$} & $63.49$\tiny{$\pm 6.15$} & $61.25$\tiny{$\pm 7.83$}\\
 & GPN & $63.75$\tiny{$\pm 8.43$} & \small{n/a} & \small{n/a} & \small{n/a} & \small{n/a} & \small{n/a} & \small{n/a} & \small{n/a} & \underline{$66.96$}\tiny{$\pm 7.94$} & \npboldmath{$67.79$}\tiny{$\pm 5.93$} & $61.50$\tiny{$\pm 4.74$} & \small{n/a} & $54.32$\tiny{$\pm 9.25$} & $64.94$\tiny{$\pm 7.58$} & $63.31$\tiny{$\pm 8.19$} & $61.74$\tiny{$\pm 6.70$} & $65.38$\tiny{$\pm 9.24$}\\
 & SGC & $64.36$\tiny{$\pm 5.81$} & $68.78$\tiny{$\pm 3.81$} & \npboldmath{$74.51$}\tiny{$\pm 4.79$} & $65.14$\tiny{$\pm 7.04$} & $72.91$\tiny{$\pm 5.03$} & \small{n/a} & \small{n/a} & \small{n/a} & $67.71$\tiny{$\pm 5.66$} & \underline{$73.05$}\tiny{$\pm 4.78$} & $68.76$\tiny{$\pm 4.42$} & $65.76$\tiny{$\pm 4.13$} & $55.91$\tiny{$\pm 8.35$} & $70.29$\tiny{$\pm 6.11$} & $67.61$\tiny{$\pm 5.85$} & $67.19$\tiny{$\pm 7.03$} & $68.18$\tiny{$\pm 6.85$}\\
\hline \multirow{7}{*}{\rotatebox[origin=c]{90}{\makecell{AmazonPhotos}}} & GCN & $85.76$\tiny{$\pm 3.16$} & $83.37$\tiny{$\pm 2.57$} & $82.20$\tiny{$\pm 2.98$} & $87.07$\tiny{$\pm 3.40$} & \small{n/a} & $85.92$\tiny{$\pm 2.50$} & \underline{$87.72$}\tiny{$\pm 4.07$} & $79.40$\tiny{$\pm 7.22$} & $75.73$\tiny{$\pm 8.56$} & $82.28$\tiny{$\pm 3.15$} & $76.27$\tiny{$\pm 5.99$} & $81.76$\tiny{$\pm 3.61$} & \npboldmath{$88.58$}\tiny{$\pm 2.45$} & $81.39$\tiny{$\pm 7.57$} & $81.12$\tiny{$\pm 5.41$} & $86.82$\tiny{$\pm 4.51$} & $84.73$\tiny{$\pm 4.33$}\\
 & APPNP & $86.14$\tiny{$\pm 4.30$} & $87.07$\tiny{$\pm 1.60$} & $86.24$\tiny{$\pm 2.13$} & \underline{$88.49$}\tiny{$\pm 2.49$} & \small{n/a} & \small{n/a} & \small{n/a} & \small{n/a} & $81.93$\tiny{$\pm 7.54$} & $85.52$\tiny{$\pm 2.33$} & $78.44$\tiny{$\pm 6.60$} & $88.24$\tiny{$\pm 1.60$} & \npboldmath{$90.14$}\tiny{$\pm 1.73$} & $86.07$\tiny{$\pm 4.41$} & $84.31$\tiny{$\pm 4.21$} & $86.65$\tiny{$\pm 3.87$} & $85.28$\tiny{$\pm 4.15$}\\
 & Ensemble & $87.97$\tiny{$\pm 2.14$} & $84.82$\tiny{$\pm 3.48$} & $83.32$\tiny{$\pm 4.04$} & $88.10$\tiny{$\pm 3.30$} & \small{n/a} & \small{n/a} & \small{n/a} & \small{n/a} & $79.82$\tiny{$\pm 7.86$} & $83.36$\tiny{$\pm 4.01$} & $78.60$\tiny{$\pm 5.09$} & $83.51$\tiny{$\pm 3.15$} & \npboldmath{$89.96$}\tiny{$\pm 1.61$} & \underline{$89.28$}\tiny{$\pm 2.36$} & $83.92$\tiny{$\pm 6.09$} & $85.53$\tiny{$\pm 4.31$} & $87.17$\tiny{$\pm 3.75$}\\
 & MC-Dropout & $86.04$\tiny{$\pm 3.39$} & $83.37$\tiny{$\pm 3.20$} & $80.08$\tiny{$\pm 7.52$} & \underline{$87.53$}\tiny{$\pm 3.21$} & \small{n/a} & \small{n/a} & \small{n/a} & \small{n/a} & $79.07$\tiny{$\pm 6.75$} & $80.45$\tiny{$\pm 9.57$} & $72.42$\tiny{$\pm 6.13$} & $82.11$\tiny{$\pm 6.62$} & \npboldmath{$88.85$}\tiny{$\pm 2.26$} & $76.69$\tiny{$\pm 8.01$} & $78.69$\tiny{$\pm 5.88$} & $76.44$\tiny{$\pm 8.43$} & $87.09$\tiny{$\pm 2.69$}\\
 & BGCN & \underline{$79.22$}\tiny{$\pm 7.13$} & $69.90$\tiny{$\pm 9.01$} & $73.08$\tiny{$\pm 9.53$} & \npboldmath{$82.32$}\tiny{$\pm 4.67$} & \small{n/a} & \small{n/a} & \small{n/a} & \small{n/a} & $59.80$\tiny{$\pm 14.79$} & $69.34$\tiny{$\pm 10.50$} & $62.35$\tiny{$\pm 12.45$} & $76.64$\tiny{$\pm 6.02$} & $77.46$\tiny{$\pm 10.16$} & $77.81$\tiny{$\pm 8.13$} & $73.15$\tiny{$\pm 9.69$} & $78.66$\tiny{$\pm 8.01$} & $77.17$\tiny{$\pm 7.59$}\\
 & GPN & $64.59$\tiny{$\pm 8.87$} & \small{n/a} & \small{n/a} & \small{n/a} & \small{n/a} & \small{n/a} & \small{n/a} & \small{n/a} & $56.95$\tiny{$\pm 9.79$} & \underline{$71.68$}\tiny{$\pm 9.41$} & $59.87$\tiny{$\pm 9.57$} & \small{n/a} & \npboldmath{$72.46$}\tiny{$\pm 10.43$} & $55.44$\tiny{$\pm 7.09$} & $71.29$\tiny{$\pm 8.57$} & $71.33$\tiny{$\pm 9.67$} & $70.29$\tiny{$\pm 11.14$}\\
 & SGC & $87.04$\tiny{$\pm 2.72$} & $89.13$\tiny{$\pm 1.79$} & $85.20$\tiny{$\pm 2.79$} & $87.95$\tiny{$\pm 2.43$} & \npboldmath{$90.57^\dagger$}\tiny{$\pm 2.83$} & \small{n/a} & \small{n/a} & \small{n/a} & $72.18$\tiny{$\pm 8.94$} & $85.08$\tiny{$\pm 2.70$} & $72.66$\tiny{$\pm 6.71$} & $89.34$\tiny{$\pm 2.13$} & \underline{$90.52$}\tiny{$\pm 1.51$} & $90.22$\tiny{$\pm 2.48$} & $78.75$\tiny{$\pm 7.06$} & $85.71$\tiny{$\pm 3.10$} & $84.25$\tiny{$\pm 5.51$}\\
\hline \multirow{7}{*}{\rotatebox[origin=c]{90}{\makecell{AmazonComputers}}} & GCN & $77.25$\tiny{$\pm 3.74$} & $68.69$\tiny{$\pm 7.28$} & $64.17$\tiny{$\pm 9.53$} & $77.93$\tiny{$\pm 2.95$} & \small{n/a} & $76.92$\tiny{$\pm 2.00$} & \underline{$78.42$}\tiny{$\pm 2.18$} & $62.61$\tiny{$\pm 11.82$} & $67.04$\tiny{$\pm 6.67$} & $62.32$\tiny{$\pm 9.62$} & $60.49$\tiny{$\pm 8.10$} & $71.51$\tiny{$\pm 4.11$} & \npboldmath{$78.94$}\tiny{$\pm 2.69$} & $62.35$\tiny{$\pm 11.46$} & $69.54$\tiny{$\pm 8.69$} & $77.08$\tiny{$\pm 4.57$} & $77.84$\tiny{$\pm 3.28$}\\
 & APPNP & $77.87$\tiny{$\pm 2.37$} & $73.64$\tiny{$\pm 4.12$} & $72.78$\tiny{$\pm 4.66$} & $78.86$\tiny{$\pm 3.23$} & \small{n/a} & \small{n/a} & \small{n/a} & \small{n/a} & $70.73$\tiny{$\pm 6.69$} & $76.96$\tiny{$\pm 3.02$} & $69.24$\tiny{$\pm 5.44$} & $77.85$\tiny{$\pm 2.75$} & \npboldmath{$80.31$}\tiny{$\pm 2.32$} & $67.25$\tiny{$\pm 9.27$} & $71.19$\tiny{$\pm 10.07$} & $78.44$\tiny{$\pm 4.37$} & \underline{$79.87$}\tiny{$\pm 3.42$}\\
 & Ensemble & $77.88$\tiny{$\pm 2.90$} & $69.20$\tiny{$\pm 5.50$} & $66.15$\tiny{$\pm 4.00$} & $78.79$\tiny{$\pm 4.64$} & \small{n/a} & \small{n/a} & \small{n/a} & \small{n/a} & $71.33$\tiny{$\pm 3.58$} & $65.49$\tiny{$\pm 5.88$} & $66.32$\tiny{$\pm 5.77$} & $71.42$\tiny{$\pm 5.38$} & \npboldmath{$80.50$}\tiny{$\pm 2.38$} & $76.41$\tiny{$\pm 3.94$} & $73.71$\tiny{$\pm 7.31$} & $77.38$\tiny{$\pm 5.33$} & \underline{$80.38$}\tiny{$\pm 4.37$}\\
 & MC-Dropout & $75.86$\tiny{$\pm 3.93$} & $63.65$\tiny{$\pm 11.06$} & $62.67$\tiny{$\pm 8.51$} & $75.07$\tiny{$\pm 3.73$} & \small{n/a} & \small{n/a} & \small{n/a} & \small{n/a} & $70.53$\tiny{$\pm 7.71$} & $59.78$\tiny{$\pm 11.42$} & $58.54$\tiny{$\pm 7.70$} & $72.24$\tiny{$\pm 4.90$} & \npboldmath{$78.17$}\tiny{$\pm 3.71$} & $55.12$\tiny{$\pm 11.72$} & $60.21$\tiny{$\pm 11.38$} & $65.25$\tiny{$\pm 10.33$} & \underline{$77.00$}\tiny{$\pm 3.59$}\\
 & BGCN & $63.23$\tiny{$\pm 9.50$} & $39.84$\tiny{$\pm 15.30$} & $45.25$\tiny{$\pm 13.93$} & $65.87$\tiny{$\pm 9.88$} & \small{n/a} & \small{n/a} & \small{n/a} & \small{n/a} & $35.80$\tiny{$\pm 15.73$} & $45.34$\tiny{$\pm 12.66$} & $44.23$\tiny{$\pm 12.92$} & $63.64$\tiny{$\pm 7.52$} & $62.32$\tiny{$\pm 13.64$} & \npboldmath{$67.87$}\tiny{$\pm 10.76$} & $53.71$\tiny{$\pm 13.66$} & \underline{$67.06$}\tiny{$\pm 8.51$} & $64.32$\tiny{$\pm 11.97$}\\
 & GPN & $59.89$\tiny{$\pm 13.72$} & \small{n/a} & \small{n/a} & \small{n/a} & \small{n/a} & \small{n/a} & \small{n/a} & \small{n/a} & $32.00$\tiny{$\pm 8.75$} & $60.63$\tiny{$\pm 10.59$} & $46.42$\tiny{$\pm 16.56$} & \small{n/a} & \npboldmath{$64.79$}\tiny{$\pm 10.60$} & $45.85$\tiny{$\pm 13.64$} & $45.95$\tiny{$\pm 14.32$} & $63.25$\tiny{$\pm 3.79$} & \underline{$64.24$}\tiny{$\pm 6.71$}\\
 & SGC & $78.35$\tiny{$\pm 2.44$} & \underline{$80.98$}\tiny{$\pm 2.86$} & $73.01$\tiny{$\pm 4.68$} & $78.49$\tiny{$\pm 2.34$} & $80.69$\tiny{$\pm 2.65$} & \small{n/a} & \small{n/a} & \small{n/a} & $64.92$\tiny{$\pm 0.75$} & $75.69$\tiny{$\pm 3.56$} & $72.31$\tiny{$\pm 3.54$} & $78.24$\tiny{$\pm 3.16$} & \npboldmath{$81.19^\dagger$}\tiny{$\pm 2.25$} & $68.70$\tiny{$\pm 8.90$} & $67.44$\tiny{$\pm 10.26$} & $73.46$\tiny{$\pm 6.06$} & $78.38$\tiny{$\pm 4.05$}\\
\Xhline{3\arrayrulewidth}
\end{tabular}

}
\end{sidewaystable}

\begin{figure}[ht]
    \centering
    \begin{subfigure}{.49\textwidth}
        \centering
\input{files/pgfs/galaxy_badge/cora_ml_galaxy_badge.pgf}
      \caption{\textcolor{black}{CoraML.}}
    \end{subfigure}
     \hfill
     \begin{subfigure}{.49\textwidth}
       \centering
 
\input{files/pgfs/galaxy_badge/citeseer_galaxy_badge.pgf}
       \caption{\textcolor{black}{Citeseer.}}
     \end{subfigure}
    \begin{subfigure}{.49\textwidth}
        \centering

\input{files/pgfs/galaxy_badge/pubmed_galaxy_badge.pgf}
      \caption{\textcolor{black}{PubMed.}}
    \end{subfigure}
     \hfill
     \begin{subfigure}{.49\textwidth}
       \centering
 
\input{files/pgfs/galaxy_badge/amazon_photos_galaxy_badge.pgf}
       \caption{\textcolor{black}{Amazon Photos.}}
     \end{subfigure}
     \begin{subfigure}{.49\textwidth}
       \centering
 
\input{files/pgfs/galaxy_badge/amazon_computers_galaxy_badge.pgf}
       \caption{\textcolor{black}{Amazon Computers.}}
     \end{subfigure}
    \caption{\textcolor{black}{US using the i.i.d. GALAXY and BADGE baselines versus our approach (ESP) and random sampling.}}
    \label{fig:galaxy_badge}
\end{figure}

\section{Computing Aleatoric and Total Confidence on CSBMs}\label{appendix:computing_alea_and_total_on_csbms}

Computing the aleatoric confidence is straightforward directly employing \Cref{def:conf_alea} using Bayes rule and the generative process up to a tractable normalization constant.

\begin{equation}
    \confidence^\alea(i,c) \propto p(\matr A, \matr X \mid \vect y_{-i}^\gt, \vect y_i=c) p(\vect y_i = c)
\end{equation}

Computing the epistemic confidence, however, turns out to have exponential complexity in the number of unlabeled nodes $\lvert \unlabeled-i\rvert$:

\begin{align}
    \confidence^\epi(i,c) &= p(\vect y_i = c \mid \matr A, \matr X, \vect y_\labeled^\gt) \\
    &= \sum_{\vect y_{\unlabeled - i}} p(\vect y_i = c = \vect y_{\unlabeled - 1} \mid \matr A, \matr X, \vect y_\labeled^\gt)\\
    &\propto \sum_{\vect y_{\unlabeled - i} } p(\matr A, \matr X \mid \vect y_i = c, \vect y_{\unlabeled - i}, \vect y_\labeled^\gt) p (\vect y_i = c, \vect y_{\unlabeled - i})
\end{align}

\textbf{Approximating Epistemic Confidence}. For larger graphs, this quickly becomes intractable and we therefore rely on a variational mean-field approximation of the joint distribution $p(\vect y_\unlabeled \mid \matr A, \matr X, \vect y_\labeled)$ to obtain marginals similar to \cite{mariadassou2010uncovering, jaakkola2000tutorial}.

\begin{equation}
    p(\vect y_\unlabeled \mid \matr A, \matr X, \vect y_\labeled) \approx q(\vect y_\unlabeled) := \prod_{i \in \unlabeled}q_i(\vect y_i) 
\end{equation}

Since the variational distributions $q_i$ are discrete, we can fully describe them with parameters $\gamma_{i,c} := q_i(\vect y_i = c)$. The ELBO of this variational problem is given by:

\begin{align}
    J(\gamma) &= \log p(\matr X, \matr A, \vect y_\labeled) + \kl[\bigg]{q(\vect y_\unlabeled)}{p(\vect y_\unlabeled \mid \matr A, \matr X, \vect y_\labeled)}\\
    &= \E_{\vect y_\unlabeled \sim q}\left[\log p(\matr A, \matr X, \vect y_\unlabeled \mid \vect y_\labeled)\right] +\mathbb{H}\left[q(\vect y_\unlabeled)\right] + \text{const} \\ 
    &= \sum_{i,c} \gamma_{i,c} \log p(\vect y_i = c) + \sum_{i < j} \sum_{c, c^\prime} \gamma_{i,c} \gamma_{j, c^\prime} \log p(\matr A_{i,j} \mid \vect y_i = c, \vect y_j = c^\prime) \\
    &+ \sum_{i, c} \gamma_{i,c} \log p(\matr X_i \mid \vect y_i = c)+ \text{const}
\end{align}

Here, we introduced $\gamma_{j,c} = \mathbbm{1}(c = \vect y_j^\gt)$ for all $j \in \labeled$ for convenience. This gives rise to a constrained optimization problem where $\sum_c \gamma_{i,c} = 1$ for all $i \in \unlabeled$. Solving this problem analytically gives rise to the equations:

\begin{equation}
    \gamma_{i,c} \propto \exp \left(\log p(\vect y_i = c) + \sum_{j \neq i}\sum_{c^\prime} \gamma_{j,c^\prime} \log p(\matr A_{i,j} \mid \vect y_i = c, \vect y_j = c^\prime) + \log p(\matr X_i \mid \vect y_i = c) \right)
\end{equation}

The marginal probabilities $\gamma_{i,c}$ can then directly be optimized by the resulting fixed point iteration scheme, where after each iteration the probabilities are normalized:

\begin{equation}
    \log \gamma_{i,c}^{(t+1)} = \log p(\vect y_i = c) + \sum_{j \neq i}\sum_{c^\prime} \gamma_{j,c^\prime}^{(t)} \log p(\matr A_{i,j} \mid \vect y_i = c, \vect y_j = c^\prime) + \log p(\matr X_i \mid \vect y_i = c)
    \label{eq:total_confidence_fixed_point_iteration}
\end{equation}

The quality of the acquisition function given by the epistemic confidence relies on the quality of the approximated marginals $p(\vect y_i \mid \matr A, \matr X, \vect y_\labeled)$, i.e. the total confidence $\confidence^\total(i, \cdot)$. To verify that the proposed variational scheme indeed provides reasonable approximations, we report the absolute approximation errors $\lvert q(\vect y_i = c) - p(\vect y_i = c \mid \matr A, \matr X, \vect y_\labeled)\rvert$ for CSBM graphs such that exact computation of the true marginals is tractable, i.e. up to $12$ nodes.

\begin{figure}[ht]
    \centering
    \input{files/pgfs/sbm/absolute_approximation_error.pgf}
    \caption{Absolute error distribution between approximate total confidence $q(\vect y_i)$ and true total confidence $p(\vect y_i\mid \matr A, \matr X, \vect y_\labeled)$ for graphs of different sizes.}
    \label{fig:approximation_error}
\end{figure}

\Cref{fig:approximation_error} shows the distribution of approximation errors averaged over five different samples from a CSBM generative distribution with a structural SNR of $\sigma_A = 2.0$ and a feature SNR $\sigma_X = 1.0$ as well as an expected node degree $\E\left[\text{deg}(v)\right] = 4.0$. In general, we expect the approximation to be of higher quality the more decoupled the marginals $p(\vect y_i \mid \matr A, \matr X, \vect y_\labeled)$ are \citep{jaakkola2000tutorial}. While the median error does not exceed $5\%$, we observe some outliers for larger graphs. In such cases, the employed approximation is inaccurate. Nonetheless, in \Cref{fig:al_csbm} we observe that even in the face of sometimes poor approximations, the proposed uncertainty framework achieves strong results. We suspect a stronger approximation to perform even better.

\section{Uncertainty Sampling with Ground-Truth Uncertainty}\label{appendix:us_with_gnd}

\begin{figure}[h!]
    \centering
    \begin{subfigure}[t]{0.49\textwidth}
        \centering
\input{files/pgfs/sbm/csbm_bayes_1000.pgf}
      \caption{AL using ground-truth uncertainty.}
    \end{subfigure}
    \begin{subfigure}[t]{0.49\textwidth}
        \centering
\input{files/pgfs/sbm/csbm_baselines_1000.pgf}
      \caption{AL using contemporary US strategies.}
    \end{subfigure}
    \caption{US on a CSBM with 1000 nodes and 4 classes. Ground-truth epistemic uncertainty significantly outperforms other estimators and random queries. Contemporary US can not outperform random sampling.}
    \label{fig:al_csbm_1000}
\end{figure}

In \Cref{fig:al_csbm_1000}, we supplement our findings from \Cref{fig:al_csbm} for graphs with $1000$ nodes and $4$ classes sampled from the same CSBM distribution. Again, we find epistemic US to be the strongest approach in terms of the Bayesian classifier while total uncertainty can not match its performance but also outperforms random sampling. We also observe a decrease in performance when the generative process is modelled incorrectly. Also contemporary estimators fail to outperform random queries, confirming the results reported in \Cref{sec:gnd_csbm} for larger graph sizes.

\begin{figure}[h!]
    \centering
    \begin{subfigure}[t]{0.49\textwidth}
        \centering
\input{files/pgfs/sbm/csbm_baselines_non_uq_100.pgf}
    \end{subfigure}
    \caption{{Performance of traditional AL strategies on a CSBM with 100 nodes and 7 classes.}}
    \label{fig:al_Csbm_non_uq_baselines}
\end{figure}

\subsection{Ablation of the Proposed Acquisition Strategy on Different CSBM Configurations}\label{appendix:csbm_grid}

We also ablate the proposed acquisition strategy on different configurations of the CSBM. That is, we sample graphs from a CSBM with homogeneous and symmetric affiliation matrices and vary both the structural SNR $\frac{p}{q}$ as well as the feature SNR $\frac{\delta_X}{\sigma_X}$. The underlying CSBM has $100$ nodes and $4$ classes and labels are sampled from a uniform prior (see \Cref{appendix:datasets}).

We perform AL on five graphs sampled from each configuration independently and measure the absolute improvement Uncertainty Sampling achieves in terms of AUC and accuracy after a budget of $5C=20$ is exhausted. In \Cref{fig:sbm_grid_auc}, we show the average improvement in AUC for each SNR configuration. We also supply a similar visualization for the accuracy after the budget is exhausted in \Cref{fig:sbm_grid_acc}.

\begin{figure*}[h!]
    \centering
    \begin{subfigure}[t]{0.46\textwidth}
        \centering
        \input{files/pgfs/sbm/sbm_grid_100_delta_auc.pgf}
        \caption{Improvement in AUC}
        \label{fig:sbm_grid_auc}
    \end{subfigure}%
    ~ 
    \begin{subfigure}[t]{0.46\textwidth}
        \centering
        \input{files/pgfs/sbm/sbm_grid_100_delta_accuracy.pgf}
        \caption{Improvement in test accuracy}
        \label{fig:sbm_grid_acc}
    \end{subfigure}
    \caption{Evaluating the absolute improvement of Uncertainty Sampling using epistemic uncertainty over random acquisition for different structure and feature SNRs.}
\end{figure*}

We find that our approach achieves the strongest improvement in both metrics when the structural SNR $\sigma_A$ is neither too high nor too small: Large values make the classification problem too easy and hence AL strategies do not have to carefully query informative nodes as the overall performance is strong even in very label scare regimes. Interestingly, our ground-truth uncertainty estimator shows strong merit when the node features are noisy, indicating that it is crucial to pick structurally informative nodes in the graph in these regimes.

At the same time, when the structural SNR is too low (in particular drops below $1.0$), our method fails to outperform random acquisition: We attribute this to the mean-field approximation to the total confidence $\confidence^\total$ described in \Cref{appendix:computing_alea_and_total_on_csbms}: We observed that for graphs with low structural SNR, the fixed point iteration of \Cref{eq:total_confidence_fixed_point_iteration} does not converge. Hence, the approximated marginal probabilities are poor and both the prediction as well as the uncertainty estimation based on it deteriorate.

\section{Visualization of Ground-Truth Uncertainty on a Toy Example}

\begin{figure}[H]
    \centering
    \begin{subfigure}[t]{0.49\textwidth}
        \centering
        \includegraphics[width=0.9\textwidth]{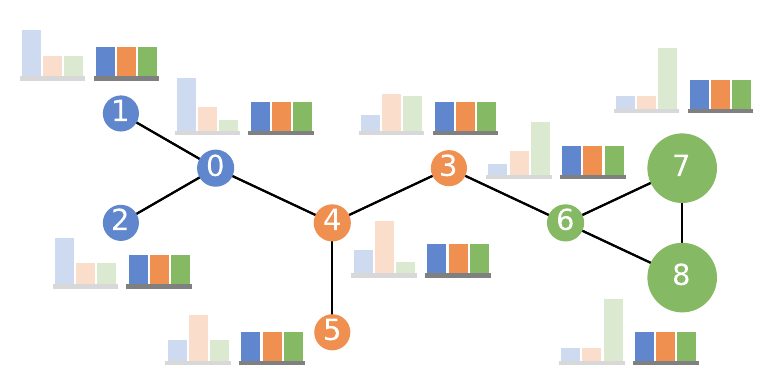}
        \caption{Initial state}
        \label{fig:toy_uniform_step_0}
    \end{subfigure}%
    ~ 
    \begin{subfigure}[t]{0.49\textwidth}
        \centering
        \includegraphics[width=0.9\textwidth]{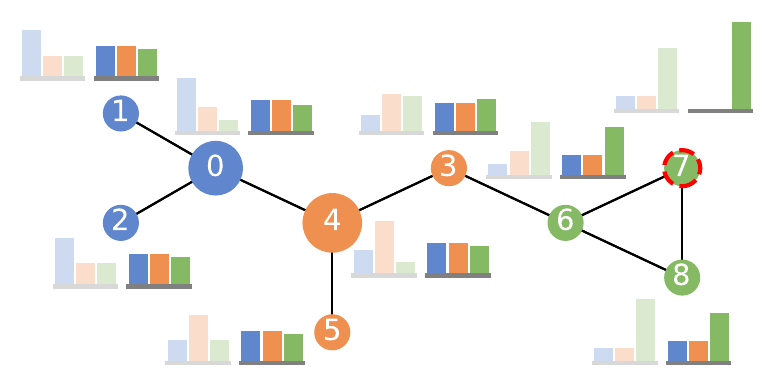}
        \caption{State after first acquisition}
        \label{fig:toy_uniform_step_1}
    \end{subfigure}
    \caption{Example of CSBM with 3 classes and 9 nodes. \textbf{Left}, we can see the initial state without any labeled node. On the \textbf{right}, we can see the state after we acquired node 7.}
\end{figure}

In the following, we illustrate the behavior of this acquisition function on a small CSBM graph. Figure \ref{fig:toy_uniform_step_0} shows a sample graph with three classes and nine nodes. The greyed-out histograms represent the aleatoric confidence. The distributions correspond to the total confidence, and the size of the nodes indicates the inverse of the ratio of both, i.e., the epistemic uncertainty, which guides our acquisition. We note that nodes $7$ and $8$ are the most promising candidates, as their aleatoric prediction is confident, and in the initial step since all nodes are unlabeled, the total confidence is uniform across all nodes due to the homogenous structure of the affiliation matrix.

\begin{figure}[H]
\centering
\includegraphics[width=0.25\linewidth]{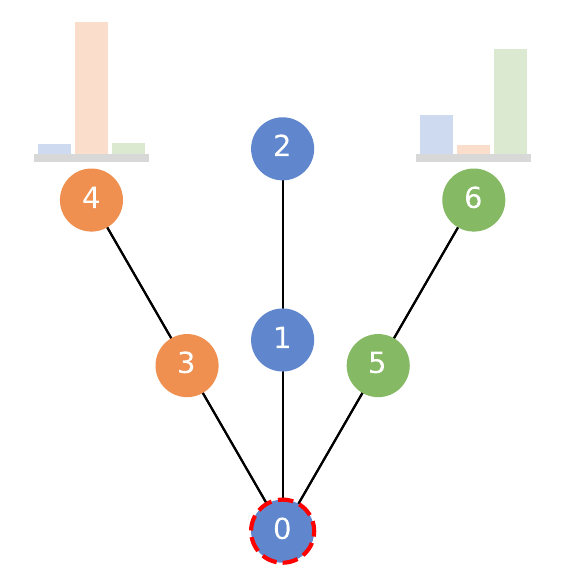}
\caption{Aleatoric uncertainty is different for symmetric nodes when the affiliation matrix is not symmetric itself.}
\label{fig:toy_alea_inhomo}
\end{figure}%
Figure \ref{fig:toy_uniform_step_1} depicts the subsequent iteration following the acquisition of node $7$. Due to the additional information introduced, node $8$ exhibits reduced epistemic uncertainty. Consequently, the most promising nodes to consider next are nodes $0$ and $4$, as their total confidence remains relatively low, and they display higher aleatoric confidence than their neighbors. This is because both nodes connect to two nodes from the same class, bolstering their confidence levels.

\begin{figure}[H]
    \centering
    \begin{subfigure}[t]{0.49\textwidth}
        \centering
        \includegraphics[width=0.8\textwidth]{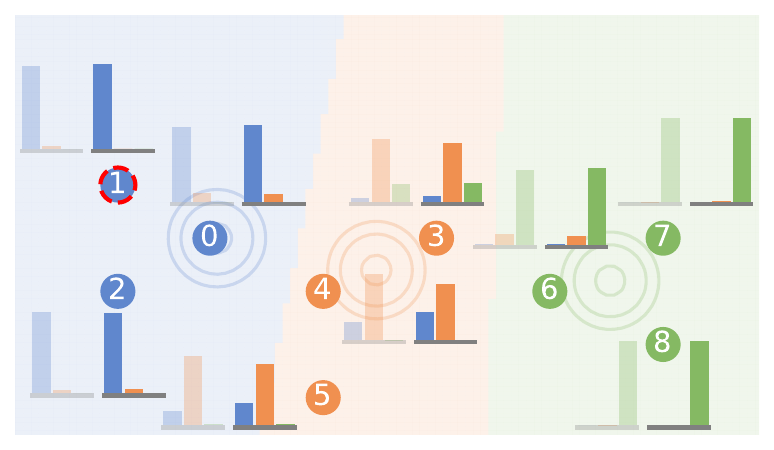}
        \caption{Decision boundary given only the features}
        \label{fig:toy_features_before}
    \end{subfigure}%
    ~ 
    \begin{subfigure}[t]{0.49\textwidth}
        \centering
        \includegraphics[width=0.8\textwidth]{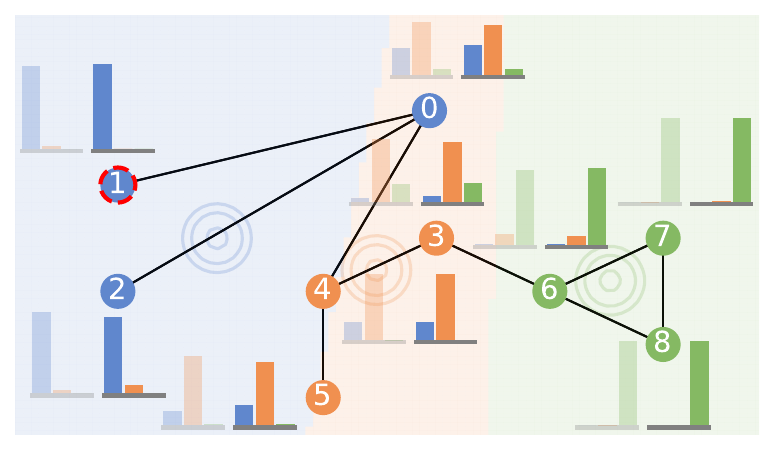}
        \caption{Decision boundary given the features \textbf{and} structure}
        \label{fig:toy_features_after}
    \end{subfigure}
    \caption{Example of CSBM with one labeled node ($1$). The shaded area represents the decision of the classifier. \textbf{Left}, only using the feature information. \textbf{Right} feature and structural information.}
\end{figure}

Figure \ref{fig:toy_features_before} presents predictions made solely on feature information, with shaded regions and concentric circles demonstrating decision boundaries and class feature distributions, respectively. It is evident that incorporating feature information significantly increases confidence. Then, in Figure \ref{fig:toy_features_after}, we adjust the position of node $0$ away from the blue center and introduce the structural information. The shaded regions in this figure represent the prediction for node $0$ and a noticeable shift to the right compared to Figure \ref{fig:toy_features_before} can be observed. This shift demonstrates the impact of structural information on the classifier's decision boundary.

Finally, Figure \ref{fig:toy_alea_inhomo} underscores the significance of modeling non-existing edges in uncertainty estimation. Despite their symmetric neighborhoods, the aleatoric uncertainty (without feature information) differs for nodes 4 and 6 as in the underlying CSBM model the green class is less likely to not associate with the blue class. This reveals that sensible uncertainty estimators need to also utilize the \textit{absence of edges}.

{\section{Approximating Disentangled Uncertainty on Real-World Data}\label{appendix:gt_approximation}}

\subsection{Multiple Pseudo-Labels (MP)}

\Cref{eq:epi_as_ratio} computes epistemic confidence as a ratio of total and aleatoric confidences. For a given classifier $f_\theta(\matr A, \matr X)$ that should be improved with AL, we interpret its predictive distribution as total confidence which is defined in \Cref{def:conf_total}. 

 \begin{equation}
    {\confidence^\total(i, c) = \mathbb{E}_{p({{\theta}} \mid \matr A, \matr X, \vect y_{\labeled}^\gt)} \left[ \prob{\vect y_i = c \mid \matr A, \matr X, \vect y_{\labeled}=\vect y_{\labeled}^\gt, {{\theta}}} \right]
    \approx f_\theta(\matr A, \matr X)_{i, c}}
 \end{equation}

 Note that total confidence is defined as the marginal predictive distribution associated with a node $i$ after conditioning on the training set $\vect y_\labeled = \vect y_\labeled^\gt$. This conditioning is represented by the posterior over the model parameters $p({{\theta}} \mid \matr A, \matr X, \vect y_{\labeled}^\gt)$ that is obtained through training the classifier. In the case of deterministic GNNs, this posterior collapses to a point estimate.

 Estimating the aleatoric confidence defined in \Cref{def:conf_alea} is more challenging as the predictive distribution of a node $i$ needs to be conditioned on unavailable ground-truth labels $\vect y_{\unlabeled - i}^\gt$. We approximate those by using the predictions of $f_\theta$ as pseudo-labels:

 \begin{equation}
     \hat{\vect y}_j = \argmax_{c \in [C]} f_\theta(\matr A, \matr X)_{j, c}
 \end{equation}

 Similar to the approximation for total confidence, we employ the predictive distribution of a classifier to estimate aleatoric confidence. Since the set of labels that this classifier must be conditioned on the pseudo-labels $\hat{\vect y}_{\unlabeled - i}$ as well, a separate classifier must be trained. For each node $i$ that the aleatoric confidence should be predicted, we get a different set of labels the classifier needs to be trained on: $\vect y_\labeled^\gt \cup \hat{\vect y}_{\unlabeled - i}$. This implies that $\mathcal{O}(n)$ auxiliary classifiers need to be trained in each iteration of AL using this approximate method. For each classifier $f_{\hat{\theta}_i}$, we use its predictive distribution as an estimate of aleatoric confidence:

 \begin{equation}
     {\confidence^\alea(i, c) = \mathbb{E}_{p({\hat{\theta}_i} \mid \matr A, \matr X, \vect y_{\labeled}^\gt, \hat{\vect y}_{\unlabeled - i})} \left[ \prob{\vect y_i = c \mid \matr A, \matr X, \vect y_{\labeled}=\vect y_{\labeled}^\gt, \vect y_{\unlabeled - i} = \hat{\vect y}_{\unlabeled - i},{\hat{\theta}}} \right]
    \approx f_{\hat{\theta}_i}(\matr A, \matr X)_{i, c}}
    \label{eq:lhs_alea}
 \end{equation}

Again, the posterior $p({\hat{\theta}_i} \mid \matr A, \matr X, \vect y_{\labeled}^\gt, \hat{\vect y}_{\unlabeled - i})$ over the weights of the auxiliary classifier are obtained through training on both available ground-truth labels as well as pseudo labels. This implies that each auxiliary classifier $f_{\hat{\theta}_i}$ is trained on $n - 1$ labeled nodes. To get a scalar estimate of epistemic uncertainty that serves as an acquisition proxy, we take the ratio of both approximations according to \Cref{eq:epi_as_ratio} and evaluate it at $c = \hat{\vect y}_i$ as the true class label of node $i$ is not available:

\begin{equation}
    \confidence^\epi(i, \hat{\vect y}_i) = \frac{\confidence^\alea(i, \hat{\vect y}_i)}{\confidence^\total(i, \hat{\vect y}_i)}
\end{equation}

When employing the MP approximation as a strategy, in each iteration, we query the label of node $i^* = \argmax_{i \in \unlabeled} \confidence^\epi(i, \hat{\vect y}_i)$ that maximizes (approximate) epistemic uncertainty.

\subsection{Expected Single Pseudo-Label (ESP)}\label{sec:esp}

An alternative approach is to not approximate epistemic uncertainty as a ratio of total and aleatoric confidences but directly estimate the right-hand side of \Cref{lem:acquisition} which we prove to be equivalent. Both in the numerator and denominator, we need to compute joint probabilities over $\vect y_{\unlabeled - i}$ which we approximate as a product of marginal probabilities predicted by the classifier. To that end, the joint probability in the denominator is conditioned on $\vect y_\labeled$ only, so we again can make use of the predictive distribution of the classifier $f_\theta$ to be trained. As the denominator of \Cref{lem:acquisition} requires evaluating the joint probability at the unavailable ground-truth labels $\vect y_{\unlabeled - i}^\gt$, we again have to rely on the pseudo-labels $\hat{\vect y}$ predicted by $f_\theta$.

\begin{equation}
    \begin{split}
    \prob{\vect y_{\unlabeled - i} = \vect y_{\unlabeled - i}^\gt \mid \matr A, \matr X, \vect y_\labeled
} &\approx \prob{\vect y_{\unlabeled - i} = \hat{\vect y}_{\unlabeled - i} \mid \matr A, \matr X, \vect y_\labeled} \\
&\approx \prod_{j \in \unlabeled - i} f_\theta(\matr A, \matr X)_{j, \hat{\vect y}_j} \\
&= \prod_{j \in \unlabeled - i} f_\theta(\matr A, \matr X)_{j, \hat{\vect y}_j} \frac{f_\theta(\matr A, \matr X)_{i, \hat{\vect y}_i}}{f_\theta(\matr A, \matr X)_{i, \hat{\vect y}_i}} \\
&= \frac{\prod_{j \in \unlabeled} f_\theta(\matr A, \matr X)_{j, \hat{\vect y}_j}}{f_\theta(\matr A, \matr X)_{i, \hat{\vect y}_i}} \\
&\propto \frac{1}{f_\theta(\matr A, \matr X)_{i, \hat{\vect y}_i}}
\end{split}
\label{eq:approximating_rhs_denom}
\end{equation}

In the last line of \Cref{eq:approximating_rhs_denom}, we recognized that the term $\prod_{j \in \unlabeled} f_\theta(\matr A, \matr X)_{j, \hat{\vect y}_j}$ is independent of node $i$ which epistemic uncertainty is to be approximated for. Since for AL, we query a node that maximizes epistemic uncertainty, we can discard constant multiplicative factors that are the same for the computation of each node.

For any node $i$, approximating the numerator of \Cref{lem:acquisition} requires conditioning on its unavailable true label $\vect y_i^\gt$. One possible remedy would be to again use the pseudo-label $\hat{\vect y}_i$ instead. However, since we need to only condition on one additional approximate label, it is also feasible to factor in the belief of $f_\theta$ about this label more accurately by taking an expectation over all possible realizations $c \in [C]$ with respect to the predictive distribution $f_\theta(\matr A, \matr X)_{i, :}$. In contrast, when computing MP approximation using \Cref{eq:lhs_alea}, we condition on all $\vect y_{\unlabeled - i}$ simultaneously: Taking an expectation with respect to the belief of the classifier $f_\theta$ would thus require $\mathcal{O}(C^{\vert\unlabeled\vert})$ evaluations, each of which involves training a surrogate model from scratch, which is intractable. 

Similar to computing the MP approximation, we emulate conditioning on additional observations by training a separate classifier on augmented data. That is, for conditioning on the observation $\vect y_i = c$, we train a model $f_{\hat{\theta}_{i, c}}$ and interpret its predictive distribution as the marginals of the joint distribution we try to approximate. Plugging everything together, we can estimate the numerator of \Cref{lem:acquisition} for a node $i$ as follows:

\begin{equation}
    \begin{split}
    \prob{\vect y_{\unlabeled - i} = \vect y_{\unlabeled - i}^\gt \mid \matr A, \matr X, \vect y_\labeled, \vect y_i = \vect y_i^\gt} &\approx \mathbb{E}_{c \sim f_\theta(\matr A, \matr X)_{i, :}} \left[\prob{\vect y_{\unlabeled - i} = \vect y_{\unlabeled - i}^\gt \mid \matr A, \matr X, \vect y_\labeled, \vect y_i = c}\right] \\
    &= \sum_c \prob{\vect y_{\unlabeled - i} = \vect y_{\unlabeled - i}^\gt \mid \matr A, \matr X, \vect y_\labeled, \vect y_i = c} f_\theta(\matr A, \matr X)_{i, c} \\
    & \approx \sum_c \left(\prod_{j \in \unlabeled - i} f_{\hat{\theta}_{i,c}}(\matr A, \matr X)_{j, \Tilde{\vect y}_j^{(i, c)}} \right) f_\theta(\matr A, \matr X)_{i, c}
\end{split}
\label{eq:approximating_rhs_num}
\end{equation}

Note that here we approximate $\vect y_{\unlabeled - i}^\gt$ not with the pseudo-labels of $f_\theta$, but instead use the pseudo-labels of the surrogate classifier $f_{\hat{\theta}_{i,c}}$ which are defined as:

\begin{equation}
    \Tilde{\vect y}_{j}^{(i, c)} := \argmax_{c \in [C]} f_{\hat{\theta}_{i, c}}(\matr A, \matr X)_{j, c}
\end{equation}

We can now combine both \Cref{eq:approximating_rhs_denom,eq:approximating_rhs_num} and compute an estimate of the epistemic uncertainty associated with a node $i$:

\begin{equation}
    \begin{split}
        \uncertainty^\epi(i, \vect y_i^\gt) &= \frac{\prob{\vect y_{\unlabeled - i} = \vect y_{\unlabeled - i}^\gt \mid \matr A, \matr X, \vect y_\labeled, \vect y_i = \vect y_i^\gt}}{\prob{\vect y_{\unlabeled - i} = \vect y_{\unlabeled - i}^\gt \mid \matr A, \matr X, \vect y_\labeled
}} \\
&\approx \frac{\sum_c \left(\prod_{j \in \unlabeled - i} f_{\hat{\theta}_{i,c}}(\matr A, \matr X)_{j, \Tilde{\vect y}_j^{(i, c)}} \right) f_\theta(\matr A, \matr X)_{i, c}}{\frac{\prod_{j \in \unlabeled} f_\theta(\matr A, \matr X)_{j, \hat{\vect y}_j}}{f_\theta(\matr A, \matr X)_{i, \hat{\vect y}_i}}} \\
&= \frac{\sum_c \left(\prod_{j \in \unlabeled - i} f_{\hat{\theta}_{i,c}}(\matr A, \matr X)_{j, \Tilde{\vect y}_j^{(i, c)}} \right) f_\theta(\matr A, \matr X)_{i, c}}{\prod_{j \in \unlabeled} f_\theta(\matr A, \matr X)_{j, \hat{\vect y}_j}}{f_\theta(\matr A, \matr X)_{i, \hat{\vect y}_i}} \\
&\propto \sum_c \left(\prod_{j \in \unlabeled - i} f_{\hat{\theta}_{i,c}}(\matr A, \matr X)_{j, \Tilde{\vect y}_j^{(i, c)}} \right) f_\theta(\matr A, \matr X)_{i, c} f_\theta(\matr A, \matr X)_{i, \hat{\vect y}_i}
    \end{split}
\end{equation}

\subsection{Implementation}

\textbf{Efficiency}. Both MP and ESP algorithms require training auxiliary classifiers on augmented datasets in order to estimate probabilities conditioned on unavailable observations. In each iteration of AL, the MP approximation requires training $\vert \unlabeled \vert \in \mathcal{O}(n)$ classifiers. The ESP approximation additionally takes an expectation over the unobserved class of each node, resulting in the training of $\mathcal{O}(nc)$ additional models per query. Notably, the auxiliary models used in the MP approximation are trained on $n - 1$ labels while the classifiers $f_{\hat{\theta}_{i, c}}$ the RP algorithm relies are only trained on $\vert \labeled \vert + 1$ labels. Therefore, even though the complexity of the MP is better in theory, in practice the runtime of the ESP approximation is significantly shorter than the MP paradigm. In fact, the MP approximation did not finish an AL run within 72 hours for two datasets (see \Cref{tab:auc_ours}).

{\textbf{Backbone Architecture and Training.} Because of the aforementioned efficiency limitations, we use an SGC model as the backbone classifier for our proposed framework. Effectively, SGC is a logistic regression model fit to diffused node features $\matr X$. We use the SAGA solver which is efficient for larger datasets to approximate the aleatoric uncertainty of the ESP as it uses pseudo-labels of all nodes, and we rely on liblinear in all other cases. Furthermore, we account for class imbalances and use a regularization weight of $\lambda = 1.0$. To mimic the GNNs used by other baselines, we diffuse the node features $\matr X$ for $2$ iterations. While acquisition requires training and evaluating auxiliary models $f_{\hat{\theta}}$ on pseudo-labels $\hat{\vect y}$ or $\tilde{\vect y}$ for both approximation frameworks, we only train the underlying classifier $f_\theta$ that we report numbers on using ground-truth labels iteratively revealed by the oracle.}

\textbf{Discussion.} We also point out reasons for the proposed method to not be at its full efficacy yet due to the various assumptions and approximations we make. Specifically, different sources of error in estimating epistemic uncertainty can stem from
\begin{inparaenum}[(i)]
    \item The classifiers $f_\theta$ and $f_{\hat{\theta}_{i,c}}$ may not faithfully model the true generative process, as described in \Cref{sec:gnd_csbm}, which results in suboptimal performance.
    \item The pseudo-labels $\hat{\vect y}$ may not match the true labels $\vect y$, which in turn leads to errors in approximating aleatoric confidence and querying epistemic confidence at the correct label.
    \item The classifiers may be poorly calibrated, a tendency exhibited by some GNN architectures \citep{hsu2022makes}. In fact, we observed similar experiments using a GCN instead of an SGC as a backbone to be unsuccessful.
\end{inparaenum}
While both MP and ESP approximation paradigms aim to estimate the same quantity, they rely on different assumptions and approximations. Therefore, it is expected that they behave differently in practice. Nonetheless, they both are indicative of the potential properly disentangled uncertainty brings to AL. Indeed, as \Cref{tab:auc_ours} verifies even under various approximations disentangling uncertainty greatly improves US. We suspect the ESP to perform significantly better because it relies less on pseudo-labels: It factors in the belief of the classifier $f_\theta$ more accurately and only trains surrogate models on one unobserved class label. 

\begin{figure}
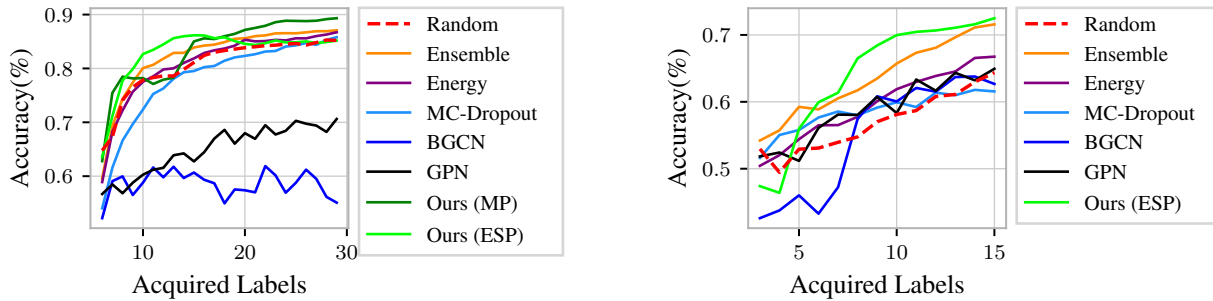
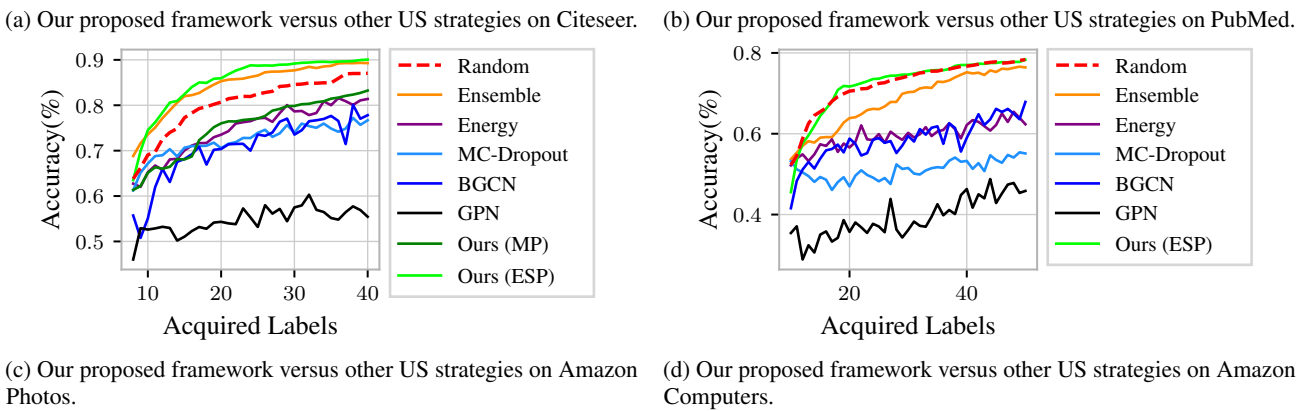

    \centering
    \begin{subfigure}[t]{0.49\textwidth}
        \centering
\input{files/pgfs/ours/citeseer_ours_vs_baselines.pgf}
      \caption{{Our proposed framework versus other US strategies on Citeseer.}}
      \label{fig:ours_citeseer}
    \end{subfigure}
    \hfill
    \begin{subfigure}[t]{0.49\textwidth}
        \centering
\input{files/pgfs/ours/pubmed_ours_vs_baselines.pgf}
      \caption{{Our proposed framework versus other US strategies on PubMed.}}
      \label{fig:ours_pubmed}
    \end{subfigure}
    
    \begin{subfigure}[t]{0.49\textwidth}
        \centering
\input{files/pgfs/ours/amazon_photos_ours_vs_baselines.pgf}
      \caption{{Our proposed framework versus other US strategies on Amazon Photos.}}
      \label{fig:ours_amazon_photos}
    \end{subfigure}
    \hfill
    \begin{subfigure}[t]{0.49\textwidth}
        \centering
\input{files/pgfs/ours/amazon_computers_ours_vs_baselines.pgf}
      \caption{{Our proposed framework versus other US strategies on Amazon Computers.}}
      \label{fig:ours_amazon_computers}
    \end{subfigure}
    \caption{{Our proposed uncertainty disentanglement framework applied to an SGC classifier using the MP or ESP approximation.}}
\end{figure}

\Cref{fig:ours_cora_ml,fig:ours_citeseer,fig:ours_pubmed,fig:ours_amazon_computers,fig:ours_amazon_photos} showcase the practical applicability of this framework. We compare both MP and ESP approximations to the best-performing US and non-US AL strategies. MP outperforms other uncertainty estimators on Citeseer, while showing worse-than-random performance on Amazon Photos. The ESP approximation consistently yields strong results that outperform other epistemic uncertainty estimators. This underlines that disentangled epistemic uncertainty, in many instances, has the potential to be an effective guide for US. Future work can build upon the results of our analysis by, for example, finding more sensible and efficient approximations to uncertainty disentangling and equipping estimators with the capabilities to describe a broader family of data-generating processes.

\subsection{\textcolor{black}{Approximating Ground-Truth Uncertainty without Graph Inductive Biases}\label{appendix:gt_features_only}}

\textcolor{black}{We verify the impact of considering network effects when approximating ground-truth uncertainty to apply the proposed algorithm. Again, we use an SGC backbone and compute uncertainty using the aforementioned ESP approach, see \Cref{sec:esp}. We ablate estimating the marginal probabilities from the classifier that is aware of the network (by diffusing the input features) against predictions from the same model that ignores the graph. Notably, we only omit network effects when approximating uncertainty. Both training and evaluation is done with consideration for instance interdependence to enable a fair comparison between the two US strategies.
}

\begin{figure}[ht]
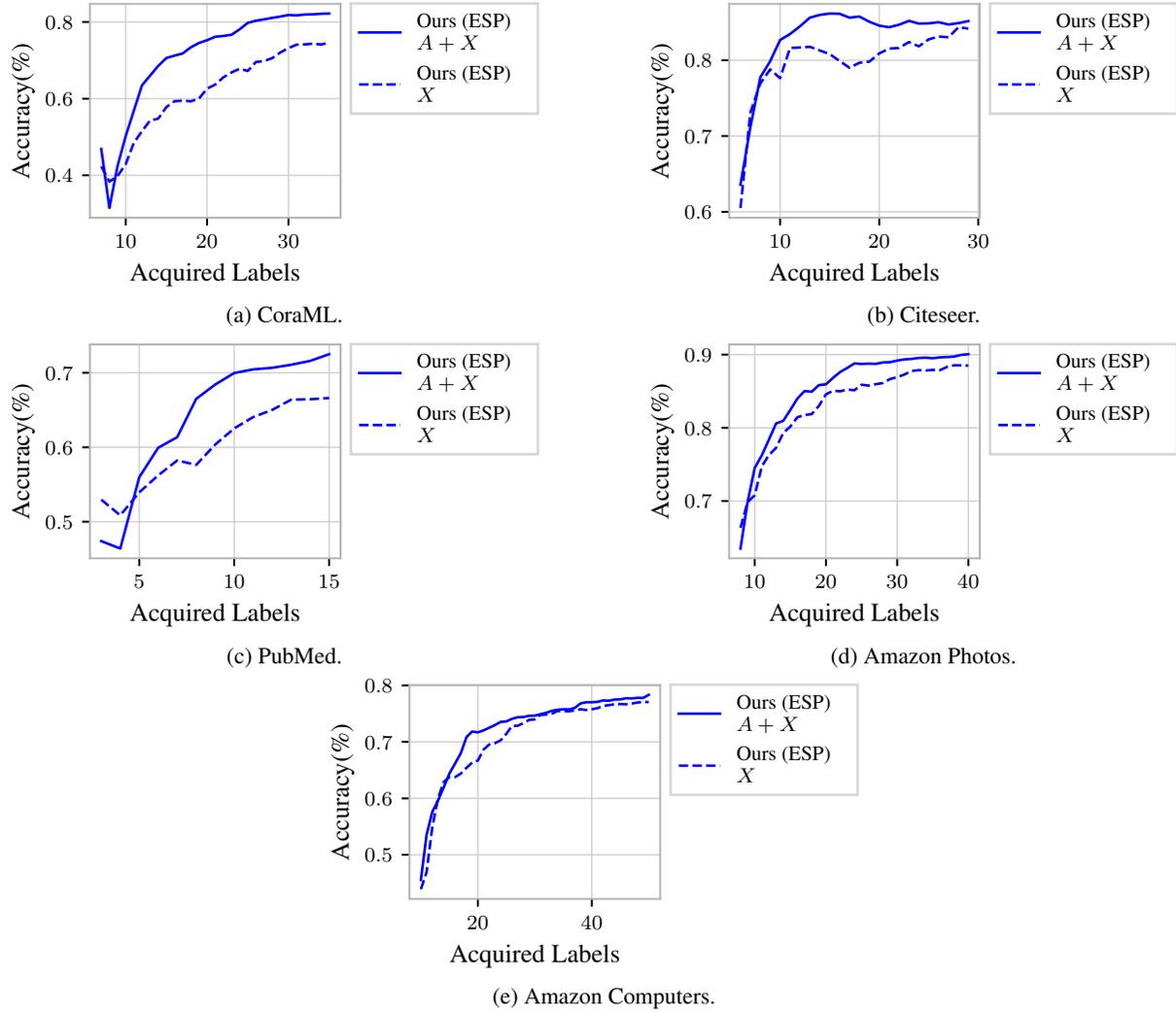

    \centering
    \begin{subfigure}{.49\textwidth}
        \centering
\input{files/pgfs/ours/features_only/cora_ml_esp_features.pgf}
      \caption{\textcolor{black}{CoraML.}}
    \end{subfigure}
     \hfill
     \begin{subfigure}{.49\textwidth}
       \centering
 \input{files/pgfs/ours/features_only/citeseer_esp_features.pgf}
       \caption{\textcolor{black}{Citeseer.}}
     \end{subfigure}
    \begin{subfigure}{.49\textwidth}
        \centering
\input{files/pgfs/ours/features_only/pubmed_esp_features.pgf}
      \caption{\textcolor{black}{PubMed.}}
    \end{subfigure}
     \hfill
     \begin{subfigure}{.49\textwidth}
       \centering
 \input{files/pgfs/ours/features_only/amazon_photos_esp_features.pgf}
       \caption{\textcolor{black}{Amazon Photos.}}
     \end{subfigure}
     \begin{subfigure}{.49\textwidth}
       \centering
 \input{files/pgfs/ours/features_only/amazon_computers_esp_features.pgf}
       \caption{\textcolor{black}{Amazon Computers.}}
     \end{subfigure}
    \caption{\textcolor{black}{US using our proposed approximation (ESP) utilizing the features ($\matr A + \matr X$) and network effects versus only using features ($\matr X$).}}
    \label{fig:ours_features_only}
\end{figure}
 
 \textcolor{black}{\Cref{fig:ours_features_only} shows that in all settings, performance deteriorates when ignoring network effects. This is in line with the discussion in \Cref{sec:iid_data_discussion}: The graph allows a faithful approximation of the underlying generative process and enables the ESP algorithm to identify informative queries. As has been shown by \citep{wu2019simplifying}, modelling instance interdependence sufficiently mitigates the need for complex feature transformations. While a recent line of works regarding Energy-based models (EBMs) \citep{grathwohl2019your} explores how to interpret classifiers on i.i.d. data as surrogates for the data generating process, for such problems typically significantly more complex feature transformations are required. Since we can not rely on strong inductive biases implied by the graph structure, faithfully approximating the generating process and thus also ground-truth uncertainty becomes more challenging.}

\end{document}